\documentclass{article}

\PassOptionsToPackage{numbers, compress}{natbib}
\usepackage[preprint]{neurips_2025}
\input{_preamble}

\usepackage[utf8]{inputenc} %
\usepackage[T1]{fontenc}    %
\usepackage{hyperref}       %
\usepackage{url}            %
\usepackage{booktabs}       %
\usepackage{amsfonts}       %
\usepackage{nicefrac}       %
\usepackage{microtype}      %
\usepackage{xcolor}         %

\title{\ourBenchmark: A Spatio-temporal Scenario Benchmark for Multi-modal Large Language Models in Autonomous Driving}

\author{%
  Christian Fruhwirth-Reisinger$^{1,2,}$\thanks{Equal contribution. Corresponding authors: \texttt{\{reisinger, dusan.malic\}@tugraz.at}} \quad Du\v{s}an Mali\'c$^{1,2,}$\footnotemark[1] \quad Wei Lin$^{3}$ \\ \quad \textbf{David Schinagl}$^{1}$ \quad \textbf{Samuel Schulter} $^{4,}$\thanks{This work is independent of the author’s employment at Amazon.} \quad \textbf{Horst Possegger}$^{1,2}$\\
  \\
  $^1$Institute of Visual Computing, Graz University of Technology\\
  $^2$Christian Doppler Laboratory for Embedded Machine Learning\\
  $^3$Institute for Machine Learning, Johannes Kepler University Linz\\
  $^4$Amazon\\
  \\
  Code: \url{https://github.com/LRP-IVC/STSBench}\\
  Data: \url{https://huggingface.co/datasets/ivc-lrp/STSBench}
}

\begin{document}
\doparttoc %
\faketableofcontents %

\maketitle

\begin{abstract}
    We introduce \ourBenchmark, a scenario-based framework to benchmark the holistic understanding of vision-language models (VLMs) for autonomous driving.
    The framework automatically mines pre-defined traffic scenarios from any dataset using ground-truth annotations, provides an intuitive user interface for efficient human verification, and generates multiple-choice questions for model evaluation. 
    Applied to the NuScenes dataset, we present \ourBenchmarkNus, the first benchmark that evaluates the spatio-temporal reasoning capabilities of VLMs based on comprehensive 3D perception. 
    Existing benchmarks typically target off-the-shelf or fine-tuned VLMs for images or videos from a single viewpoint and focus on semantic tasks such as object recognition, dense captioning, risk assessment, or scene understanding.
    In contrast, \ourBenchmarkNus evaluates driving expert VLMs for end-to-end driving, operating on videos from multi-view cameras or LiDAR. It specifically assesses their ability to reason about both ego-vehicle actions and complex interactions among traffic participants, a crucial capability for autonomous vehicles.
    The benchmark features \nScenarios diverse scenarios spanning multiple views and frames, resulting in \nQuestions human-verified multiple-choice questions. 
    A thorough evaluation uncovers critical shortcomings in existing models’ ability to reason about fundamental traffic dynamics in complex environments. 
    These findings highlight the urgent need for architectural advances that explicitly model spatio-temporal reasoning. 
    By addressing a core gap in spatio-temporal evaluation, \ourBenchmark enables the development of more robust and explainable VLMs for autonomous driving.
\end{abstract}

\section{Introduction}
\label{sec:introduction}
\begin{figure}[t!]
    \centering
    \begin{subfigure}[b]{0.23\textwidth}
        \centering
        \tcbincludegraphics[boxrule=1pt, left=2mm, right=2mm, top=2mm, bottom=2mm, colframe=black, colback=white, width=1\textwidth]{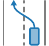}
        \caption{Ego.}
        \label{fig:intro:stsbench_scenario_examples:ego}
    \end{subfigure}
    \hfill
    \begin{subfigure}[b]{0.23\textwidth}
        \centering
        \tcbincludegraphics[boxrule=1pt,left=2mm, right=2mm, top=2mm, bottom=2mm, colframe=black, colback=white, width=1\textwidth]{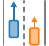}
        \caption{Agent-to-Ego.}
        \label{fig:intro:stsbench_scenario_examples:ego_agent}
    \end{subfigure}
    \hfill
    \begin{subfigure}[b]{0.23\textwidth}
        \centering
        \tcbincludegraphics[boxrule=1pt,left=2mm, right=2mm, top=2mm, bottom=2mm, colframe=black, colback=white, width=1\textwidth]{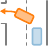}
        \caption{Agent.}
        \label{fig:intro:stsbench_scenario_examples:agent}
    \end{subfigure}
    \hfill
    \begin{subfigure}[b]{0.23\textwidth}
        \centering
        \tcbincludegraphics[boxrule=1pt,left=2mm, right=2mm, top=2mm, bottom=2mm, colframe=black, colback=white, width=1\textwidth]{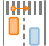}
        \caption{Agent-to-Agent.}
        \label{fig:intro:stsbench_scenario_examples:agent_agent}
    \end{subfigure}
    \caption{\textbf{\ourBenchmark scenario categories.}
    The benchmark covers common ego-vehicle (blue) actions, \eg, \emph{ego lane change}~(\subref{fig:intro:stsbench_scenario_examples:ego}) and interactions with agents (orange), \eg, \emph{ego overtaking agent}~(\subref{fig:intro:stsbench_scenario_examples:ego_agent}), important for vehicle control. 
    In addition, to test for complex spatio-temporal understanding, we evaluate agent actions, \eg, \emph{agent left turn}~(\subref{fig:intro:stsbench_scenario_examples:agent}), and interactions between agents, \eg, \emph{agent waiting for pedestrian to cross}~(\subref{fig:intro:stsbench_scenario_examples:agent_agent}).
    }
    \label{fig:intro:stsbench_scenario_examples}
\end{figure}

The rapid development of increasingly powerful vision-language models (VLMs)~\cite{bai_2023_qwenVL, chen_2024_gpt4far, chen_2024_internvl, dai_2023_instructblip, li_2022_blip, li_2023_blip2, liu_2023_llava, liu_2023_improvedllava, liu_2024_llavanext} has sparked significant interest in applying them to end-to-end autonomous driving~\cite{fu_2025_orion, hwang_2024_emma, jiang_2024_senna, li_2025_gpvl, pan_2024_vlp, shao_2024_lmdrive, song_2025_insightdrive, wang_2024_omnidrive, zhao_2025_sce2drivex, zhou_2025_hermes, zhou_2025_opendrivevla}.
These models aim to enhance trust in fully autonomous systems by providing human-interpretable decisions in natural language~\cite{xie_2025_drivebench}. 
Unlike pre-trained generalist VLMs, these driving experts operate on consecutive multi-view images or LiDAR scans to understand a scene comprehensively and are fine-tuned for planning and controlling the ego vehicle. 

To achieve end-to-end driving, where raw sensor inputs are directly mapped to driving actions, most models~\cite{jiang_2024_senna, wang_2024_omnidrive, zhou_2025_opendrivevla} predict future trajectories through waypoints or control signals for the ego-vehicle. 
A common evaluation strategy is to perform open-loop planning on NuScenes~\cite{caesar_2020_nuscenes}, as most other real-world planning datasets lack raw visual inputs alongside map data and agent trajectories, or are synthetic benchmarks~\cite{jia_2024_benchtodrive} simulated in CARLA~\cite{dosovitskiy_2017_carla}.
A drawback of the open-loop evaluation on \nusc, however, is the relatively small and unbalanced validation split~\cite{zhiqi_2024_isegostatus} in which, for approximately 75\% of all cases, the correct action is "continue to drive straight".
Moreover, evaluating L2 error between predicted and ground truth waypoints, or measuring the accuracy of ego-action predictions, offers limited insight into the decision-making and reasoning capabilities of language-based end-to-end driving experts.
Therefore, further testing of these models is crucial.

A growing number of benchmarks assess the understanding of VLMs in the context of automated driving~\cite{guo_2024_drivemllm, li_2024_codalm, malla_2023_drama, marcu_2024_lingoqa, sachdeva_2024_rank2tell, sima_2024_drivelm, tian_2025_nuscenesspatialqa, xie_2025_drivebench, zeng_2025_dvbench}. 
They typically focus on specific abilities such as spatial reasoning in camera images~\cite{guo_2024_drivemllm, tian_2025_nuscenesspatialqa}, recognition and prediction of ego-vehicle actions~\cite{marcu_2024_lingoqa, sima_2024_drivelm}, the handling of visual corruptions~\cite{xie_2025_drivebench}, or understanding risk~\cite{malla_2023_drama, sachdeva_2024_rank2tell} and critical driving scenarios~\cite{zeng_2025_dvbench}. 
However, most of these benchmarks target general-purpose VLMs that operate on single images or monocular videos and are not designed to evaluate whether models can jointly reason over spatially distributed and temporally extended inputs from multi-view video or LiDAR data, which is an essential capability for coherent understanding in complex, real-world driving scenes.
Furthermore, most existing action- or event-based benchmarks evaluate only the behavior of the ego-vehicle. While this focus is reasonable for planning and control tasks, it overlooks the broader situational understanding required for safe driving, such as anticipating interactions between other traffic participants. Even if they are not of immediate importance, driving models should have the ability to understand such actions or the future consequences of events.

Another major challenge lies in developing annotation schemes for benchmarks applicable to existing perception or planning datasets without requiring extensive manual effort. Existing driving-related benchmarks~\cite{jin_2024_tod3cap, kim_2018_bddx, marcu_2024_lingoqa, sachdeva_2024_rank2tell, sima_2024_drivelm, tian_2024_drivevlm, xie_2025_drivebench} are typically tied to a specific dataset through manual annotations or extensive human verification.
However, although many driving datasets~\cite{caesar_2020_nuscenes, mao_2021_once, sun_2020_waymo, wilson_2021_argoverse2} are recorded using multi-camera and LiDAR systems, their sensor setups differ significantly regarding camera type, placement, and orientation. As a result, a model with 3D understanding~\cite{li_2022_bevformer, wang_2023_streampetr} trained and evaluated on one dataset cannot be straightforwardly assessed using a benchmark built for another, since the projection from images to 3D space depends on the dataset-specific calibration. This makes it necessary to create separate benchmark annotations for each dataset, which is both time-consuming and costly when done manually.

To address the identified issues, we introduce \ourBenchmark, a generalizable framework for automatically mining spatio-temporal driving scenarios from existing datasets with rich ground-truth annotations. 
The framework identifies traffic scenarios, following a pre-defined scenario catalog, that reflect real-world interactions by leveraging information such as 3D bounding boxes and tracks, object class labels, ego-motion data, and HD maps. 
The catalog contains all scenarios and their definitions. It assigns each scenario a list of negative scenarios that do not occur in the same scene.
In addition to this fully automated extraction procedure, we provide a visual inspection tool that enables fast and effortless human verification of the mined scenarios.
Inspectors are tasked to check for false positive scenarios and remove negative ones that also apply for the corresponding scenario (\eg, for a vehicle increasing its speed while performing a \emph{right turn}, \emph{accelerate} would be removed).
From the verified scenarios, \ourBenchmark automatically constructs a multiple-choice benchmark that asks models to identify which interactions occur in a given scene. 
Questions may concern ego-vehicle behavior, the actions of other agents (vehicles, pedestrians, or cyclists), interactions between agents and the ego-vehicle, or between multiple third-party agents. Examples of the different categories are illustrated in~Fig.~\ref{fig:intro:stsbench_scenario_examples}. 
The workflow is simple: automatically mine scenarios, verify them with minimal effort, and convert them into a structured spatio-temporal reasoning task.

Furthermore, we instantiate \ourBenchmark on the validation split of the NuScenes dataset, which remains the most commonly used training and evaluation dataset for vision-language models in autonomous driving. 
Unlike existing benchmarks that focus narrowly on ego-centric actions in images or monocular videos, our benchmark, \ourBenchmarkNus, explicitly targets spatio-temporal reasoning involving both ego and non-ego agents across multiple views and time steps (see Table~\ref{tab:intro:benchmark_comparison}). 
\ourBenchmarkNus comprises \nScenarios scenarios resulting in a total of \nQuestions challenging multiple-choice questions.

We conduct a detailed evaluation of various models that fall into one of three categories: text-only large language models (LLMs), off-the-shelf VLMs, or driving expert VLMs. 
While LLMs receive ground truth trajectories, off-the-shelf VLMs operate on multiple images. Expert models are designed to deal with consecutive multi-view images. 
With just trajectory information available, LLMs outperform both VLM counterparts significantly. 
Our evaluations highlight that state-of-the-art models across all categories provide limited spatio-temporal reasoning capabilities.
This is especially notable for more challenging scenarios (involving interactions between other agents), which require a truly holistic understanding of the scene.

\begingroup
    \setlength\tabcolsep{5.5pt}
    \begin{table}[t!]
        \caption{\textbf{Task-specific driving benchmarks.} 
       Autonomous driving benchmarks created from the \nusc dataset focusing on various tasks grouped by dataset source. $\dagger$:~we do not consider simple status classification annotations such as \emph{moving}, \emph{walking}, \etc as temporal reasoning. Annotation modalities denote single images (S.I.), multi-view images (M.I.), and multi-view videos (M.V.). Evaluation types are: Visual Question Answering~(VQA), Multiple Choice~(MC), Numerical (NUM), and Open-loop Planning (OLP). }
        \label{tab:intro:benchmark_comparison}
         \scriptsize
        \begin{center}
            \begin{tabular*}{\textwidth}{lccccccccc}
                \toprule
                \multirow{2}{*}{Benchmark Name}  & Anno.    & Anno.  & Human  & Spat.  & Temp.            & Multi-view & Third Party & Eval.      & Pub.   \\
                                                 & Modality & Type   & Verif. & Reas.  & Reas.$^\dagger$  & Events     & Interaction & Type       & Avail. \\
                \midrule                                                                                                                                            
                \drivemllm                       & S.I.     & Auto   & \cmark & \cmark & \xmark           & \xmark     & \xmark      & MC, NUM     & \cmark \\
                \nuscenesmqa                     & M.I.     & Auto   & \xmark & \cmark & \xmark           & \xmark     & \xmark      & VQA         & \cmark \\
                \nuscenesqa                      & M.I.     & Auto   & \xmark & \cmark & \xmark           & \xmark     & \xmark      & VQA         & \cmark \\
                \drivelm                         & M.I.     & Manual & \cmark & \cmark & \xmark           & \cmark     & \cmark      & VQA         & \cmark \\
                \nuscenesspatialqa               & M.I.     & Auto   & \xmark & \cmark & \xmark           & \xmark     & \xmark      & MC, NUM     & \xmark \\
                \drivebench                      & M.I.     & Manual & \xmark & \cmark & \xmark           & \cmark     & \cmark      & VQA         & \cmark \\
                \nuinstruct                      & M.V.     & Auto   & \xmark & \cmark & \cmark           & \cmark     & \xmark      & VQA         & \cmark \\
                \todcap                          & M.V.     & Manual & \cmark & \cmark & \xmark           & \xmark     & \xmark      & VQA         & \cmark \\
                \drivelmmoone                    & M.I.     & Auto   & \cmark & \cmark & \cmark           & \xmark     & \xmark      & VQA, MC     & \cmark \\
                \omnidrive                       & M.V.     & Auto   & \cmark & \cmark & \cmark           & \xmark     & \xmark      & VQA, OLP    & \cmark \\
                \midrule                                                                                                                                   
                \rowcolor{gray!20}                                  
                STSBench3D (ours)                & M.V.     & Auto   & \cmark & \cmark & \cmark           & \cmark     & \cmark      & MC          & \cmark \\
                \bottomrule        
            \end{tabular*}
        \end{center}
    \end{table}
\endgroup

\section{Related Work}
\label{sec:relatedwork}
\paragraph{Driving datasets and benchmarks with text annotations.}

Autonomous driving (AD) is an extensive field of research that has led to the creation of numerous datasets~\cite{caesar_2020_nuscenes, Cordts_2016_cityscapes, geiger_2012_kitti, mao_2021_once, sun_2020_waymo,  wilson_2021_argoverse2, yu_2020_bdd} for various perception tasks. 
These datasets have been enriched with text annotations to facilitate language-based model training for specific tasks in the AD domain.
Following common practices in visual instruction tuning~\cite{dai_2023_instructblip, liu_2023_llava, li_2023_blip2}, annotations have been added mostly for separate multi-view images, focusing on tasks such as grounding~\cite{deruyttere_2019_talk2car, vasudevan_2018_cityscapesref}, ego-action prediction~\cite{kim_2018_bddx, xu_2020_bddoia}, open-loop planning~\cite{sima_2024_drivelm, tian_2024_drivevlm}, risk assessment~\cite{ding_2023_hilmd, malla_2023_drama}, spatial reasoning~\cite{guo_2024_drivemllm, tian_2025_nuscenesspatialqa}, or visual question answering~\cite{marcu_2024_lingoqa, nie_2023_reason2drive, li_2024_codalm}.
To address the lack of 3D understanding, EML~\cite{zhou_2024_embodied} introduces text annotations that incorporate question-answer pairs about 2D-to-3D spatio-temporal relations.
The first attempts to evaluate the spatial capabilities of vision-language models for AD are the benchmarks \drivemllm, \nuscenesmqa, and \nuscenesspatialqa. They asses VLMs for their ability to measure distances and understand relative positions within camera images. 
However, since autonomous driving necessitates a holistic understanding of dynamic scenes, various datasets that incorporate multi-view video~\cite{ding_2024_holistic, wang_2024_omnidrive} or 3D~\cite{qian_2024_nuscenesqa, jin_2024_tod3cap, wu_2025_nuprompt, sachdeva_2024_rank2tell} annotations emerged. 
Datasets such as Nuscenes-QA~\cite{qian_2024_nuscenesqa}, DriveLM~\cite{sima_2024_drivelm}, OmniDrive~\cite{wang_2024_omnidrive}, and NuInstruct~\cite{ding_2024_holistic} propose visual question-answering frameworks aimed at scene understanding, chain-of-thought reasoning and counterfactual reasoning.
Despite their extensive annotations, these datasets rely predominantly on question–answer pairs that emphasize semantics~\cite{qian_2024_nuscenesqa, sima_2024_drivelm} and spatial relations~\cite{ding_2024_holistic}, with limited temporal context, particularly beyond ego-vehicle interactions.
\ourBenchmarkNus specifically targets these gaps and tests the spatio-temporal reasoning capabilities of end-to-end driving models. 

\paragraph{Vision-language models for end-to-end driving.}
Vision-Language Models~(VLMs)~\cite{bai_2023_qwenVL, chen_2024_internvl, deitke_2024_molmo, alayrac_2022_flamingo, li_2023_blip2, liu_2023_improvedllava, dai_2023_instructblip} have attracted a lot of attention due to their exceptional zero-shot capabilities.
These capabilities have also raised interest in applying these models to end-to-end AD for open-loop
and closed-loop 
planning. 
Early methods directly apply VLMs to the front-view camera images of an autonomous vehicle~\cite{arai_2024_covla, ma_2024_dolphins, sima_2024_drivelm, xu_2024_drivegpt4, zheng_2024_doe, zhou_2024_embodied} to predict future trajectories or control signals in text form.
However, a holistic understanding of the traffic scene is crucial for realistic driving scenes with highly dynamic scenarios.
Therefore, another line of work operates on multi-view images and videos~\cite{jiang_2024_senna, huang_2024_rdadriver, li_2025_gpvl, wang_2024_omnidrive}. 
To cope with the increasing number of image tokens caused by additional views and multiple video frames, Senna~\cite{jiang_2024_senna} compresses each view via temporal attention for path planning. 
Another technique for dealing with multiple views is encoding into Bird's Eye View (BEV) features that are later aligned with the underlying LLM.
While GPVL~\cite{li_2025_gpvl} and BEV-InMM~\cite{ding_2024_holistic} utilize a BEVFormer~\cite{li_2022_bevformer} backbone, OmniDrive~\cite{wang_2024_omnidrive} uses StreamPETR~\cite{wang_2023_streampetr}. 
More recent approaches combine end-to-end driving models~\cite{hu_2023_uniad, jiang_2023_vad} with VLMs~\cite{fu_2025_orion, pan_2024_vlp, song_2025_insightdrive, zhou_2025_opendrivevla} to increase the contextual understanding and provide reasoning alongside future trajectories. 
Despite their excellent performance on the planning task, the variety of methods and their handling of available input modalities raise questions about their environmental understanding (\ie, reasoning capabilities).
Therefore, we conduct a detailed analysis about the spatio-temporal reasoning capabilities of language-based end-to-end driving models by applying \ourBenchmark to instantiate \ourBenchmarkNus on \nusc data.

\section{Spatio-temporal Scenario Benchmark}
\label{sec:dataset}
With \ourBenchmark, we introduce a benchmark framework designed to evaluate the spatio-temporal reasoning capabilities of vision-language models (VLMs) in autonomous driving.
While most existing benchmarks focus on off-the-shelf or fine-tuned VLMs operating on single images or monocular videos, our benchmark targets expert driving models. 
These experts are expected to have a comprehensive 3D understanding of dynamic scenes and therefore need to process multi-view, LiDAR, or a combined video input signal that enables holistic reasoning.
The development of our benchmark is motivated by two observations:

\paragraph{There is a gap in assessing the spatio-temporal understanding of expert driving models.} 
Recent efforts have adapted VLMs for driving~\cite{hwang_2024_emma, jiang_2024_senna, wang_2024_omnidrive} or extended existing planning models such as UniAD~\cite{hu_2023_uniad} with LLMs~\cite{zhou_2025_opendrivevla} to improve interpretability and trust. However, these expert models are usually evaluated on the \nusc dataset using predicted waypoints or control signals, with metrics such as L2 error or collision rate. Even if these scores are excellent, they do not guarantee that the model’s decisions are grounded in a correct understanding of other traffic participants or scene dynamics.
Although several benchmarks have been developed to test off-the-shelf VLMs for spatial~\cite{guo_2024_drivemllm, inoue_2024_nuscenes_mqa, tian_2025_nuscenesspatialqa} or temporal~\cite{ding_2024_holistic, ishaq_2025_drivelmmo1} reasoning, most are restricted to single frames or monocular views.
NuInstruct~\cite{ding_2024_holistic} remains an exception by providing multi-view video-based questions, but it is automatically generated and lacks human verification, making it better suited for training than evaluation.
Additionally, the temporal reasoning aspect of the NuInstruct benchmark is limited to rather simple motion states, such as whether an agent is moving or stopped.

\paragraph{Existing benchmarks are tailored for a specific dataset and can hardly be transferred or extended by additional scenarios.} 
Encoding a 3D scene from multiple 2D images implicitly requires knowledge of the transformations that relate each image to a shared 3D space. Most recent detection encoders~\cite{li_2022_bevformer, wang_2023_streampetr}, commonly used in end-to-end driving models, learn this mapping during training. As a result, they become tied to a fixed sensor setup for which they can reason accurately about object sizes and distances. This reliance on specific camera setups makes it difficult to evaluate a model trained on one dataset using the benchmark of another, unless domain adaptation or generalization strategies are employed. Therefore, it is essential to have a fast and generalizable framework applicable to various datasets with different sensor configurations. Moreover, existing benchmarks are often built by manually annotating data~\cite{jin_2024_tod3cap, sima_2024_drivelm}, making them time-consuming and expensive to replicate. Even when using the same dataset, adding a new benchmark task typically requires inspecting and re-annotating a large number of samples. These challenges underscore the need for a fast and generalizable framework for benchmark creation.

To address these challenges, we introduce \ourBenchmark, a framework for fast and scalable traffic scenario mining and verification. In Section~\ref{sec:benchmark:framework}, we describe the annotation pipeline that transforms raw dataset annotations into structured multiple-choice evaluations for spatio-temporal reasoning. Section~\ref{sec:benchmark:nuscenes} details the instantiation of this framework on the \nusc dataset, resulting in our \ourBenchmarkNus benchmark.

\subsection{Annotation Framework}
\label{sec:benchmark:framework}
\begin{figure}[t!]
    \centering
        \includegraphics[width=1\textwidth]{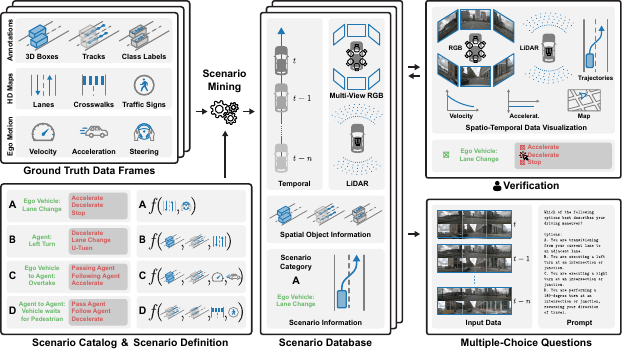}
    \caption{\textbf{\ourBenchmark workflow.} (Best viewed on screen)}
    \label{fig:benchmark:annotation_workflow}
\end{figure}

Our annotation framework aims to create an accurate benchmark for any given perception dataset (\eg,~\cite{caesar_2020_nuscenes, mao_2021_once, sun_2020_waymo}), considering all available ground truth annotations for the respective dataset. 
These annotations include 3D bounding boxes, tracking identifiers, object class labels, ego-motion data (\eg,~velocity, acceleration, steering angle), or HD map data (\eg,~lanes, lane boundaries, speed limits, crosswalks).
In Fig.~\ref{fig:benchmark:annotation_workflow}, we illustrate the benchmark creation workflow of \ourBenchmark. 

\paragraph{1) Scenario catalog.} 
To build a coherent benchmark, we define a scenario catalog containing all relevant scenarios (\eg, lane change, overtake, following, \etc). In addition, we assign negative scenarios for each entry of the catalog. 
Negatives are other scenarios that do not occur in the corresponding scenario.
Assume we have an \emph{overtaking} scenario in which a vehicle in motion passes another vehicle, also in motion, in the adjacent lane. 
Closely related scenarios would be \emph{passing}, where only one of the two vehicles is moving, and \emph{acceleration}, where a vehicle increases its speed.
For an actual \emph{overtaking} scenario, both \emph{passing} and \emph{acceleration} are valuable negatives to test the reasoning capabilities: Holistic spatiotemporal understanding implies that also closely related scenarios can be distinguished.

\paragraph{2) Scenario definition.}
In order to enable automatic mining, we define heuristics for the traffic scenarios we want to test, taking into account available ground truth annotations. 
For instance, an important scenario crucial for driving experts is recognizing \emph{lane changes} for the ego-vehicle and other vehicles in the scene. 
The complexity of this action requires knowing the position of the corresponding vehicle throughout multiple frames \wrt lane boundaries. 
The event gets recognized as such if the boundary is crossed and a vehicle transitions from one lane to another.

\paragraph{3) Scenario mining.}
We automatically mine the pre-defined scenarios and save them in a scenario database. The database contains consecutive images of all available views, LiDAR data, the spatial coordinates of all objects involved in the scenario, and the extracted scenario information, including the found scenario and assigned negatives. 

\paragraph{4) Verification.} 
Human verification is used to ensure annotation quality. Rather than reviewing full sequences frame by frame, the annotator only needs to perform two simple checks: confirm or reject the presence of a mined scenario (false positive) and verify that the negative examples are truly invalid (false negatives). 
For instance, in an \emph{overtaking} scenario, a model might also detect \emph{acceleration}, which could be an invalid negative. These checks are lightweight and fast, allowing for efficient quality control without the burden of traditional manual annotation.

\paragraph{5) Question generation.} 
Finally, \ourBenchmark generates multiple-choice questions asking which of the provided scenario examples occur in the scene. For different scenario types, \ie, ego, agent, ego-to-agent, and agent-to-agent, we provide fixed questions containing the required spatial positions of the occurring objects. The questions are designed to have one correct answer and, by default, provide five possible choices. Further details are provided in Appendix~\ref{sec:supp:details_question_gen}.

\subsection{\ourBenchmarkNus Benchmark Construction}
\label{sec:benchmark:nuscenes}
We leverage \ourBenchmark to mine scenarios and subsequently derive multiple-choice questions from a real-world dataset for evaluating spatio-temporal reasoning capabilities of end-to-end driving models. 

\paragraph{Data Source.}
Since most expert driving models~\cite{huang_2024_drivemm, jin_2024_tod3cap, tian_2024_drivevlm, wang_2024_omnidrive} operate on the multi-view videos or LiDAR scans of NuScenes~\cite{caesar_2020_nuscenes}, we construct our benchmark on this large-scale autonomous driving dataset with rich 3D annotations in a multi-sensor setup. In particular, we automatically gather scenarios from all 150 scenes of the validation set, considering only annotated key frames. Therefore, we leverage manually annotated 3D tracks and class labels, ego-motion data (\eg, velocity) from the inertial measurement unit (IMU), and lanes, lane boundaries, and road markings (\eg, crosswalks) from the available HD map data.

In contrast to prior benchmarks, focusing primarily on ego-vehicle actions that mainly occur in the front-view, \ourBenchmarkNus evaluates spatio-temporal reasoning across a broader set of interactions and multiple views. This includes reasoning about other agents and their interactions with the ego-vehicle or with one another. To support this, we define four distinct scenario categories:

\paragraph{1) Ego-vehicle scenarios.}
The first category includes all actions related exclusively to the ego-vehicle, such as \emph{acceleration/deceleration}, \emph{left/right turn}, or \emph{lane change}. Important for control decisions and collision prevention, driving models must be aware of the ego-vehicle status and behavior. Although these scenarios are part of existing benchmarks in different forms and relatively straightforward to detect, they provide valuable negatives for scenarios with ego-agent interactions.

\paragraph{2) Agent scenarios.}
Similar to ego-vehicle scenarios, agent scenarios involve a single agent in the scene. However, this category additionally contains vulnerable road users such as pedestrians and cyclists. Pedestrians, contrary to vehicles, perform actions such as \emph{walking}, \emph{running}, or \emph{crossing}. Awareness of other traffic participants and their actions is crucial when it comes to risk assessment, planning the next ego action, or analyzing the situation in a dynamic environment. In contrast to ego-vehicle actions, other road users may be occluded or far away and, therefore, pose a particular challenge. 

\paragraph{3) Ego-to-agent scenarios.}
The third category of scenarios describes ego-related agent actions. Directly influencing the driving behavior of each other, this category is similarly important to the ego-vehicle scenarios \wrt the immediate control decisions. Ego-agent scenarios contain maneuvers such as \emph{overtaking}, \emph{passing}, \emph{following}, or \emph{leading}. The scenarios focus on agents in the immediate vicinity of the ego-vehicle and direct interactions.

\paragraph{4) Agent-to-agent scenarios.}
The most challenging group of scenarios concerns interactions between two agents, not considering the ego-vehicle. These scenarios describe the spatio-temporal relationship between objects. For instance, a vehicle that \emph{overtakes} another vehicle in motion or pedestrians \emph{moving alongside} each other. The latter is a perfect example of interactions that do not actively influence the driving behavior of the expert model. However, we argue that a holistic understanding of the scene should not be restricted to the immediate surroundings of the ego-vehicle.

\subsection{Benchmark Statistics}

\begin{figure}
    \begin{subfigure}[b]{0.49\textwidth}
        \includegraphics[width=\linewidth]{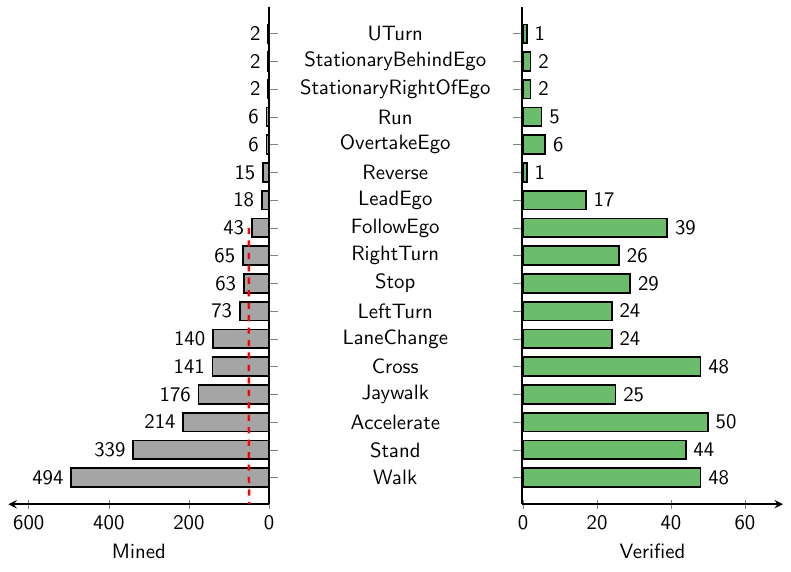}
        \caption*{Agent scenarios}
        \label{fig:benchmark:stsnu_scenarios:agent}
    \end{subfigure}
    \hfill
    \begin{subfigure}[b]{0.49\textwidth}
        \includegraphics[width=\linewidth]{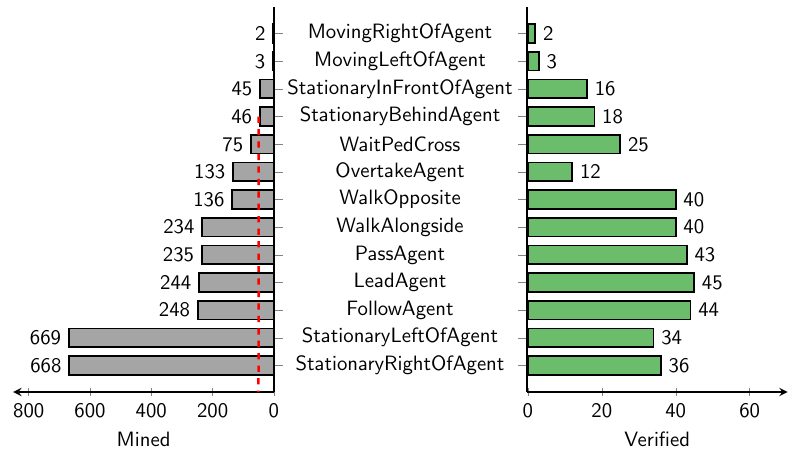}
        \caption*{Agent-to-Agent scenarios}
        \label{fig:benchmark:stsnu_scenarios:agent_agent}
    \end{subfigure}\\[.2em]
    
    \begin{subfigure}[b]{0.49\textwidth}
        \includegraphics[width=\linewidth]{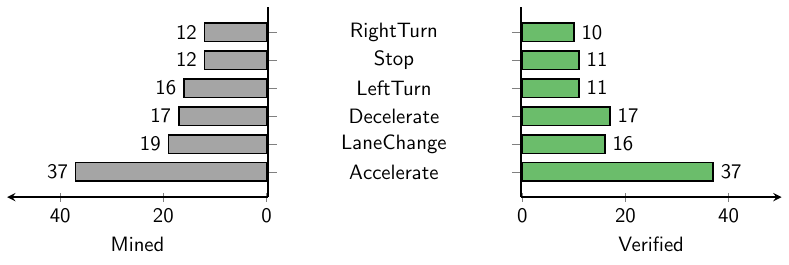}
        \caption*{Ego-vehicle scenarios}
        \label{fig:benchmark:stsnu_scenarios:ego}
    \end{subfigure}
    \hfill
    \begin{subfigure}[b]{0.49\textwidth}
        \includegraphics[width=\linewidth]{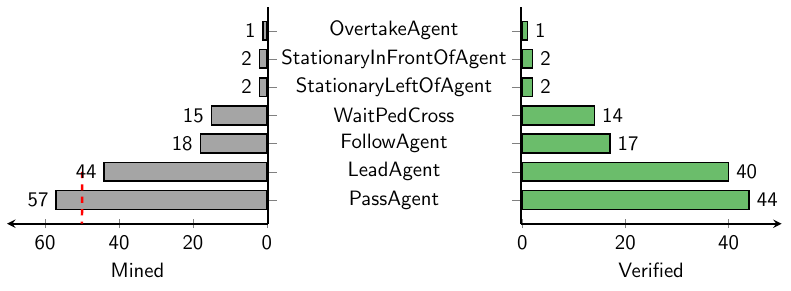}
        \caption*{Ego-to-Agent scenarios}
        \label{fig:benchmark:stsnu_scenarios:ego_agent}
    \end{subfigure}

\caption{\textbf{Scenario statistics.} 
Number of mined scenarios in total (gray) and the remaining samples (green) after sub-sampling and verification. Scenarios with more than 50 samples (dashed red line) have been sub-sampled considering spatial distribution, occlusion, and distance to the ego-vehicle.}
\label{fig:benchmark:stsnu_scenarios}
\end{figure}

\begin{wrapfigure}{r}{0.53\textwidth}
   \centering
   \includegraphics[width=0.51\textwidth]{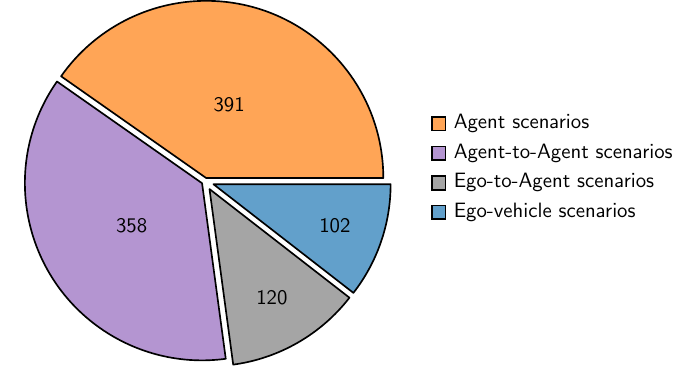}
   \caption{\textbf{Scenario distribution.}
   Number of scenarios per category.}
   \label{fig:benchmark:pieplot}
\end{wrapfigure}

The scenario catalog of our \ourBenchmarkNus benchmark comprises \nScenarios different scenario descriptions. Using this catalog, \ourBenchmark has automatically mined \nScenariosTotal scenarios from 150 sequences of the \nusc validation set. 
To ensure a better balance of the benchmark, we sub-sampled over-represented scenarios based on occlusion rate and spatial distribution of agents. Hence, with this optional step, we removed very difficult examples of highly occluded agents and objects that are far away. 
The remaining  \nScenariosSampled scenarios have gone through human verification and finally resulted in 971 multiple-choice questions offering at least four possible answers per question, of which exactly one is correct. 
We provide detailed numbers of mined, sub-sampled, and remaining scenarios after human verification in Fig.~\ref{fig:benchmark:stsnu_scenarios}. 
The distribution of scenarios assigned to the four proposed categories is illustrated in Fig~\ref{fig:benchmark:pieplot}.
We can see that a large proportion of the scenarios cover the two more difficult scenario categories, \ie, agent and agent-to-agent. 
Despite this, the ego and ego-to-agent scenarios, some of which were also examined in other benchmarks, are sufficiently represented. 
We provide additional statistics and details of \ourBenchmarkNus in Appendix~\ref{sec:supp:benchmark_details}.

\paragraph{Verification and correction.} 
To ensure accuracy, the verification was carried out by three different people with a European class B driver's license. 
Prior to the validation procedure, the human driving experts were briefed to gain a general understanding of the scenarios and what requirements they need to meet. 
The agreement on positive samples was $85.6$\% (1017 agreements out of 1188) while there was a disagreement of $20.8$\% (247 out of 1188) on negative samples. 
There were mainly differences in the perception of recognizable or unrecognizable agents and misunderstandings of interactions between two agents. 
On average, the three inspectors spent $4.78$ hours each verifying the data, which is around $14.5$ seconds per sample and inspector.
A more detailed analysis and verification times are provided in Appendix~\ref{sec:supp:benchmark_details:verification_details}.
For the final \ourBenchmarkNus benchmark, we merged positive scenarios applying a majority voting and kept all negatives with full agreement over all reviewers.

\section{Experiments}
\label{sec:evaluation}
\begin{table}[t]
    \centering
    \caption{\textbf{Overall performance.}
    Performance comparison of LLMs, off-the-shelf VLMs, and driving expert VLMs. Accuracy grouped by scenario categories. The best results are highlighted in bold.}
    \label{tab:experiments:main_evaluation}
    \begin{tabular}{l c c c c c}
    	\toprule
    	& \textbf{Ego} & \textbf{Ego-to-Agent} & \textbf{Agent} & \textbf{Agent-to-Agent} & \textbf{Average} \\
    	\midrule
    	\llama &  16.63 & 19.73 & 13.87 & 21.84 & 18.01 \\
    	\deepseek & 31.09 & 71.49 & 41.98 & 44.24 & 47.20 \\
    	\gptfo & \bestresult{63.63} & \bestresult{75.75} & 45.59 & 43.38 & \bestresult{57.08} \\
    	\midrule
    	\internvlob &  19.98 & 38.17 & 20.47 & 29.88 & 27.12\\
    	\quenvlsb & 35.62 & 26.64 & 36.58 & 37.47 & 34.07 \\
    	\internvlab & 38.38 & 40.82 & \bestresult{53.33} & \bestresult{51.77} & 46.07 \\
    	\midrule
    	\sennavlm & \phantom{0}8.81 & 44.44 & 26.10 & 31.45 & 27.70 \\
    	\omnidrive & 24.40 & 42.97 & 23.78 & 26.15 & 29.33 \\
    	\drivemm & 43.49 & 48.31 & 38.00 & 28.26 & 39.51 \\
    	\bottomrule
    \end{tabular}
\end{table}

The evaluation on \ourBenchmarkNus follows a simple protocol.
For each scenario, we measure accuracy as the proportion of correctly answered multiple-choice questions.
To account for scenario imbalance, we report the overall accuracy as a weighted mean across all scenarios.
Beyond its simplicity, this evaluation method offers the significant advantage of being both interpretable and comparable across models.

\paragraph{Baselines.}
We evaluate a range of models from three categories: large-language models (LLMs), off-the-shelf vision-language models (VLMs), and driving expert VLMs. First, we task LLMs to infer the correct maneuver given ground-truth perception data. Therefore, we selected two open-source models, \deepseek and \llama, and one closed-source model, \gptfo.  
Second, for the evaluation of VLMs without fine-tuning, we use \quenvlsb, \internvlob, and \internvlab. 
Third, as representative driving expert models, we evaluate \sennavlm, \omnidrive, and \drivemm.

\paragraph{Evaluation setting.}
Since only driving expert models are designed to handle multi-view video data, we adapted the input format for LLMs and off-the-shelf VLMs in our evaluation.
To simulate a perfect perception system, we provide LLMs with the GPS positions of the ego-vehicle and, when relevant, the trajectories of involved agents, along with the task description, a multiple-choice question, and scenario definitions for the available answers. 
This setup serves as a simple baseline for comparison.
For off-the-shelf VLMs, we supply a series of images and an adapted text prompt. 
Because scenarios can span multiple viewpoints, we select, at each time step, the image corresponding to the camera view in which the relevant part of the scenario occurs. The associated camera view metadata is also provided.
Driving expert models, on the other hand, receive full multi-view image sequences in addition to the text prompt, in which we refer to the involved agents.
Further implementation details and input formatting for all models are provided in Appendix~\ref{sec:supp:prompt_examples}.

\paragraph{Overall performance and analysis.}
The evaluation results in Table~\ref{tab:experiments:main_evaluation} indicate that VLMs, particularly driving expert VLMs, do not have a spatio-temporal understanding of dynamic traffic scenes. 
Driving expert VLMs are good at basic perception tasks, \ie, observing traffic participants near the vehicle, but struggle with ego-vehicle and agent-to-agent scenarios, thus demonstrating insufficient holistic spatio-temporal understanding.
In comparison, without image inputs, the top-performing LLM outperforms its visual counterparts by a significant margin, especially for relatively simple ego-vehicle ($63.63\%$ vs. $43.49\%$) and ego-to-agent ($75.75\%$ vs. $48.31\%$) scenarios. 
\gptfo, the advanced reasoning model, performs particularly well and reaches an average accuracy of $57.08\%$. 
\drivemm is the best performing expert model on average. 
However, all off-the-shelf VLMs and the top-two LLMs show superior performance in the difficult agent-to-agent scenarios. 

An interesting observation is the considerable gap between \drivemm and the other expert models, particularly for the ego scenarios. 
While OmniDrive projects multi-view image features into BEV, DriveMM directly processes multi-view videos. 
It is worth noting that the StreamPETR~\cite{wang_2023_streampetr} encoder of \omnidrive has initially been designed for perception tasks such as 3D object detection, where it is important to model the surroundings of the ego-vehicle. 
The results suggest that these representations may hamper the reasoning performance \wrt ego actions.
We provide a more detailed analysis in Appendix~\ref{sec:supp:detailed_analysis}.

\section{Conclusion}
\label{sec:conclusion}
In this paper, we introduced \ourBenchmark, a framework for automatic scenario mining from large-scale autonomous driving datasets with rich ground truth annotations. The framework also includes a fast and lightweight verification interface, enabling the effortless creation of high-quality benchmarks for spatio-temporal reasoning in multi-view video or LiDAR data.
We applied \ourBenchmark to the \nus dataset resulting in \ourBenchmarkNus, which comprises \nQuestions multiple-choice questions derived from \nScenarios diverse traffic scenarios. This benchmark provides a rigorous evaluation of driving expert models in terms of their spatio-temporal understanding from a holistic, scene-level perspective. 
Our evaluation revealed that, despite recent progress, current driving expert models still show significant limitations in spatio-temporal reasoning, highlighting the need for further research in this area.

\paragraph{Limitations.}
The major issue with publicly available large-scale driving datasets is that they mostly have clean and homogeneous data collection and filtering processes. For instance, \nus has been recorded in Boston and Singapore, and contains hardly unexpected or dangerous driving behavior. This is perfectly fine for perception tasks and also desirable from a human safety perspective. However, this is disadvantageous for the automatic mining of diverse traffic scenarios. Nevertheless, evaluations on available data already pose significant challenges for driving expert models regarding their spatio-temporal reasoning capabilities.
Another bottleneck of automatic scenario mining is the careful design of heuristics. For example, spatio-temporal processes have variable length. For instance, the time frame to detect \emph{u-turns} is significantly longer than, for example, \emph{lane changes}, especially when the agent has to interrupt the maneuver because of the current traffic situation. However, considering the rich annotations available, a simple set of heuristics can already provide sufficient pre-selection of traffic scenarios.

\paragraph{Social impact.}
Our work contributes to safe automated or autonomous driving systems: With STSnu, we highlight the lack of holistic scene understanding of state-of-the-art models.
To mitigate the limitations (\eg potential geographic bias, cannot test for safety-critical driving behavior due to lack of such data), we also open-source STSBench, a framework to easily extract and (most importantly) efficiently verify such benchmarks from other datasets.
We explicitly rely on heuristics to ensure that the extracted benchmark scenarios are deterministic, easily reproducible and intuitive.
We believe that our framework is a valuable and easy-to-use tool to guide future research on driving expert models towards better holistic scene understanding capabilities, in order to achieve safe and trustworthy systems.

\bibliographystyle{plainnat}
\bibliography{_main}

\begin{thebibliography}{76}
\providecommand{\natexlab}[1]{#1}
\providecommand{\url}[1]{\texttt{#1}}
\expandafter\ifx\csname urlstyle\endcsname\relax
  \providecommand{\doi}[1]{doi: #1}\else
  \providecommand{\doi}{doi: \begingroup \urlstyle{rm}\Url}\fi

\bibitem[AI(2024)]{meta_2025_llama3}
Meta AI.
\newblock {LLaMA 3.2: Open Foundation and Instruction Models}, 2024.

\bibitem[Alayrac et~al.(2022)Alayrac, Donahue, Luc, Miech, Barr, Hasson, Lenc,
  Mensch, Millican, Reynolds, Ring, Rutherford, Cabi, Han, Gong, Samangooei,
  Monteiro, Menick, Borgeaud, Brock, Nematzadeh, Sharifzadeh, Binkowski,
  Barreira, Vinyals, Zisserman, and Simonyan]{alayrac_2022_flamingo}
Jean-Baptiste Alayrac, Jeff Donahue, Pauline Luc, Antoine Miech, Iain Barr,
  Yana Hasson, Karel Lenc, Arthur Mensch, Katie Millican, Malcolm Reynolds,
  Roman Ring, Eliza Rutherford, Serkan Cabi, Tengda Han, Zhitao Gong, Sina
  Samangooei, Marianne Monteiro, Jacob Menick, Sebastian Borgeaud, Andy Brock,
  Aida Nematzadeh, Sahand Sharifzadeh, Mikolaj Binkowski, Ricardo Barreira,
  Oriol Vinyals, Andrew Zisserman, and Karen Simonyan.
\newblock {Flamingo: a Visual Language Model for Few-Shot Learning}.
\newblock In \emph{NeurIPS}, 2022.

\bibitem[Arai et~al.(2024)Arai, Miwa, Sasaki, Yamaguchi, Watanabe, Aoki, and
  Yamamoto]{arai_2024_covla}
Hidehisa Arai, Keita Miwa, Kento Sasaki, Yu~Yamaguchi, Kohei Watanabe, Shunsuke
  Aoki, and Issei Yamamoto.
\newblock {CoVLA: Comprehensive Vision-Language-Action Dataset for Autonomous
  Driving}.
\newblock In \emph{WACV}, 2024.

\bibitem[Bai et~al.(2023)Bai, Bai, Yang, Wang, Tan, Wang, Lin, Zhou, and
  Zhou]{bai_2023_qwenVL}
Jinze Bai, Shuai Bai, Shusheng Yang, Shijie Wang, Sinan Tan, Peng Wang, Junyang
  Lin, Chang Zhou, and Jingren Zhou.
\newblock {Qwen-VL: A Versatile Vision-Language Model for Understanding,
  Localization, Text Reading, and Beyond}.
\newblock \emph{arXiv preprint arXiv:2308.12966}, 2023.

\bibitem[Bai et~al.(2025)Bai, Chen, Liu, Wang, Ge, Song, Dang, Wang, Wang,
  Tang, Zhong, Zhu, Yang, Li, Wan, Wang, Ding, Fu, Xu, Ye, Zhang, Xie, Cheng,
  Zhang, Yang, Xu, and Lin]{shuai_2025_quenvl}
Shuai Bai, Keqin Chen, Xuejing Liu, Jialin Wang, Wenbin Ge, Sibo Song, Kai
  Dang, Peng Wang, Shijie Wang, Jun Tang, Humen Zhong, Yuanzhi Zhu, Mingkun
  Yang, Zhaohai Li, Jianqiang Wan, Pengfei Wang, Wei Ding, Zheren Fu, Yiheng
  Xu, Jiabo Ye, Xi~Zhang, Tianbao Xie, Zesen Cheng, Hang Zhang, Zhibo Yang,
  Haiyang Xu, and Junyang Lin.
\newblock {Qwen2.5-VL Technical Report}.
\newblock \emph{arXiv preprint arXiv:2502.13923}, 2025.

\bibitem[Caesar et~al.(2020)Caesar, Bankiti, Lang, Vora, Liong, Xu, Krishnan,
  Pan, Baldan, and Beijbom]{caesar_2020_nuscenes}
Holger Caesar, Varun Bankiti, Alex~H. Lang, Sourabh Vora, Venice~Erin Liong,
  Qiang Xu, Anush Krishnan, Yu~Pan, Giancarlo Baldan, and Oscar Beijbom.
\newblock {nuScenes: A multimodal dataset for autonomous driving}.
\newblock In \emph{CVPR}, 2020.

\bibitem[Cao et~al.(2024)Cao, Zhou, Ma, Ye, Cui, Tang, Cao, Liang, Wang, Rehg,
  and Zheng]{cao_2024_maplm}
Xu~Cao, Tong Zhou, Yunsheng Ma, Wenqian Ye, Can Cui, Kun Tang, Zhipeng Cao,
  Kaizhao Liang, Ziran Wang, James Rehg, and Chao Zheng.
\newblock {MAPLM: A Real-World Large-Scale Vision-Language Dataset for Map and
  Traffic Scene Understanding}.
\newblock In \emph{CVPR}, 2024.

\bibitem[Chen et~al.(2024{\natexlab{a}})Chen, Wang, Tian, Ye, Gao, Cui, Tong,
  Hu, Luo, Ma, et~al.]{chen_2024_gpt4far}
Zhe Chen, Weiyun Wang, Hao Tian, Shenglong Ye, Zhangwei Gao, Erfei Cui, Wenwen
  Tong, Kongzhi Hu, Jiapeng Luo, Zheng Ma, et~al.
\newblock How far are we to gpt-4v? closing the gap to commercial multimodal
  models with open-source suites.
\newblock \emph{Science China Information Sciences}, 67\penalty0 (12):\penalty0
  220101, 2024{\natexlab{a}}.

\bibitem[Chen et~al.(2024{\natexlab{b}})Chen, Wu, Wang, Su, Chen, Xing, Zhong,
  Zhang, Zhu, Lu, et~al.]{chen_2024_internvl}
Zhe Chen, Jiannan Wu, Wenhai Wang, Weijie Su, Guo Chen, Sen Xing, Muyan Zhong,
  Qinglong Zhang, Xizhou Zhu, Lewei Lu, et~al.
\newblock {Internvl: Scaling up vision foundation models and aligning for
  generic visual-linguistic tasks}.
\newblock In \emph{CVPR}, 2024{\natexlab{b}}.

\bibitem[Cordts et~al.(2016)Cordts, Omran, Ramos, Rehfeld, Enzweiler, Benenson,
  Franke, Roth, and Schiele]{Cordts_2016_cityscapes}
Marius Cordts, Mohamed Omran, Sebastian Ramos, Timo Rehfeld, Markus Enzweiler,
  Rodrigo Benenson, Uwe Franke, Stefan Roth, and Bernt Schiele.
\newblock {The Cityscapes Dataset for Semantic Urban Scene Understanding}.
\newblock In \emph{CVPR}, 2016.

\bibitem[Dai et~al.(2023)Dai, Li, Li, Tiong, Zhao, Wang, Li, Fung, and
  Hoi]{dai_2023_instructblip}
Wenliang Dai, Junnan Li, Dongxu Li, Anthony Meng~Huat Tiong, Junqi Zhao,
  Weisheng Wang, Boyang Li, Pascale Fung, and Steven Hoi.
\newblock {InstructBLIP: Towards General-purpose Vision-Language Models with
  Instruction Tuning}.
\newblock In \emph{NeurIPS}, 2023.

\bibitem[DeepSeek-AI(2024)]{deepseekai_2024_deepseekv3}
DeepSeek-AI.
\newblock Deepseek-v3 technical report.
\newblock \emph{arXiv preprint arXiv:2412.19437}, 2024.

\bibitem[Deitke et~al.(2024)Deitke, Clark, Lee, Tripathi, Yang, Park, Salehi,
  Muennighoff, Lo, Soldaini, Lu, Anderson, Bransom, Ehsani, Ngo, Chen, Patel,
  Yatskar, Callison-Burch, Head, Hendrix, Bastani, VanderBilt, Lambert, Chou,
  Chheda, Sparks, Skjonsberg, Schmitz, Sarnat, Bischoff, Walsh, Newell,
  Wolters, Gupta, Zeng, Borchardt, Groeneveld, Nam, Lebrecht, Wittlif,
  Schoenick, Michel, Krishna, Weihs, Smith, Hajishirzi, Girshick, Farhadi, and
  Kembhavi]{deitke_2024_molmo}
Matt Deitke, Christopher Clark, Sangho Lee, Rohun Tripathi, Yue Yang, Jae~Sung
  Park, Mohammadreza Salehi, Niklas Muennighoff, Kyle Lo, Luca Soldaini, Jiasen
  Lu, Taira Anderson, Erin Bransom, Kiana Ehsani, Huong Ngo, YenSung Chen, Ajay
  Patel, Mark Yatskar, Chris Callison-Burch, Andrew Head, Rose Hendrix, Favyen
  Bastani, Eli VanderBilt, Nathan Lambert, Yvonne Chou, Arnavi Chheda, Jenna
  Sparks, Sam Skjonsberg, Michael Schmitz, Aaron Sarnat, Byron Bischoff, Pete
  Walsh, Chris Newell, Piper Wolters, Tanmay Gupta, Kuo-Hao Zeng, Jon
  Borchardt, Dirk Groeneveld, Crystal Nam, Sophie Lebrecht, Caitlin Wittlif,
  Carissa Schoenick, Oscar Michel, Ranjay Krishna, Luca Weihs, Noah~A. Smith,
  Hannaneh Hajishirzi, Ross Girshick, Ali Farhadi, and Aniruddha Kembhavi.
\newblock {Molmo and PixMo: Open Weights and Open Data for State-of-the-Art
  Vision-Language Models}, 2024.

\bibitem[Deruyttere et~al.(2019)Deruyttere, Vandenhende, Grujicic, Van~Gool,
  and Moens]{deruyttere_2019_talk2car}
Thierry Deruyttere, Simon Vandenhende, Dusan Grujicic, Luc Van~Gool, and
  Marie-Francine Moens.
\newblock {T}alk2{C}ar: Taking control of your self-driving car.
\newblock In \emph{EMNLP-IJCNLP}, 2019.

\bibitem[Ding et~al.(2023)Ding, Han, Xu, Zhang, and Li]{ding_2023_hilmd}
Xinpeng Ding, Jianhua Han, Hang Xu, Wei Zhang, and Xiaomeng Li.
\newblock {HiLM-D: Towards High-Resolution Understanding in Multimodal Large
  Language Models for Autonomous Driving}.
\newblock \emph{arXiv preprint arXiv:2308.12966}, 2023.

\bibitem[Ding et~al.(2024)Ding, Han, Xu, Laing, Hang, Zhang, and
  Li]{ding_2024_holistic}
Xinpeng Ding, Jinahua Han, Hang Xu, Xiaodan Laing, Xu~Hang, Wei Zhang, and
  Xiaomeng Li.
\newblock {Holistic Autonomous Driving Understanding by Bird's-Eye-View
  Injected Multi-Modal Large Models}.
\newblock In \emph{CVPR}, 2024.

\bibitem[Dosovitskiy et~al.(2017)Dosovitskiy, Ros, Codevilla, Lopez, and
  Koltun]{dosovitskiy_2017_carla}
Alexey Dosovitskiy, German Ros, Felipe Codevilla, Antonio Lopez, and Vladlen
  Koltun.
\newblock {CARLA: An Open Urban Driving Simulator}.
\newblock In \emph{CoRL}, 2017.

\bibitem[Fu et~al.(2025)Fu, Zhang, Zhao, Cui, Liang, Zhang, Zhang, Xie, Wang,
  and Bai]{fu_2025_orion}
Haoyu Fu, Diankun Zhang, Zongchuang Zhao, Jianfeng Cui, Dingkang Liang, Chong
  Zhang, Dingyuan Zhang, Hongwei Xie, Bing Wang, and Xiang Bai.
\newblock {ORION: A Holistic End-to-End Autonomous Driving Framework by
  Vision-Language Instructed Action Generation}.
\newblock \emph{arXiv preprint arXiv:2503.19755}, 2025.

\bibitem[Geiger et~al.(2012)Geiger, Lenz, and Urtasun]{geiger_2012_kitti}
Andreas Geiger, Philip Lenz, and Raquel Urtasun.
\newblock {Are we ready for autonomous driving? The KITTI vision benchmark
  suite}.
\newblock In \emph{CVPR}, 2012.

\bibitem[Guo et~al.(2024)Guo, Ruijun, Yiqun, Yuhang, Zhang, and
  Chen]{guo_2024_drivemllm}
Xianda Guo, Zhang Ruijun, Duan Yiqun, He~Yuhang, Chenming Zhang, and Long Chen.
\newblock {DriveMLLM: A Benchmark for Spatial Understanding with Multimodal
  Large Language Models in Autonomous Driving}.
\newblock \emph{arXiv preprint arXiv:2411.13112}, 2024.

\bibitem[Hu et~al.(2023)Hu, Yang, Chen, Li, Sima, Zhu, Chai, Du, Lin, Wang, Lu,
  Jia, Liu, Dai, Qiao, and Li]{hu_2023_uniad}
Yihan Hu, Jiazhi Yang, Li~Chen, Keyu Li, Chonghao Sima, Xizhou Zhu, Siqi Chai,
  Senyao Du, Tianwei Lin, Wenhai Wang, Lewei Lu, Xiaosong Jia, Qiang Liu,
  Jifeng Dai, Yu~Qiao, and Hongyang Li.
\newblock {Planning-oriented Autonomous Driving}.
\newblock In \emph{CVPR}, 2023.

\bibitem[Huang et~al.(2024{\natexlab{a}})Huang, Fen, Yan, Xiao, Jie, Zhong,
  Liang, and Ma]{huang_2024_drivemm}
Zhijian Huang, Chengjian Fen, Feng Yan, Baihui Xiao, Zequn Jie, Yujie Zhong,
  Xiaodan Liang, and Lin Ma.
\newblock Drivemm: All-in-one large multimodal model for autonomous driving.
\newblock \emph{arXiv preprint arXiv:2412.07689}, 2024{\natexlab{a}}.

\bibitem[Huang et~al.(2024{\natexlab{b}})Huang, Tang, Chen, Lin, Jie, Ma, Wang,
  and Liang]{huang_2024_rdadriver}
Zhijian Huang, Tao Tang, Shaoxiang Chen, Sihao Lin, Zequn Jie, Lin Ma, Guangrun
  Wang, and Xiaodan Liang.
\newblock {Making Large Language Models Better Planners with Reasoning-Decision
  Alignment}.
\newblock In \emph{ECCV}, 2024{\natexlab{b}}.

\bibitem[Hwang et~al.(2024)Hwang, Xu, Lin, Hung, Ji, Choi, Huang, He,
  Covington, Sapp, Zhou, Guo, Anguelov, and Tan]{hwang_2024_emma}
Jyh-Jing Hwang, Runsheng Xu, Hubert Lin, Wei-Chih Hung, Jingwei Ji, Kristy
  Choi, Di~Huang, Tong He, Paul Covington, Benjamin Sapp, Yin Zhou, James Guo,
  Dragomir Anguelov, and Mingxing Tan.
\newblock Emma: End-to-end multimodal model for autonomous driving.
\newblock \emph{arXiv preprint arXiv:2410.23262}, 2024.

\bibitem[Inoue et~al.(2024)Inoue, Yada, Tanahashi, and
  Yamaguchi]{inoue_2024_nuscenes_mqa}
Yuichi Inoue, Yuki Yada, Kotaro Tanahashi, and Yu~Yamaguchi.
\newblock {NuScenes-MQA: Integrated Evaluation of Captions and QA for
  Autonomous Driving Datasets Using Markup Annotations}.
\newblock In \emph{WACVW}, 2024.

\bibitem[Ishaq et~al.(2025)Ishaq, Lahoud, More, Thawakar, Thawkar, Dissanayake,
  Ahsan, Li, Khan, Cholakkal, Laptev, Anwer, and Khan]{ishaq_2025_drivelmmo1}
Ayesha Ishaq, Jean Lahoud, Ketan More, Omkar Thawakar, Ritesh Thawkar, Dinura
  Dissanayake, Noor Ahsan, Yuhao Li, Fahad~Shahbaz Khan, Hisham Cholakkal, Ivan
  Laptev, Rao~Muhammad Anwer, and Salman Khan.
\newblock Drivelmm-o1: A step-by-step reasoning dataset and large multimodal
  model for driving scenario understanding.
\newblock \emph{arXiv preprint arXiv:2503.10621}, 2025.

\bibitem[Jia et~al.(2024)Jia, Yang, Li, Zhang, and Yan]{jia_2024_benchtodrive}
Xiaosong Jia, Zhenjie Yang, Qifeng Li, Zhiyuan Zhang, and Junchi Yan.
\newblock {Bench2Drive: Towards Multi-Ability Benchmarking of Closed-Loop
  End-To-End Autonomous Driving}.
\newblock In \emph{NeurIPS}, 2024.

\bibitem[Jiang et~al.(2023)Jiang, Chen, Xu, Liao, Chen, Zhou, Zhang, Liu,
  Huang, and Wang]{jiang_2023_vad}
Bo~Jiang, Shaoyu Chen, Qing Xu, Bencheng Liao, Jiajie Chen, Helong Zhou, Qian
  Zhang, Wenyu Liu, Chang Huang, and Xinggang Wang.
\newblock {VAD: Vectorized Scene Representation for Efficient Autonomous
  Driving}.
\newblock In \emph{ICCV}, 2023.

\bibitem[Jiang et~al.(2024)Jiang, Chen, Liao, Zhang, Yin, Zhang, Huang, Liu,
  and Wang]{jiang_2024_senna}
Bo~Jiang, Shaoyu Chen, Bencheng Liao, Xingyu Zhang, Wei Yin, Qian Zhang, Chang
  Huang, Wenyu Liu, and Xinggang Wang.
\newblock {Senna: Bridging Large Vision-Language Models and End-to-End
  Autonomous Driving}.
\newblock \emph{arXiv preprint arXiv:2410.22313}, 2024.

\bibitem[Jin et~al.(2024)Jin, Zheng, Li, Li, Zheng, Hu, Liu, Zhu, Yan, Sun,
  Zhan, Jia, Long, Chen, and Zhao]{jin_2024_tod3cap}
Bu~Jin, Yupeng Zheng, Pengfei Li, Weize Li, Yuhang Zheng, Sujie Hu, Xinyu Liu,
  Jinwei Zhu, Zhijie Yan, Haiyang Sun, Kun Zhan, Peng Jia, Xiaoxiao Long, Yilun
  Chen, and Hao Zhao.
\newblock {TOD3Cap}: {Towards} {3D} {Dense} {Captioning} in {Outdoor} {Scenes}.
\newblock In \emph{ECCV}, 2024.

\bibitem[Kim et~al.(2018)Kim, Rohrbach, Darrell, Canny, and
  Akata]{kim_2018_bddx}
Jinkyu Kim, Anna Rohrbach, Trevor Darrell, John Canny, and Zeynep Akata.
\newblock {Textual Explanations for Self-Driving Vehicles}.
\newblock In \emph{ECCV}, 2018.

\bibitem[Li et~al.(2022{\natexlab{a}})Li, Li, Xiong, and Hoi]{li_2022_blip}
Junnan Li, Dongxu Li, Caiming Xiong, and S.~Hoi.
\newblock {BLIP}: {Bootstrapping} {Language}-{Image} {Pre}-training for
  {Unified} {Vision}-{Language} {Understanding} and {Generation}.
\newblock In \emph{ICML}, 2022{\natexlab{a}}.

\bibitem[Li et~al.(2023)Li, Li, Savarese, and Hoi]{li_2023_blip2}
Junnan Li, Dongxu Li, Silvio Savarese, and Steven Hoi.
\newblock {BLIP}-2: bootstrapping language-image pre-training with frozen image
  encoders and large language models.
\newblock In \emph{ICML}, 2023.

\bibitem[Li et~al.(2025{\natexlab{a}})Li, Wang, Li, Liao, He, and
  Peng]{li_2025_gpvl}
Tengpeng Li, Hanli Wang, Xianfei Li, Wenlong Liao, Tao He, and Pai Peng.
\newblock {Generative Planning with 3D-vision Language Pre-training for
  End-to-End Autonomous Driving}.
\newblock \emph{arXiv preprint arXiv:2501.08861}, 2025{\natexlab{a}}.

\bibitem[Li et~al.(2025{\natexlab{b}})Li, Zhang, Chen, Liu, Li, Gao, Hong,
  Tian, Zhao, Li, et~al.]{li_2024_codalm}
Yanze Li, Wenhua Zhang, Kai Chen, Yanxin Liu, Pengxiang Li, Ruiyuan Gao,
  Lanqing Hong, Meng Tian, Xinhai Zhao, Zhenguo Li, et~al.
\newblock {Automated Evaluation of Large Vision-Language Models on Self-driving
  Corner Cases}.
\newblock In \emph{WACV}, 2025{\natexlab{b}}.

\bibitem[Li et~al.(2022{\natexlab{b}})Li, Wang, Li, Xie, Sima, Lu, Qiao, and
  Dai]{li_2022_bevformer}
Zhiqi Li, Wenhai Wang, Hongyang Li, Enze Xie, Chonghao Sima, Tong Lu, Yu~Qiao,
  and Jifeng Dai.
\newblock {BEVFormer: Learning Bird’s-Eye-View Representation from
  Multi-Camera Images via Spatiotemporal Transformers}.
\newblock In \emph{ECCV}, 2022{\natexlab{b}}.

\bibitem[Li et~al.(2024)Li, Yu, Lan, Li, Kautz, Lu, and
  Alvarez]{zhiqi_2024_isegostatus}
Zhiqi Li, Zhiding Yu, Shiyi Lan, Jiahan Li, Jan Kautz, Tong Lu, and Jose~M.
  Alvarez.
\newblock { Is Ego Status All You Need for Open-Loop End-to-End Autonomous
  Driving? }.
\newblock In \emph{CVPR}, 2024.

\bibitem[Liu et~al.(2023)Liu, Li, Wu, and Lee]{liu_2023_llava}
Haotian Liu, Chunyuan Li, Qingyang Wu, and Yong~Jae Lee.
\newblock {Visual Instruction Tuning}.
\newblock In \emph{NeurIPS}, 2023.

\bibitem[Liu et~al.(2024{\natexlab{a}})Liu, Li, Li, and
  Lee]{liu_2023_improvedllava}
Haotian Liu, Chunyuan Li, Yuheng Li, and Yong~Jae Lee.
\newblock {Improved Baselines with Visual Instruction Tuning}.
\newblock In \emph{CVPR}, 2024{\natexlab{a}}.

\bibitem[Liu et~al.(2024{\natexlab{b}})Liu, Li, Li, Li, Zhang, Shen, and
  Lee]{liu_2024_llavanext}
Haotian Liu, Chunyuan Li, Yuheng Li, Bo~Li, Yuanhan Zhang, Sheng Shen, and
  Yong~Jae Lee.
\newblock Llava-next: Improved reasoning, ocr, and world knowledge,
  2024{\natexlab{b}}.

\bibitem[Ma et~al.(2024)Ma, Cao, Sun, Pavone, and Xiao]{ma_2024_dolphins}
Yingzi Ma, Yulong Cao, Jiachen Sun, Marco Pavone, and Chaowei Xiao.
\newblock {Dolphins: Multimodal Language Model for Driving}.
\newblock In \emph{ECCV}, 2024.

\bibitem[Malla et~al.(2023)Malla, Choi, Dwivedi, Choi, and
  Li]{malla_2023_drama}
Srikanth Malla, Chiho Choi, Isht Dwivedi, Joon~Hee Choi, and Jiachen Li.
\newblock {DRAMA: Joint Risk Localization and Captioning in Driving}.
\newblock In \emph{WACV}, 2023.

\bibitem[Mao et~al.(2021)Mao, Niu, Jiang, Liang, Li, Ye, Zhang, Li, Yu, and
  Xu]{mao_2021_once}
Jiageng Mao, Minzhe Niu, Chenhan Jiang, Xiaodan Liang, Yamin Li, Chaoqiang Ye,
  Wei Zhang, Zhenguo Li, Jie Yu, and Chunjing Xu.
\newblock {One Million Scenes for Autonomous Driving: ONCE Dataset}.
\newblock In \emph{NeurIPS}, 2021.

\bibitem[Marcu et~al.(2024)Marcu, Chen, Hünermann, Karnsund, Hanotte,
  Chidananda, Nair, Badrinarayanan, Kendall, Shotton, and
  Sinavski]{marcu_2024_lingoqa}
Ana-Maria Marcu, Long Chen, Jan Hünermann, Alice Karnsund, Benoit Hanotte,
  Prajwal Chidananda, Saurabh Nair, Vijay Badrinarayanan, Alex Kendall, Jamie
  Shotton, and Oleg Sinavski.
\newblock {LingoQA: Video Question Answering for Autonomous Driving}.
\newblock In \emph{ECCV}, 2024.

\bibitem[Nie et~al.(2024)Nie, Peng, Wang, Cai, Han, Xu, and
  Zhang]{nie_2023_reason2drive}
Ming Nie, Renyuan Peng, Chunwei Wang, Xinyue Cai, Jianhua Han, Hang Xu, and
  Li~Zhang.
\newblock {Reason2Drive: Towards Interpretable and Chain-based Reasoning for
  Autonomous Driving}.
\newblock In \emph{ECCV}, 2024.

\bibitem[OpenAI et~al.(2024)OpenAI, Achiam, Adler, Agarwal, Ahmad, Akkaya,
  Aleman, Almeida, Altenschmidt, Altman, Anadkat, and et~al.]{openai_2024_gpt4}
OpenAI, Josh Achiam, Steven Adler, Sandhini Agarwal, Lama Ahmad, Ilge Akkaya,
  Florencia~Leoni Aleman, Diogo Almeida, Janko Altenschmidt, Sam Altman,
  Shyamal Anadkat, and et~al.
\newblock {GPT-4 Technical Report}.
\newblock \emph{arXiv preprint arXiv:2303.08774}, 2024.

\bibitem[Pan et~al.(2024)Pan, Yaman, Nesti, Mallik, Allievi, Velipasalar, and
  Ren]{pan_2024_vlp}
Chenbin Pan, Burhaneddin Yaman, Tommaso Nesti, Abhirup Mallik, Alessandro~G
  Allievi, Senem Velipasalar, and Liu Ren.
\newblock {VLP: Vision Language Planning for Autonomous Driving}.
\newblock In \emph{CVPR}, 2024.

\bibitem[Qian et~al.(2024)Qian, Chen, Zhuo, Jiao, and
  Jiang]{qian_2024_nuscenesqa}
Tianwen Qian, Jingjing Chen, Linhai Zhuo, Yang Jiao, and Yu-Gang Jiang.
\newblock {NuScenes}-{QA}: {A} {Multi}-{Modal} {Visual} {Question} {Answering}
  {Benchmark} for {Autonomous} {Driving} {Scenario}.
\newblock In \emph{AAAI}, 2024.

\bibitem[Radford et~al.(2021)Radford, Wook~Kim, Hallacy, Ramesh, Goh, Agarwal,
  Sastry, Askell, Mishkin, Clark, Krueger, and Sutskever]{radford_2021_clip}
Alec Radford, Jong Wook~Kim, Chris Hallacy, Aditya Ramesh, Gabriel Goh,
  Sandhini Agarwal, Girish Sastry, Amanda Askell, Pamela Mishkin, Jack Clark,
  Gretchen Krueger, and Ilya Sutskever.
\newblock {Learning Transferable Visual Models From Natural Language
  Supervision}.
\newblock In \emph{ICML}, 2021.

\bibitem[{Rerun Development Team}(2024)]{rerunteam_2025_rerunsdk}
{Rerun Development Team}.
\newblock {Rerun: A Visualization SDK for Multimodal Data}, 2024.
\newblock URL \url{https://www.rerun.io}.
\newblock Available from https://www.rerun.io/ and
  https://github.com/rerun-io/rerun.

\bibitem[Sachdeva et~al.(2024)Sachdeva, Agarwal, Chundi, Roelofs, Li,
  Kochenderfer, Choi, and Dariush]{sachdeva_2024_rank2tell}
Enna Sachdeva, Nakul Agarwal, Suhas Chundi, Sean Roelofs, Jiachen Li, Mykel
  Kochenderfer, Chiho Choi, and Behzad Dariush.
\newblock {Rank2Tell}: {A} {Multimodal} {Driving} {Dataset} for {Joint}
  {Importance} {Ranking} and {Reasoning}.
\newblock In \emph{WACV}, 2024.

\bibitem[Shao et~al.(2024)Shao, Hu, Wang, Song, Waslander, Liu, and
  Li]{shao_2024_lmdrive}
Hao Shao, Yuxuan Hu, Letian Wang, Guanglu Song, Steven~L. Waslander, Yu~Liu,
  and Hongsheng Li.
\newblock {LMDrive: Closed-Loop End-to-End Driving with Large Language Models}.
\newblock In \emph{CVPR}, 2024.

\bibitem[Sima et~al.(2024)Sima, Renz, Chitta, Chen, Zhang, Xie, Luo, Geiger,
  and Li]{sima_2024_drivelm}
Chonghao Sima, Katrin Renz, Kashyap Chitta, Li~Chen, Hanxue Zhang, Chengen Xie,
  Ping Luo, Andreas Geiger, and Hongyang Li.
\newblock {DriveLM}: {Driving} with {Graph} {Visual} {Question} {Answering}.
\newblock In \emph{ECCV}, 2024.

\bibitem[Song et~al.(2025)Song, Guo, Wu, Wei, and Chen]{song_2025_insightdrive}
Ruiqi Song, Xianda Guo, Hangbin Wu, Qinggong Wei, and Long Chen.
\newblock {InsightDrive: Insight Scene Representation for End-to-End Autonomous
  Driving}.
\newblock \emph{arXiv preprint arXiv:2503.13047}, 2025.

\bibitem[Sun et~al.(2020)Sun, Kretzschmar, Dotiwalla, Chouard, Patnaik, Tsui,
  Guo, Zhou, Chai, Caine, et~al.]{sun_2020_waymo}
Pei Sun, Henrik Kretzschmar, Xerxes Dotiwalla, Aurelien Chouard, Vijaysai
  Patnaik, Paul Tsui, James Guo, Yin Zhou, Yuning Chai, Benjamin Caine, et~al.
\newblock {Scalability in Perception for Autonomous Driving: Waymo Open
  Dataset}.
\newblock In \emph{CVPR}, 2020.

\bibitem[Tian et~al.(2025)Tian, Mao, Zhang, Jiang, Zhou, and
  Tu]{tian_2025_nuscenesspatialqa}
Kexin Tian, Jingrui Mao, Yunlong Zhang, Jiwan Jiang, Yang Zhou, and Zhengzhong
  Tu.
\newblock {NuScenes-SpatialQA: A Spatial Understanding and Reasoning Benchmark
  for Vision-Language Models in Autonomous Driving}.
\newblock \emph{arXiv preprint arXiv:2504.03164}, 2025.

\bibitem[Tian et~al.(2024)Tian, Gu, Li, Liu, Zhao, Wang, Zhan, Jia, Lang, and
  Zhao]{tian_2024_drivevlm}
Xiaoyu Tian, Junru Gu, Bailin Li, Yicheng Liu, Zhiyong Zhao, Yang Wang, Kun
  Zhan, Peng Jia, Xianpeng Lang, and Hang Zhao.
\newblock {DriveVLM: The Convergence of Autonomous Driving and Large
  Vision-Language Models}.
\newblock In \emph{CoRL}, 2024.

\bibitem[Vasudevan et~al.(2018)Vasudevan, Dai, and
  Van~Gool]{vasudevan_2018_cityscapesref}
Arun~Balajee Vasudevan, Dengxin Dai, and Luc Van~Gool.
\newblock {Object Referring in Videos With Language and Human Gaze}.
\newblock In \emph{CVPR}, 2018.

\bibitem[Wang et~al.(2023)Wang, Liu, Wang, Li, and Zhang]{wang_2023_streampetr}
Shihao Wang, Yingfei Liu, Tiancai Wang, Ying Li, and Xiangyu Zhang.
\newblock {Exploring Object-Centric Temporal Modeling for Efficient Multi-View
  3D Object Detection}.
\newblock In \emph{ICCV}, 2023.

\bibitem[Wang et~al.(2024)Wang, Yu, Jiang, Lan, Shi, Chang, Kautz, Li, and
  Alvarez]{wang_2024_omnidrive}
Shihao Wang, Zhiding Yu, Xiaohui Jiang, Shiyi Lan, Min Shi, Nadine Chang, Jan
  Kautz, Ying Li, and Jose~M. Alvarez.
\newblock {OmniDrive}: A holistic llm-agent framework for autonomous driving
  with 3d perception, reasoning and planning.
\newblock \emph{arXiv preprint arXiv:2405.01533}, 2024.

\bibitem[Wei et~al.(2022)Wei, Wang, Schuurmans, Bosma, Xia, Chi, Le, Zhou,
  et~al.]{wei_2022_cot}
Jason Wei, Xuezhi Wang, Dale Schuurmans, Maarten Bosma, Fei Xia, Ed~Chi, Quoc~V
  Le, Denny Zhou, et~al.
\newblock Chain-of-thought prompting elicits reasoning in large language
  models.
\newblock \emph{NeurIPS}, 2022.

\bibitem[Wilson et~al.(2021)Wilson, Qi, Agarwal, Lambert, Singh, Khandelwal,
  Pan, Kumar, Hartnett, Pontes, Ramanan, Carr, and
  Hays]{wilson_2021_argoverse2}
Benjamin Wilson, William Qi, Tanmay Agarwal, John Lambert, Jagjeet Singh,
  Siddhesh Khandelwal, Bowen Pan, Ratnesh Kumar, Andrew Hartnett,
  Jhony~Kaesemodel Pontes, Deva Ramanan, Peter Carr, and James Hays.
\newblock {Argoverse 2: Next Generation Datasets for Self-driving Perception
  and Forecasting}.
\newblock In \emph{NeurIPS}, 2021.

\bibitem[Winter et~al.(2025)Winter, Azer, and Flohr]{winter_2025_bevdriver}
Katharina Winter, Mark Azer, and Fabian~B. Flohr.
\newblock {BEVDriver: Leveraging BEV Maps in LLMs for Robust Closed-Loop
  Driving}.
\newblock \emph{arXiv preprint arXiv:2503.03074}, 2025.

\bibitem[Wu et~al.(2025)Wu, Han, Wang, Liu, Zhang, and Shen]{wu_2025_nuprompt}
Dongming Wu, Wencheng Han, Tiancai Wang, Yingfei Liu, Xiangyu Zhang, and
  Jianbing Shen.
\newblock Language {Prompt} for {Autonomous} {Driving}.
\newblock In \emph{AAAI}, 2025.

\bibitem[Xie et~al.(2025)Xie, Kong, Dong, Sima, Zhang, Chen, Liu, and
  Pan]{xie_2025_drivebench}
Shaoyuan Xie, Lingdong Kong, Yuhao Dong, Chonghao Sima, Wenwei Zhang, Qi~Alfred
  Chen, Ziwei Liu, and Liang Pan.
\newblock {Are VLMs Ready for Autonomous Driving? An Empirical Study from the
  Reliability, Data, and Metric Perspectives}.
\newblock \emph{arXiv preprint arXiv:2501.04003}, 2025.

\bibitem[Xu et~al.(2020)Xu, Yang, Gong, Lin, Wu, Li, and
  Vasconcelos]{xu_2020_bddoia}
Yiran Xu, Xiaoyin Yang, Lihang Gong, Hsuan-Chu Lin, Tz-Ying Wu, Yunsheng Li,
  and Nuno Vasconcelos.
\newblock { Explainable Object-Induced Action Decision for Autonomous Vehicles
  }.
\newblock In \emph{CVPR}, 2020.

\bibitem[Xu et~al.(2024)Xu, Zhang, Xie, Zhao, Guo, Wong, Li, and
  Zhao]{xu_2024_drivegpt4}
Zhenhua Xu, Yujia Zhang, Enze Xie, Zhen Zhao, Yong Guo, Kwan-Yee~K Wong,
  Zhenguo Li, and Hengshuang Zhao.
\newblock {Drivegpt4: Interpretable end-to-end autonomous driving via large
  language model}.
\newblock \emph{IEEE Robotics and Automation Letters}, 2024.

\bibitem[Yu et~al.(2020)Yu, Chen, Wang, Xian, Chen, Liu, Madhavan, and
  Darrell]{yu_2020_bdd}
Fisher Yu, Haofeng Chen, Xin Wang, Wenqi Xian, Yingying Chen, Fangchen Liu,
  Vashisht Madhavan, and Trevor Darrell.
\newblock {BDD100K: A Diverse Driving Dataset for Heterogeneous Multitask
  Learning}.
\newblock In \emph{CVPR}, June 2020.

\bibitem[Zeng et~al.(2025)Zeng, Wu, Shi, Zhou, and Guo]{zeng_2025_dvbench}
Tong Zeng, Longfeng Wu, Liang Shi, Dawei Zhou, and Feng Guo.
\newblock {Are Vision LLMs Road-Ready? A Comprehensive Benchmark for
  Safety-Critical Driving Video Understanding}.
\newblock \emph{arXiv preprint arXiv:2504.14526}, 2025.

\bibitem[Zhai et~al.(2023)Zhai, Mustafa, Kolesnikov, and
  Beyer]{zhai_2023_siglip}
Xiaohua Zhai, Basil Mustafa, Alexander Kolesnikov, and Lucas Beyer.
\newblock {Sigmoid Loss for Language Image Pre-Training}.
\newblock In \emph{ICCV}, October 2023.

\bibitem[Zhao et~al.(2025)Zhao, Yuan, Li, Hu, Li, Zheng, and
  Gao]{zhao_2025_sce2drivex}
Rui Zhao, Qirui Yuan, Jinyu Li, Haofeng Hu, Yun Li, Chengyuan Zheng, and Fei
  Gao.
\newblock {Sce2DriveX: A Generalized MLLM Framework for Scene-to-Drive
  Learning}, 2025.

\bibitem[Zheng et~al.(2024{\natexlab{a}})Zheng, Zhou, Meng, Zhou, and
  Huang]{zheng_2024_llmmcqbias}
Chujie Zheng, Hao Zhou, Fandong Meng, Jie Zhou, and Minlie Huang.
\newblock {Large Language Models Are Not Robust Multiple Choice Selectors}.
\newblock In \emph{ICLR}, 2024{\natexlab{a}}.

\bibitem[Zheng et~al.(2024{\natexlab{b}})Zheng, Xia, Huang, Zuo, Zhou, and
  Lu]{zheng_2024_doe}
Wenzhao Zheng, Zetian Xia, Yuanhui Huang, Sicheng Zuo, Jie Zhou, and Jiwen Lu.
\newblock {Doe-1: Closed-Loop Autonomous Driving with Large World Model}.
\newblock \emph{arXiv preprint arXiv: 2412.09627}, 2024{\natexlab{b}}.

\bibitem[Zhou et~al.(2025{\natexlab{a}})Zhou, Liang, Tu, Chen, Ding, Zhang,
  Tan, Zhao, and Bai]{zhou_2025_hermes}
Xin Zhou, Dingkang Liang, Sifan Tu, Xiwu Chen, Yikang Ding, Dingyuan Zhang,
  Feiyang Tan, Hengshuang Zhao, and Xiang Bai.
\newblock {HERMES: A Unified Self-Driving World Model for Simultaneous 3D Scene
  Understanding and Generation}.
\newblock \emph{arXiv preprint arXiv:2501.14729}, 2025{\natexlab{a}}.

\bibitem[Zhou et~al.(2025{\natexlab{b}})Zhou, Han, Yang, Ma, and
  Knoll]{zhou_2025_opendrivevla}
Xingcheng Zhou, Xuyuan Han, Feng Yang, Yunpu Ma, and Alois~C. Knoll.
\newblock {OpenDriveVLA: Towards End-to-end Autonomous Driving with Large
  Vision Language Action Model}.
\newblock \emph{arXiv preprint arXiv:2503.23463}, 2025{\natexlab{b}}.

\bibitem[Zhou et~al.(2024)Zhou, Huang, Bu, Zeng, Li, Qiu, Zhu, Guo, Qiao, and
  Li]{zhou_2024_embodied}
Yunsong Zhou, Linyan Huang, Qingwen Bu, Jia Zeng, Tianyu Li, Hang Qiu, Hongzi
  Zhu, Minyi Guo, Yu~Qiao, and Hongyang Li.
\newblock {Embodied Understanding of Driving Scenarios}.
\newblock In \emph{ECCV}, 2024.

\end{thebibliography}

\clearpage
\appendix

\addcontentsline{toc}{section}{Appendix} %
\part{Appendix} %
\parttoc %
\clearpage
\section{Benchmark details} 
\label{sec:supp:benchmark_details}
\subsection{\ourBenchmarkNus statistics}
\begin{figure*}[t!]
    \centering
    \begin{subfigure}[t]{0.49\textwidth}
        \centering
        \begin{tikzpicture}
\pie[square,
     radius=2.52,
    color={C0!70, C1!70, C2!70, C3!70}
    ]
    {55/1, 30/2, 11/3, 4/4}
\end{tikzpicture}
        \caption{Number of camera views per scenario.}
        \label{fig:supp:stats:stsnu_views_per_scenario}
    \end{subfigure}%
    \hfill
    \begin{subfigure}[t]{0.49\textwidth}
        \centering
        \includegraphics[width=1.0\textwidth]{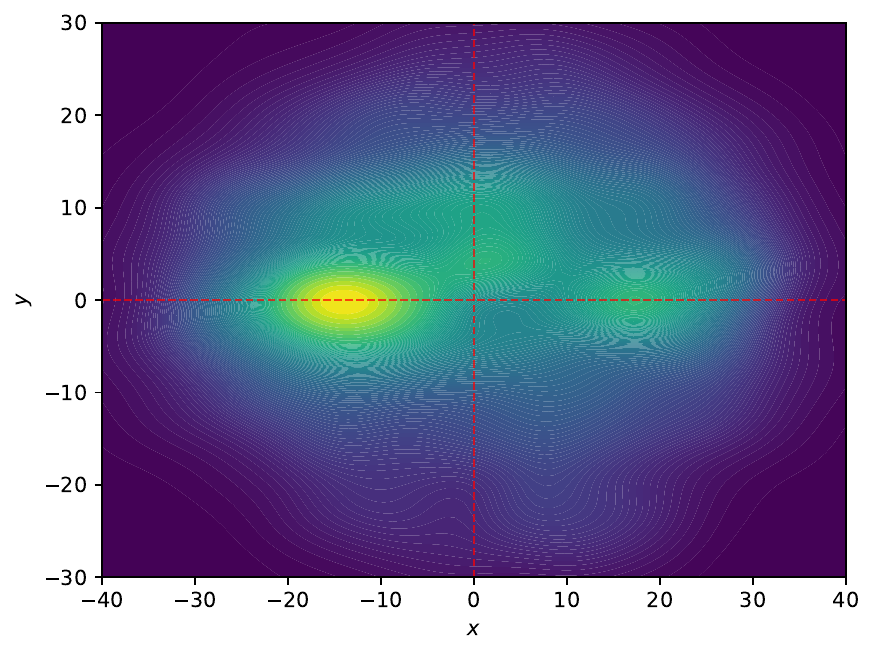}
        \caption{Heatmap of $x$-$y$ positions of different scenarios.}
        \label{fig:supp:stats:xy_heatmap}
    \end{subfigure}
    \caption{\textbf{Scenario statistics.} 
    Distribution of scenarios with agent involvement across camera views and on the $x$-$y$ plane.}
    \label{fig:supp:stats}
\end{figure*}

Since driving scenes are dynamic environments in which both the ego-vehicle and other traffic participants are in motion, a single camera view is often insufficient to capture all relevant interactions. As a result, most modern datasets~\cite{caesar_2020_nuscenes, sun_2020_waymo, wilson_2021_argoverse2} employ multi-camera systems with slightly overlapping fields of view. In our benchmark, we mine scenarios where agents are distributed across varying numbers of camera views, as illustrated in~Fig.\ref{fig:supp:stats:stsnu_views_per_scenario}. We observe that approximately half of the scenarios are confined to a single view, while the remaining scenarios span up to four views over the observed time period. The relatively high concentration of actions in a single view can be attributed to two main factors: 1) In the majority of sequences in \nus, the ego-vehicle either drives straight or remains stationary. 2) Many scenarios occur in the front or rear view of the ego-vehicle, as these cameras primarily capture the road it travels on. Each of these factors, individually and in combination, contributes to the skew in this distribution. Nevertheless, the remaining half of the scenarios span multiple views, underscoring the importance of evaluating a model’s ability to generalize across spatially distributed visual inputs.

To maintain a balanced dataset, we sub-sample the mined scenarios and remove samples from overrepresented categories. We define three criteria for agent-related scenarios to guide this selection: 1) occlusion, 2) distance to the ego-vehicle, and 3) spatial distribution. Based on these criteria, we retain scenarios that are highly visible, occur in the near surrounding of the ego-vehicle, and are spatially well-distributed around it. The first criterion is straightforward since visual systems cannot effectively reason about occluded agents. 
The second criterion prioritizes agents that are closer to the ego-vehicle, as distant objects are harder to perceive, and the benchmark is not focused on evaluating long-range detection or reasoning about small objects. The final criterion ensures a diverse set of samples with respect to camera viewpoints and spatial coverage, as illustrated in~Fig.\ref{fig:supp:stats:xy_heatmap}. We observe a distributional imbalance, with a disproportionate number of scenarios occurring in the rear area of the ego-vehicle, reflecting biases in the underlying data. However, the remaining scenarios are relatively well distributed across other spatial regions.

\newpage
\subsection{Scenario catalog}
The benchmark generation starts with the definition of a scenario catalog. It includes all scenarios that should be covered in the benchmark and assigns negative scenarios for each entry. Negative scenarios do not occur during the assigned scenario and serve later in the benchmark creation as wrong-choice options for the multiple-choice answer generation. 
In our scenario catalog, we list scenarios for ego-vehicle (Fig.~\ref{fig:supp:scenatio_catalog:ego}), other agents (Fig.~\ref{fig:supp:scenatio_catalog:agent}), interactions between ego-vehicle and agents (Fig.~\ref{fig:supp:scenatio_catalog:ego_agent}), and interactions among other agents (Fig.~\ref{fig:supp:scenatio_catalog:agent_agent}). 
For the respective categories, we define the scenarios in text form in Tables~\ref{tab:supp:scenatio_catalog:description:ego},~\ref{tab:supp:scenatio_catalog:description:agent},~\ref{tab:supp:scenatio_catalog:description:ego_agent} and,~\ref{tab:supp:scenatio_catalog:description:agent_agent}.

\newcommand{\FTcat}{}
\def\FTcat(#1,#2,#3){%
	\FTentry(#1,{{\color{black!40!white}\faBook}\hspace{0.2em}#3}, black)
	(tmp.west)++(0.8em,-0.4em)node(#2){}
	(tmp.west)++(1.5em,0)
	++(0,-1.3em) 
}

\newcommand{\FTdir}{}
\def\FTdir(#1,#2,#3){%
  \FTentry(#1,{{\color{green!60!black}\faCar}\hspace{0.2em}#3}, black)
  (tmp.west)++(0.8em,-0.4em)node(#2){}
  (tmp.west)++(2.5em,0)
  ++(0,-1.3em)
}
	
\newcommand{\FTcar}{}
\def\FTcar(#1,#2,#3){%
  \FTentry(#1,{{\color{C0}\faCar}\hspace{0.2em}#3}, black)
  (tmp.west)++(0.8em,-0.4em)node(#2){}
  (tmp.west)++(2.5em,0)
  ++(0,-1.3em)
}

\newcommand{\FTcarcar}{}
\def\FTcarcar(#1,#2,#3){%
  \FTentry(#1,{{\color{C0}\faCar\color{C1}\faCar}\hspace{0.2em}#3}, black)
  (tmp.west)++(0.8em,-0.4em)node(#2){}
  (tmp.west)++(2.5em,0)
  ++(0,-1.3em)
}

\newcommand{\FTcarperson}{}
\def\FTcarperson(#1,#2,#3){%
  \FTentry(#1,{{\color{C0}\faCar\color{C1}\faWalking}\hspace{0.2em}#3}, black)
  (tmp.west)++(0.8em,-0.4em)node(#2){}
  (tmp.west)++(2.5em,0)
  ++(0,-1.3em)
}

\newcommand{\FTperson}{}
\def\FTperson(#1,#2,#3){%
  \FTentry(#1,{{\color{C1}\faWalking}\hspace{0.2em}#3}, black)
  (tmp.west)++(0.8em,-0.4em)node(#2){}
  (tmp.west)++(2.5em,0)
  ++(0,-1.3em)
}

\newcommand{\FTagent}{}
\def\FTagent(#1,#2,#3){%
  \FTentry(#1,{{\color{C1}\faCar}\hspace{0.2em}#3}, black)
  (tmp.west)++(0.8em,-0.4em)node(#2){}
  (tmp.west)++(2.5em,0)
  ++(0,-1.3em)
}

\newcommand{\FTagentagent}{}
\def\FTagentagent(#1,#2,#3){%
  \FTentry(#1,{{\color{C1}\faCar\color{C1}\faCar}\hspace{0.2em}#3}, black)
  (tmp.west)++(0.8em,-0.4em)node(#2){}
  (tmp.west)++(2.5em,0)
  ++(0,-1.3em)
}

\newcommand{\FTpersonperson}{}
\def\FTpersonperson(#1,#2,#3){%
  \FTentry(#1,{{\color{C1}\faWalking\color{C1}\faWalking}\hspace{0.2em}#3}, black)
  (tmp.west)++(0.8em,-0.4em)node(#2){}
  (tmp.west)++(2.5em,0)
  ++(0,-1.3em)
}

\newcommand{\FTagentperson}{}
\def\FTagentperson(#1,#2,#3){%
  \FTentry(#1,{{\color{C1}\faCar\color{C1}\faWalking}\hspace{0.2em}#3}, black)
  (tmp.west)++(0.8em,-0.4em)node(#2){}
  (tmp.west)++(2.5em,0)
  ++(0,-1.3em)
}

\newcommand{\FTentry}{}
\def\FTentry(#1,#2,#3){%
  node(tmp){}
  (#1|-tmp)++(1.2em,0) node(tmp)[anchor=west,#3]{ \tiny\tt #2}
  (#1)|-(tmp.west)
  ++(0,-0.7em) %
}
\newcommand{\FTroot}{tmp.west}

\begin{figure*}[b!]
    \centering
    \begin{subfigure}[t]{0.5\textwidth}
        \centering
        \begin{tikzpicture}
\draw[color=black!60!white]
\FTcat(\FTroot,group_ego,ego){
    \FTcar(group_ego,ego_LEFT_TURN,left turn){
        \FTentry(ego_LEFT_TURN,stop, red)
        \FTentry(ego_LEFT_TURN,decelerate, red)
        \FTentry(ego_LEFT_TURN,right turn, red)
        \FTentry(ego_LEFT_TURN,u turn, red)
        \FTentry(ego_LEFT_TURN,lane change, red)
        \FTentry(ego_LEFT_TURN,reverse, red)
        \FTentry(ego_LEFT_TURN,accelerate, red)
    }
    \FTcar(group_ego,ego_ACCELERATE,accelerate){
        \FTentry(ego_ACCELERATE,stop, red)
        \FTentry(ego_ACCELERATE,decelerate, red)
        \FTentry(ego_ACCELERATE,left turn, red)
        \FTentry(ego_ACCELERATE,right turn, red)
        \FTentry(ego_ACCELERATE,u turn, red)
        \FTentry(ego_ACCELERATE,lane change, red)
        \FTentry(ego_ACCELERATE,reverse, red)
    }
    \FTcar(group_ego,ego_RIGHT_TURN,right turn){
        \FTentry(ego_RIGHT_TURN,stop, red)
        \FTentry(ego_RIGHT_TURN,decelerate, red)
        \FTentry(ego_RIGHT_TURN,left turn, red)
        \FTentry(ego_RIGHT_TURN,u turn, red)
        \FTentry(ego_RIGHT_TURN,reverse, red)
        \FTentry(ego_RIGHT_TURN,lane change, red)
        \FTentry(ego_RIGHT_TURN,accelerate, red)
    }
};
\end{tikzpicture}
    \end{subfigure}%
    ~ 
    \begin{subfigure}[t]{0.5\textwidth}
        \centering
        \begin{tikzpicture}
\draw[color=black!60!white]
\FTcat(\FTroot,group_ego,ego){
    \FTcar(group_ego,ego_LANE_CHANGE,lane change){
        \FTentry(ego_LANE_CHANGE,stop, red)
        \FTentry(ego_LANE_CHANGE,decelerate, red)
        \FTentry(ego_LANE_CHANGE,left turn, red)
        \FTentry(ego_LANE_CHANGE,right turn, red)
        \FTentry(ego_LANE_CHANGE,u turn, red)
        \FTentry(ego_LANE_CHANGE,reverse, red)
        \FTentry(ego_LANE_CHANGE,accelerate, red)
    }
    \FTcar(group_ego,ego_DECELERATE,decelerate){
        \FTentry(ego_DECELERATE,stop, red)
        \FTentry(ego_DECELERATE,left turn, red)
        \FTentry(ego_DECELERATE,right turn, red)
        \FTentry(ego_DECELERATE,u turn, red)
        \FTentry(ego_DECELERATE,lane change, red)
        \FTentry(ego_DECELERATE,reverse, red)
        \FTentry(ego_DECELERATE,accelerate, red)
    }
    \FTcar(group_ego,ego_STOP,stop){
        \FTentry(ego_STOP,decelerate, red)
        \FTentry(ego_STOP,left turn, red)
        \FTentry(ego_STOP,right turn, red)
        \FTentry(ego_STOP,u turn, red)
        \FTentry(ego_STOP,lane change, red)
        \FTentry(ego_STOP,reverse, red)
        \FTentry(ego_STOP,accelerate, red)
    }
};
\end{tikzpicture}
    \end{subfigure}
    \caption{\textbf{Scenario catalog.} 
    Ego scenarios with assigned negative scenarios in red.}
    \label{fig:supp:scenatio_catalog:ego}
\end{figure*}

\begin{figure*}[t!]
    \centering
    \begin{subfigure}[t]{0.5\textwidth}
        \centering
        \begin{tikzpicture}
\draw[color=black!60!white]
\FTcat(\FTroot,group_ego-agent,ego-agent){
    \FTcarcar(group_ego-agent,ego_LEAD_AGENT,lead agent){
        \FTentry(ego_LEAD_AGENT,stationary behind agent, red)
        \FTentry(ego_LEAD_AGENT,pass agent, red)
        \FTentry(ego_LEAD_AGENT,stationary right of agent, red)
        \FTentry(ego_LEAD_AGENT,stationary in front of agent, red)
        \FTentry(ego_LEAD_AGENT,stop, red)
        \FTentry(ego_LEAD_AGENT,right turn, red)
        \FTentry(ego_LEAD_AGENT,accelerate, red)
        \FTentry(ego_LEAD_AGENT,moving right of agent, red)
        \FTentry(ego_LEAD_AGENT,u turn, red)
        \FTentry(ego_LEAD_AGENT,follow agent, red)
        \FTentry(ego_LEAD_AGENT,stationary left of agent, red)
        \FTentry(ego_LEAD_AGENT,moving left of agent, red)
        \FTentry(ego_LEAD_AGENT,lane change, red)
        \FTentry(ego_LEAD_AGENT,overtake agent, red)
        \FTentry(ego_LEAD_AGENT,reverse, red)
        \FTentry(ego_LEAD_AGENT,wait ped cross, red)
        \FTentry(ego_LEAD_AGENT,decelerate, red)
        \FTentry(ego_LEAD_AGENT,left turn, red)
}
    \FTcarcar(group_ego-agent,ego_PASS_AGENT,pass agent){
        \FTentry(ego_PASS_AGENT,stationary behind agent, red)
        \FTentry(ego_PASS_AGENT,stationary right of agent, red)
        \FTentry(ego_PASS_AGENT,stationary in front of agent, red)
        \FTentry(ego_PASS_AGENT,stop, red)
        \FTentry(ego_PASS_AGENT,right turn, red)
        \FTentry(ego_PASS_AGENT,left turn, red)
        \FTentry(ego_PASS_AGENT,accelerate, red)
        \FTentry(ego_PASS_AGENT,moving right of agent, red)
        \FTentry(ego_PASS_AGENT,u turn, red)
        \FTentry(ego_PASS_AGENT,follow agent, red)
        \FTentry(ego_PASS_AGENT,stationary left of agent, red)
        \FTentry(ego_PASS_AGENT,moving left of agent, red)
        \FTentry(ego_PASS_AGENT,lane change, red)
        \FTentry(ego_PASS_AGENT,overtake agent, red)
        \FTentry(ego_PASS_AGENT,reverse, red)
        \FTentry(ego_PASS_AGENT,wait ped cross, red)
        \FTentry(ego_PASS_AGENT,lead agent, red)
        \FTentry(ego_PASS_AGENT,decelerate, red)
}
    \FTcarperson(group_ego-agent,ego_WAIT_PED_CROSS,wait ped cross){
        \FTentry(ego_WAIT_PED_CROSS,pass agent, red)
        \FTentry(ego_WAIT_PED_CROSS,follow ego, red)
        \FTentry(ego_WAIT_PED_CROSS,decelerate, red)
        \FTentry(ego_WAIT_PED_CROSS,reverse, red)
        \FTentry(ego_WAIT_PED_CROSS,u turn, red)
        \FTentry(ego_WAIT_PED_CROSS,overtake ego, red)
        \FTentry(ego_WAIT_PED_CROSS,follow agent, red)
        \FTentry(ego_WAIT_PED_CROSS,lead ego, red)
        \FTentry(ego_WAIT_PED_CROSS,overtake agent, red)
        \FTentry(ego_WAIT_PED_CROSS,right turn, red)
        \FTentry(ego_WAIT_PED_CROSS,lead agent, red)
        \FTentry(ego_WAIT_PED_CROSS,left turn, red)
        \FTentry(ego_WAIT_PED_CROSS,stop, red)
}
    \FTcarcar(group_ego-agent,ego_FOLLOW_AGENT,follow agent){
        \FTentry(ego_FOLLOW_AGENT,stationary behind agent, red)
        \FTentry(ego_FOLLOW_AGENT,pass agent, red)
        \FTentry(ego_FOLLOW_AGENT,stationary right of agent, red)
        \FTentry(ego_FOLLOW_AGENT,overtake agent, red)
        \FTentry(ego_FOLLOW_AGENT,stationary in front of agent, red)
        \FTentry(ego_FOLLOW_AGENT,stop, red)
        \FTentry(ego_FOLLOW_AGENT,left turn, red)
        \FTentry(ego_FOLLOW_AGENT,reverse, red)
        \FTentry(ego_FOLLOW_AGENT,accelerate, red)
        \FTentry(ego_FOLLOW_AGENT,moving right of agent, red)
        \FTentry(ego_FOLLOW_AGENT,u turn, red)
        \FTentry(ego_FOLLOW_AGENT,moving left of agent, red)
        \FTentry(ego_FOLLOW_AGENT,lane change, red)
        \FTentry(ego_FOLLOW_AGENT,stationary left of agent, red)
        \FTentry(ego_FOLLOW_AGENT,right turn, red)
        \FTentry(ego_FOLLOW_AGENT,wait ped cross, red)
        \FTentry(ego_FOLLOW_AGENT,lead agent, red)
        \FTentry(ego_FOLLOW_AGENT,decelerate, red)
    }
};
\end{tikzpicture}
    \end{subfigure}%
    ~ 
    \begin{subfigure}[t]{0.5\textwidth}
        \centering
        \begin{tikzpicture}
\draw[color=black!60!white]
\FTcat(\FTroot,group_ego-agent,ego-agent){
    \FTcarcar(group_ego-agent,ego_OVERTAKE_AGENT,overtake agent){
        \FTentry(ego_OVERTAKE_AGENT,pass agent, red)
        \FTentry(ego_OVERTAKE_AGENT,moving left of agent, red)
        \FTentry(ego_OVERTAKE_AGENT,moving right of agent, red)
        \FTentry(ego_OVERTAKE_AGENT,stationary left of agent, red)
        \FTentry(ego_OVERTAKE_AGENT,stationary right of agent, red)
        \FTentry(ego_OVERTAKE_AGENT,lead agent, red)
        \FTentry(ego_OVERTAKE_AGENT,follow agent, red)
        \FTentry(ego_OVERTAKE_AGENT,stationary in front of agent, red)
        \FTentry(ego_OVERTAKE_AGENT,stationary behind agent, red)
        \FTentry(ego_OVERTAKE_AGENT,wait ped cross, red)
        \FTentry(ego_OVERTAKE_AGENT,accelerate, red)
        \FTentry(ego_OVERTAKE_AGENT,decelerate, red)
        \FTentry(ego_OVERTAKE_AGENT,stop, red)
        \FTentry(ego_OVERTAKE_AGENT,lane change, red)
        \FTentry(ego_OVERTAKE_AGENT,u turn, red)
        \FTentry(ego_OVERTAKE_AGENT,reverse, red)
        \FTentry(ego_OVERTAKE_AGENT,left turn, red)
        \FTentry(ego_OVERTAKE_AGENT,right turn, red)
}
    \FTcarcar(group_ego-agent,ego_STATIONARY_LEFT_OF_AGENT,stationary left of agent){
        \FTentry(ego_STATIONARY_LEFT_OF_AGENT,pass agent, red)
        \FTentry(ego_STATIONARY_LEFT_OF_AGENT,overtake agent, red)
        \FTentry(ego_STATIONARY_LEFT_OF_AGENT,moving left of agent, red)
        \FTentry(ego_STATIONARY_LEFT_OF_AGENT,moving right of agent, red)
        \FTentry(ego_STATIONARY_LEFT_OF_AGENT,stationary right of agent, red)
        \FTentry(ego_STATIONARY_LEFT_OF_AGENT,lead agent, red)
        \FTentry(ego_STATIONARY_LEFT_OF_AGENT,follow agent, red)
        \FTentry(ego_STATIONARY_LEFT_OF_AGENT,stationary in front of agent, red)
        \FTentry(ego_STATIONARY_LEFT_OF_AGENT,stationary behind agent, red)
        \FTentry(ego_STATIONARY_LEFT_OF_AGENT,wait ped cross, red)
        \FTentry(ego_STATIONARY_LEFT_OF_AGENT,accelerate, red)
        \FTentry(ego_STATIONARY_LEFT_OF_AGENT,decelerate, red)
        \FTentry(ego_STATIONARY_LEFT_OF_AGENT,stop, red)
        \FTentry(ego_STATIONARY_LEFT_OF_AGENT,lane change, red)
        \FTentry(ego_STATIONARY_LEFT_OF_AGENT,u turn, red)
        \FTentry(ego_STATIONARY_LEFT_OF_AGENT,reverse, red)
        \FTentry(ego_STATIONARY_LEFT_OF_AGENT,left turn, red)
        \FTentry(ego_STATIONARY_LEFT_OF_AGENT,right turn, red)
}
    \FTcarcar(group_ego-agent,ego_STATIONARY_IN_FRONT_OF_AGENT,stationary in front of agent){
        \FTentry(ego_STATIONARY_IN_FRONT_OF_AGENT,lead agent, red)
        \FTentry(ego_STATIONARY_IN_FRONT_OF_AGENT,follow agent, red)
        \FTentry(ego_STATIONARY_IN_FRONT_OF_AGENT,stationary behind agent, red)
        \FTentry(ego_STATIONARY_IN_FRONT_OF_AGENT,pass agent, red)
        \FTentry(ego_STATIONARY_IN_FRONT_OF_AGENT,overtake agent, red)
        \FTentry(ego_STATIONARY_IN_FRONT_OF_AGENT,moving left of agent, red)
        \FTentry(ego_STATIONARY_IN_FRONT_OF_AGENT,moving right of agent, red)
        \FTentry(ego_STATIONARY_IN_FRONT_OF_AGENT,stationary left of agent, red)
        \FTentry(ego_STATIONARY_IN_FRONT_OF_AGENT,stationary right of agent, red)
        \FTentry(ego_STATIONARY_IN_FRONT_OF_AGENT,wait ped cross, red)
        \FTentry(ego_STATIONARY_IN_FRONT_OF_AGENT,accelerate, red)
        \FTentry(ego_STATIONARY_IN_FRONT_OF_AGENT,decelerate, red)
        \FTentry(ego_STATIONARY_IN_FRONT_OF_AGENT,stop, red)
        \FTentry(ego_STATIONARY_IN_FRONT_OF_AGENT,lane change, red)
        \FTentry(ego_STATIONARY_IN_FRONT_OF_AGENT,u turn, red)
        \FTentry(ego_STATIONARY_IN_FRONT_OF_AGENT,reverse, red)
        \FTentry(ego_STATIONARY_IN_FRONT_OF_AGENT,left turn, red)
        \FTentry(ego_STATIONARY_IN_FRONT_OF_AGENT,right turn, red)
    }
};
\end{tikzpicture}
    \end{subfigure}
    \caption{\textbf{Scenario catalog.} 
    Ego-Agent scenarios with assigned negative scenarios in red.}
    \label{fig:supp:scenatio_catalog:ego_agent}
\end{figure*}

\begin{figure*}[t!]
    \centering
    \begin{subfigure}[t]{0.32\textwidth}
        \centering
        \begin{tikzpicture}
\draw[color=black!60!white]
\FTcat(\FTroot,group_agent,agent){
    \FTagent(group_agent,ego_RIGHT_TURN,right turn){
        \FTentry(ego_RIGHT_TURN,stop, red)
        \FTentry(ego_RIGHT_TURN,decelerate, red)
        \FTentry(ego_RIGHT_TURN,left turn, red)
        \FTentry(ego_RIGHT_TURN,u turn, red)
        \FTentry(ego_RIGHT_TURN,lane change, red)
        \FTentry(ego_RIGHT_TURN,reverse, red)
        \FTentry(ego_RIGHT_TURN,accelerate, red)
}
    \FTagent(group_agent,ego_LANE_CHANGE,lane change){
        \FTentry(ego_LANE_CHANGE,stop, red)
        \FTentry(ego_LANE_CHANGE,decelerate, red)
        \FTentry(ego_LANE_CHANGE,left turn, red)
        \FTentry(ego_LANE_CHANGE,right turn, red)
        \FTentry(ego_LANE_CHANGE,u turn, red)
        \FTentry(ego_LANE_CHANGE,reverse, red)
        \FTentry(ego_LANE_CHANGE,accelerate, red)
}
    \FTagent(group_agent,ego_OVERTAKE_EGO,overtake ego){
        \FTentry(ego_OVERTAKE_EGO,stationary right of ego, red)
        \FTentry(ego_OVERTAKE_EGO,moving left of ego, red)
        \FTentry(ego_OVERTAKE_EGO,follow ego, red)
        \FTentry(ego_OVERTAKE_EGO,stationary in front of ego, red)
        \FTentry(ego_OVERTAKE_EGO,stop, red)
        \FTentry(ego_OVERTAKE_EGO,decelerate, red)
        \FTentry(ego_OVERTAKE_EGO,left turn, red)
        \FTentry(ego_OVERTAKE_EGO,right turn, red)
        \FTentry(ego_OVERTAKE_EGO,moving right of ego, red)
        \FTentry(ego_OVERTAKE_EGO,pass ego, red)
        \FTentry(ego_OVERTAKE_EGO,u turn, red)
        \FTentry(ego_OVERTAKE_EGO,stationary left of ego, red)
        \FTentry(ego_OVERTAKE_EGO,lead ego, red)
        \FTentry(ego_OVERTAKE_EGO,lane change, red)
        \FTentry(ego_OVERTAKE_EGO,stationary behind ego, red)
        \FTentry(ego_OVERTAKE_EGO,reverse, red)
        \FTentry(ego_OVERTAKE_EGO,accelerate, red)
}
    \FTagent(group_agent,ego_FOLLOW_EGO,follow ego){
        \FTentry(ego_FOLLOW_EGO,stationary behind ego, red)
        \FTentry(ego_FOLLOW_EGO,lead ego, red)
        \FTentry(ego_FOLLOW_EGO,stationary in front of ego, red)
        \FTentry(ego_FOLLOW_EGO,pass ego, red)
        \FTentry(ego_FOLLOW_EGO,overtake ego, red)
        \FTentry(ego_FOLLOW_EGO,stationary left of ego, red)
        \FTentry(ego_FOLLOW_EGO,stationary right of ego, red)
        \FTentry(ego_FOLLOW_EGO,moving left of ego, red)
        \FTentry(ego_FOLLOW_EGO,moving right of ego, red)
        \FTentry(ego_FOLLOW_EGO,accelerate, red)
        \FTentry(ego_FOLLOW_EGO,lane change, red)
        \FTentry(ego_FOLLOW_EGO,left turn, red)
        \FTentry(ego_FOLLOW_EGO,right turn, red)
        \FTentry(ego_FOLLOW_EGO,u turn, red)
        \FTentry(ego_FOLLOW_EGO,reverse, red)
        \FTentry(ego_FOLLOW_EGO,stop, red)
        \FTentry(ego_FOLLOW_EGO,decelerate, red)
}
    \FTagent(group_agent,ego_STOP,stop){
        \FTentry(ego_STOP,decelerate, red)
        \FTentry(ego_STOP,left turn, red)
        \FTentry(ego_STOP,right turn, red)
        \FTentry(ego_STOP,u turn, red)
        \FTentry(ego_STOP,lane change, red)
        \FTentry(ego_STOP,reverse, red)
        \FTentry(ego_STOP,accelerate, red)
}
    \FTperson(group_agent,ego_RUN,run){
        \FTentry(ego_RUN,stop, red)
        \FTentry(ego_RUN,decelerate, red)
        \FTentry(ego_RUN,cross, red)
        \FTentry(ego_RUN,accelerate, red)
        \FTentry(ego_RUN,jaywalk, red)
        \FTentry(ego_RUN,stand, red)
        \FTentry(ego_RUN,walk, red)
}
};
\end{tikzpicture}
    \end{subfigure}%
    ~ 
    \begin{subfigure}[t]{0.32\textwidth}
        \centering
        \begin{tikzpicture}
\draw[color=black!60!white]
\FTcat(\FTroot,group_agent,agent){
    \FTagent(group_agent,ego_LEAD_EGO,lead ego){
        \FTentry(ego_LEAD_EGO,stationary right of ego, red)
        \FTentry(ego_LEAD_EGO,moving left of ego, red)
        \FTentry(ego_LEAD_EGO,follow ego, red)
        \FTentry(ego_LEAD_EGO,stationary in front of ego, red)
        \FTentry(ego_LEAD_EGO,stop, red)
        \FTentry(ego_LEAD_EGO,left turn, red)
        \FTentry(ego_LEAD_EGO,reverse, red)
        \FTentry(ego_LEAD_EGO,moving right of ego, red)
        \FTentry(ego_LEAD_EGO,pass ego, red)
        \FTentry(ego_LEAD_EGO,accelerate, red)
        \FTentry(ego_LEAD_EGO,u turn, red)
        \FTentry(ego_LEAD_EGO,overtake ego, red)
        \FTentry(ego_LEAD_EGO,stationary left of ego, red)
        \FTentry(ego_LEAD_EGO,lane change, red)
        \FTentry(ego_LEAD_EGO,stationary behind ego, red)
        \FTentry(ego_LEAD_EGO,right turn, red)
        \FTentry(ego_LEAD_EGO,decelerate, red)
}
    \FTagent(group_agent,ego_REVERSE,reverse){
        \FTentry(ego_REVERSE,stop, red)
        \FTentry(ego_REVERSE,decelerate, red)
        \FTentry(ego_REVERSE,left turn, red)
        \FTentry(ego_REVERSE,accelerate, red)
        \FTentry(ego_REVERSE,u turn, red)
        \FTentry(ego_REVERSE,right turn, red)
}
    \FTagent(group_agent,ego_U_TURN,u turn){
        \FTentry(ego_U_TURN,decelerate, red)
        \FTentry(ego_U_TURN,lane change, red)
        \FTentry(ego_U_TURN,left turn, red)
        \FTentry(ego_U_TURN,right turn, red)
        \FTentry(ego_U_TURN,stop, red)
        \FTentry(ego_U_TURN,reverse, red)
}
    \FTagent(group_agent,ego_STATIONARY_RIGHT_OF_EGO,stationary right of ego){
        \FTentry(ego_STATIONARY_RIGHT_OF_EGO,overtake ego, red)
        \FTentry(ego_STATIONARY_RIGHT_OF_EGO,pass ego, red)
        \FTentry(ego_STATIONARY_RIGHT_OF_EGO,moving right of ego, red)
        \FTentry(ego_STATIONARY_RIGHT_OF_EGO,moving left of ego, red)
        \FTentry(ego_STATIONARY_RIGHT_OF_EGO,stationary left of ego, red)
        \FTentry(ego_STATIONARY_RIGHT_OF_EGO,follow ego, red)
        \FTentry(ego_STATIONARY_RIGHT_OF_EGO,lead ego, red)
        \FTentry(ego_STATIONARY_RIGHT_OF_EGO,stationary in front of ego, red)
        \FTentry(ego_STATIONARY_RIGHT_OF_EGO,stationary behind ego, red)
        \FTentry(ego_STATIONARY_RIGHT_OF_EGO,accelerate, red)
        \FTentry(ego_STATIONARY_RIGHT_OF_EGO,decelerate, red)
        \FTentry(ego_STATIONARY_RIGHT_OF_EGO,lane change, red)
        \FTentry(ego_STATIONARY_RIGHT_OF_EGO,left turn, red)
        \FTentry(ego_STATIONARY_RIGHT_OF_EGO,right turn, red)
        \FTentry(ego_STATIONARY_RIGHT_OF_EGO,u turn, red)
        \FTentry(ego_STATIONARY_RIGHT_OF_EGO,reverse, red)
        \FTentry(ego_STATIONARY_RIGHT_OF_EGO,stop, red)
}
    \FTagent(group_agent,ego_STATIONARY_BEHIND_EGO,stationary behind ego){
        \FTentry(ego_STATIONARY_BEHIND_EGO,follow ego, red)
        \FTentry(ego_STATIONARY_BEHIND_EGO,lead ego, red)
        \FTentry(ego_STATIONARY_BEHIND_EGO,stationary in front of ego, red)
        \FTentry(ego_STATIONARY_BEHIND_EGO,pass ego, red)
        \FTentry(ego_STATIONARY_BEHIND_EGO,overtake ego, red)
        \FTentry(ego_STATIONARY_BEHIND_EGO,stationary left of ego, red)
        \FTentry(ego_STATIONARY_BEHIND_EGO,stationary right of ego, red)
        \FTentry(ego_STATIONARY_BEHIND_EGO,moving left of ego, red)
        \FTentry(ego_STATIONARY_BEHIND_EGO,moving right of ego, red)
        \FTentry(ego_STATIONARY_BEHIND_EGO,accelerate, red)
        \FTentry(ego_STATIONARY_BEHIND_EGO,decelerate, red)
        \FTentry(ego_STATIONARY_BEHIND_EGO,lane change, red)
        \FTentry(ego_STATIONARY_BEHIND_EGO,left turn, red)
        \FTentry(ego_STATIONARY_BEHIND_EGO,right turn, red)
        \FTentry(ego_STATIONARY_BEHIND_EGO,u turn, red)
        \FTentry(ego_STATIONARY_BEHIND_EGO,reverse, red)
        \FTentry(ego_STATIONARY_BEHIND_EGO,stop, red)
    }
};
\end{tikzpicture}
    \end{subfigure}
    ~
    \begin{subfigure}[t]{0.32\textwidth}
        \centering
        \begin{tikzpicture}
\draw[color=black!60!white]
\FTcat(\FTroot,group_agent,agent){
    \FTperson(group_agent,ego_JAYWALK,jaywalk){
        \FTentry(ego_JAYWALK,cross, red)
        \FTentry(ego_JAYWALK,run, red)
        \FTentry(ego_JAYWALK,stand, red)
        \FTentry(ego_JAYWALK,accelerate, red)
        \FTentry(ego_JAYWALK,decelerate, red)
        \FTentry(ego_JAYWALK,stop, red)
}
    \FTagent(group_agent,ego_LEFT_TURN,left turn){
        \FTentry(ego_LEFT_TURN,decelerate, red)
        \FTentry(ego_LEFT_TURN,lane change, red)
        \FTentry(ego_LEFT_TURN,right turn, red)
        \FTentry(ego_LEFT_TURN,u turn, red)
        \FTentry(ego_LEFT_TURN,stop, red)
        \FTentry(ego_LEFT_TURN,reverse, red)
        \FTentry(ego_LEFT_TURN,accelerate, red)
}
    \FTperson(group_agent,ego_WALK,walk){
        \FTentry(ego_WALK,cross, red)
        \FTentry(ego_WALK,jaywalk, red)
        \FTentry(ego_WALK,stand, red)
        \FTentry(ego_WALK,run, red)
        \FTentry(ego_WALK,accelerate, red)
        \FTentry(ego_WALK,decelerate, red)
        \FTentry(ego_WALK,stop, red)
}
    \FTperson(group_agent,ego_STAND,stand){
        \FTentry(ego_STAND,cross, red)
        \FTentry(ego_STAND,jaywalk, red)
        \FTentry(ego_STAND,walk, red)
        \FTentry(ego_STAND,run, red)
        \FTentry(ego_STAND,accelerate, red)
        \FTentry(ego_STAND,decelerate, red)
        \FTentry(ego_STAND,stop, red)
}
    \FTperson(group_agent,ego_CROSS,cross){
        \FTentry(ego_CROSS,jaywalk, red)
        \FTentry(ego_CROSS,run, red)
        \FTentry(ego_CROSS,stand, red)
        \FTentry(ego_CROSS,accelerate, red)
        \FTentry(ego_CROSS,decelerate, red)
        \FTentry(ego_CROSS,stop, red)
}
    \FTagent(group_agent,ego_ACCELERATE,accelerate){
            \FTentry(ego_ACCELERATE,lane change, red)
            \FTentry(ego_ACCELERATE,left turn, red)
            \FTentry(ego_ACCELERATE,right turn, red)
            \FTentry(ego_ACCELERATE,u turn, red)
            \FTentry(ego_ACCELERATE,decelerate, red)
            \FTentry(ego_ACCELERATE,reverse, red)
            \FTentry(ego_ACCELERATE,stop, red)
    }
};
\end{tikzpicture}
    \end{subfigure}
    \caption{\textbf{Scenario catalog.} 
    Agent scenarios with assigned negative scenarios in red.}
    \label{fig:supp:scenatio_catalog:agent}
\end{figure*}

\begin{figure*}[t!]
    \centering
    \begin{subfigure}[t]{0.32\textwidth}
        \centering
        \begin{tikzpicture}
\draw[color=black!60!white]
\FTcat(\FTroot,group_agent-agent,agent-agent){
    \FTagentagent(group_agent-agent,ego_FOLLOW_AGENT,follow agent){
        \FTentry(ego_FOLLOW_AGENT,stationary behind agent, red)
        \FTentry(ego_FOLLOW_AGENT,lead agent, red)
        \FTentry(ego_FOLLOW_AGENT,stationary in front of agent, red)
        \FTentry(ego_FOLLOW_AGENT,pass agent, red)
        \FTentry(ego_FOLLOW_AGENT,overtake agent, red)
        \FTentry(ego_FOLLOW_AGENT,moving left of agent, red)
        \FTentry(ego_FOLLOW_AGENT,moving right of agent, red)
        \FTentry(ego_FOLLOW_AGENT,stationary left of agent, red)
        \FTentry(ego_FOLLOW_AGENT,stationary right of agent, red)
        \FTentry(ego_FOLLOW_AGENT,wait ped cross, red)
        \FTentry(ego_FOLLOW_AGENT,accelerate, red)
        \FTentry(ego_FOLLOW_AGENT,decelerate, red)
        \FTentry(ego_FOLLOW_AGENT,stop, red)
        \FTentry(ego_FOLLOW_AGENT,lane change, red)
        \FTentry(ego_FOLLOW_AGENT,u turn, red)
        \FTentry(ego_FOLLOW_AGENT,reverse, red)
        \FTentry(ego_FOLLOW_AGENT,left turn, red)
        \FTentry(ego_FOLLOW_AGENT,right turn, red)
}
    \FTagentagent(group_agent-agent,ego_STATIONARY_LEFT_OF_AGENT,stationary left of agent){
        \FTentry(ego_STATIONARY_LEFT_OF_AGENT,pass agent, red)
        \FTentry(ego_STATIONARY_LEFT_OF_AGENT,overtake agent, red)
        \FTentry(ego_STATIONARY_LEFT_OF_AGENT,moving left of agent, red)
        \FTentry(ego_STATIONARY_LEFT_OF_AGENT,moving right of agent, red)
        \FTentry(ego_STATIONARY_LEFT_OF_AGENT,stationary right of agent, red)
        \FTentry(ego_STATIONARY_LEFT_OF_AGENT,lead agent, red)
        \FTentry(ego_STATIONARY_LEFT_OF_AGENT,follow agent, red)
        \FTentry(ego_STATIONARY_LEFT_OF_AGENT,stationary in front of agent, red)
        \FTentry(ego_STATIONARY_LEFT_OF_AGENT,stationary behind agent, red)
        \FTentry(ego_STATIONARY_LEFT_OF_AGENT,wait ped cross, red)
        \FTentry(ego_STATIONARY_LEFT_OF_AGENT,accelerate, red)
        \FTentry(ego_STATIONARY_LEFT_OF_AGENT,decelerate, red)
        \FTentry(ego_STATIONARY_LEFT_OF_AGENT,stop, red)
        \FTentry(ego_STATIONARY_LEFT_OF_AGENT,lane change, red)
        \FTentry(ego_STATIONARY_LEFT_OF_AGENT,u turn, red)
        \FTentry(ego_STATIONARY_LEFT_OF_AGENT,reverse, red)
        \FTentry(ego_STATIONARY_LEFT_OF_AGENT,left turn, red)
        \FTentry(ego_STATIONARY_LEFT_OF_AGENT,right turn, red)
}
    \FTagentagent(group_agent-agent,ego_STATIONARY_RIGHT_OF_AGENT,stationary right of agent){
        \FTentry(ego_STATIONARY_RIGHT_OF_AGENT,pass agent, red)
        \FTentry(ego_STATIONARY_RIGHT_OF_AGENT,overtake agent, red)
        \FTentry(ego_STATIONARY_RIGHT_OF_AGENT,moving right of agent, red)
        \FTentry(ego_STATIONARY_RIGHT_OF_AGENT,moving left of agent, red)
        \FTentry(ego_STATIONARY_RIGHT_OF_AGENT,stationary left of agent, red)
        \FTentry(ego_STATIONARY_RIGHT_OF_AGENT,lead agent, red)
        \FTentry(ego_STATIONARY_RIGHT_OF_AGENT,follow agent, red)
        \FTentry(ego_STATIONARY_RIGHT_OF_AGENT,stationary in front of agent, red)
        \FTentry(ego_STATIONARY_RIGHT_OF_AGENT,stationary behind agent, red)
        \FTentry(ego_STATIONARY_RIGHT_OF_AGENT,wait ped cross, red)
        \FTentry(ego_STATIONARY_RIGHT_OF_AGENT,accelerate, red)
        \FTentry(ego_STATIONARY_RIGHT_OF_AGENT,decelerate, red)
        \FTentry(ego_STATIONARY_RIGHT_OF_AGENT,stop, red)
        \FTentry(ego_STATIONARY_RIGHT_OF_AGENT,lane change, red)
        \FTentry(ego_STATIONARY_RIGHT_OF_AGENT,u turn, red)
        \FTentry(ego_STATIONARY_RIGHT_OF_AGENT,reverse, red)
        \FTentry(ego_STATIONARY_RIGHT_OF_AGENT,left turn, red)
        \FTentry(ego_STATIONARY_RIGHT_OF_AGENT,right turn, red)
}
    \FTpersonperson(group_agent-agent,ego_WALK_ALONGSIDE,walk alongside){
        \FTentry(ego_WALK_ALONGSIDE,walk opposite, red)
        \FTentry(ego_WALK_ALONGSIDE,overtake agent, red)
        \FTentry(ego_WALK_ALONGSIDE,pass agent, red)
        \FTentry(ego_WALK_ALONGSIDE,run, red)
        \FTentry(ego_WALK_ALONGSIDE,stand, red)
        \FTentry(ego_WALK_ALONGSIDE,cross, red)
        \FTentry(ego_WALK_ALONGSIDE,jaywalk, red)
}
    \FTagentagent(group_agent-agent,ego_STATIONARY_IN_FRONT_OF_AGENT,stationary in front of agent){
        \FTentry(ego_STATIONARY_IN_FRONT_OF_AGENT,lead agent, red)
        \FTentry(ego_STATIONARY_IN_FRONT_OF_AGENT,follow agent, red)
        \FTentry(ego_STATIONARY_IN_FRONT_OF_AGENT,stationary behind agent, red)
        \FTentry(ego_STATIONARY_IN_FRONT_OF_AGENT,pass agent, red)
        \FTentry(ego_STATIONARY_IN_FRONT_OF_AGENT,overtake agent, red)
        \FTentry(ego_STATIONARY_IN_FRONT_OF_AGENT,moving left of agent, red)
        \FTentry(ego_STATIONARY_IN_FRONT_OF_AGENT,moving right of agent, red)
        \FTentry(ego_STATIONARY_IN_FRONT_OF_AGENT,stationary left of agent, red)
        \FTentry(ego_STATIONARY_IN_FRONT_OF_AGENT,stationary right of agent, red)
        \FTentry(ego_STATIONARY_IN_FRONT_OF_AGENT,wait ped cross, red)
        \FTentry(ego_STATIONARY_IN_FRONT_OF_AGENT,accelerate, red)
        \FTentry(ego_STATIONARY_IN_FRONT_OF_AGENT,decelerate, red)
        \FTentry(ego_STATIONARY_IN_FRONT_OF_AGENT,stop, red)
        \FTentry(ego_STATIONARY_IN_FRONT_OF_AGENT,lane change, red)
        \FTentry(ego_STATIONARY_IN_FRONT_OF_AGENT,u turn, red)
        \FTentry(ego_STATIONARY_IN_FRONT_OF_AGENT,reverse, red)
        \FTentry(ego_STATIONARY_IN_FRONT_OF_AGENT,left turn, red)
        \FTentry(ego_STATIONARY_IN_FRONT_OF_AGENT,right turn, red)
}
};
\end{tikzpicture}
    \end{subfigure}%
    ~ 
    \begin{subfigure}[t]{0.32\textwidth}
        \centering
        \begin{tikzpicture}
\draw[color=black!60!white]
\FTcat(\FTroot,group_agent-agent,agent-agent){
    \FTagentagent(group_agent-agent,ego_STATIONARY_BEHIND_AGENT,stationary behind agent){
        \FTentry(ego_STATIONARY_BEHIND_AGENT,follow agent, red)
        \FTentry(ego_STATIONARY_BEHIND_AGENT,lead agent, red)
        \FTentry(ego_STATIONARY_BEHIND_AGENT,stationary in front of agent, red)
        \FTentry(ego_STATIONARY_BEHIND_AGENT,pass agent, red)
        \FTentry(ego_STATIONARY_BEHIND_AGENT,overtake agent, red)
        \FTentry(ego_STATIONARY_BEHIND_AGENT,moving left of agent, red)
        \FTentry(ego_STATIONARY_BEHIND_AGENT,moving right of agent, red)
        \FTentry(ego_STATIONARY_BEHIND_AGENT,stationary left of agent, red)
        \FTentry(ego_STATIONARY_BEHIND_AGENT,stationary right of agent, red)
        \FTentry(ego_STATIONARY_BEHIND_AGENT,wait ped cross, red)
        \FTentry(ego_STATIONARY_BEHIND_AGENT,accelerate, red)
        \FTentry(ego_STATIONARY_BEHIND_AGENT,decelerate, red)
        \FTentry(ego_STATIONARY_BEHIND_AGENT,stop, red)
        \FTentry(ego_STATIONARY_BEHIND_AGENT,lane change, red)
        \FTentry(ego_STATIONARY_BEHIND_AGENT,u turn, red)
        \FTentry(ego_STATIONARY_BEHIND_AGENT,reverse, red)
        \FTentry(ego_STATIONARY_BEHIND_AGENT,left turn, red)
        \FTentry(ego_STATIONARY_BEHIND_AGENT,right turn, red)
}
    \FTagentagent(group_agent-agent,ego_PASS_AGENT,pass agent){
        \FTentry(ego_PASS_AGENT,overtake agent, red)
        \FTentry(ego_PASS_AGENT,moving left of agent, red)
        \FTentry(ego_PASS_AGENT,moving right of agent, red)
        \FTentry(ego_PASS_AGENT,stationary left of agent, red)
        \FTentry(ego_PASS_AGENT,stationary right of agent, red)
        \FTentry(ego_PASS_AGENT,lead agent, red)
        \FTentry(ego_PASS_AGENT,follow agent, red)
        \FTentry(ego_PASS_AGENT,stationary in front of agent, red)
        \FTentry(ego_PASS_AGENT,stationary behind agent, red)
        \FTentry(ego_PASS_AGENT,wait ped cross, red)
        \FTentry(ego_PASS_AGENT,accelerate, red)
        \FTentry(ego_PASS_AGENT,decelerate, red)
        \FTentry(ego_PASS_AGENT,stop, red)
        \FTentry(ego_PASS_AGENT,lane change, red)
        \FTentry(ego_PASS_AGENT,u turn, red)
        \FTentry(ego_PASS_AGENT,reverse, red)
        \FTentry(ego_PASS_AGENT,left turn, red)
        \FTentry(ego_PASS_AGENT,right turn, red)
        \FTentry(ego_PASS_AGENT,walk opposite, red)
        \FTentry(ego_PASS_AGENT,jaywalk, red)
        \FTentry(ego_PASS_AGENT,walk alongside, red)
        \FTentry(ego_PASS_AGENT,run, red)
        \FTentry(ego_PASS_AGENT,stand, red)
}
    \FTagentagent(group_agent-agent,ego_MOVING_LEFT_OF_AGENT,moving left of agent){
        \FTentry(ego_MOVING_LEFT_OF_AGENT,pass agent, red)
        \FTentry(ego_MOVING_LEFT_OF_AGENT,overtake agent, red)
        \FTentry(ego_MOVING_LEFT_OF_AGENT,stationary left of agent, red)
        \FTentry(ego_MOVING_LEFT_OF_AGENT,moving right of agent, red)
        \FTentry(ego_MOVING_LEFT_OF_AGENT,stationary right of agent, red)
        \FTentry(ego_MOVING_LEFT_OF_AGENT,lead agent, red)
        \FTentry(ego_MOVING_LEFT_OF_AGENT,follow agent, red)
        \FTentry(ego_MOVING_LEFT_OF_AGENT,stationary in front of agent, red)
        \FTentry(ego_MOVING_LEFT_OF_AGENT,stationary behind agent, red)
        \FTentry(ego_MOVING_LEFT_OF_AGENT,wait ped cross, red)
        \FTentry(ego_MOVING_LEFT_OF_AGENT,accelerate, red)
        \FTentry(ego_MOVING_LEFT_OF_AGENT,decelerate, red)
        \FTentry(ego_MOVING_LEFT_OF_AGENT,stop, red)
        \FTentry(ego_MOVING_LEFT_OF_AGENT,lane change, red)
        \FTentry(ego_MOVING_LEFT_OF_AGENT,u turn, red)
        \FTentry(ego_MOVING_LEFT_OF_AGENT,reverse, red)
        \FTentry(ego_MOVING_LEFT_OF_AGENT,left turn, red)
        \FTentry(ego_MOVING_LEFT_OF_AGENT,right turn, red)
}
    \FTagentagent(group_agent-agent,ego_MOVING_RIGHT_OF_AGENT,moving right of agent){
        \FTentry(ego_MOVING_RIGHT_OF_AGENT,pass agent, red)
        \FTentry(ego_MOVING_RIGHT_OF_AGENT,stationary behind agent, red)
        \FTentry(ego_MOVING_RIGHT_OF_AGENT,stationary right of agent, red)
        \FTentry(ego_MOVING_RIGHT_OF_AGENT,overtake agent, red)
        \FTentry(ego_MOVING_RIGHT_OF_AGENT,stationary in front of agent, red)
        \FTentry(ego_MOVING_RIGHT_OF_AGENT,stop, red)
        \FTentry(ego_MOVING_RIGHT_OF_AGENT,decelerate, red)
        \FTentry(ego_MOVING_RIGHT_OF_AGENT,left turn, red)
        \FTentry(ego_MOVING_RIGHT_OF_AGENT,reverse, red)
        \FTentry(ego_MOVING_RIGHT_OF_AGENT,accelerate, red)
        \FTentry(ego_MOVING_RIGHT_OF_AGENT,u turn, red)
        \FTentry(ego_MOVING_RIGHT_OF_AGENT,follow agent, red)
        \FTentry(ego_MOVING_RIGHT_OF_AGENT,moving left of agent, red)
        \FTentry(ego_MOVING_RIGHT_OF_AGENT,lane change, red)
        \FTentry(ego_MOVING_RIGHT_OF_AGENT,stationary left of agent, red)
        \FTentry(ego_MOVING_RIGHT_OF_AGENT,right turn, red)
        \FTentry(ego_MOVING_RIGHT_OF_AGENT,wait ped cross, red)
        \FTentry(ego_MOVING_RIGHT_OF_AGENT,lead agent, red)
    }
};
\end{tikzpicture}
    \end{subfigure}
    ~
    \begin{subfigure}[t]{0.32\textwidth}
        \centering
        \begin{tikzpicture}
\draw[color=black!60!white]
\FTcat(\FTroot,group_agent-agent,agent-agent){
    \FTpersonperson(group_agent-agent,ego_WALK_OPPOSITE,walk opposite){
        \FTentry(ego_WALK_OPPOSITE,walk alongside, red)
        \FTentry(ego_WALK_OPPOSITE,overtake agent, red)
        \FTentry(ego_WALK_OPPOSITE,pass agent, red)
        \FTentry(ego_WALK_OPPOSITE,run, red)
        \FTentry(ego_WALK_OPPOSITE,stand, red)
        \FTentry(ego_WALK_OPPOSITE,cross, red)
        \FTentry(ego_WALK_OPPOSITE,jaywalk, red)
}
    \FTagentagent(group_agent-agent,ego_LEAD_AGENT,lead agent){
        \FTentry(ego_LEAD_AGENT,stationary behind agent, red)
        \FTentry(ego_LEAD_AGENT,pass agent, red)
        \FTentry(ego_LEAD_AGENT,stationary right of agent, red)
        \FTentry(ego_LEAD_AGENT,overtake agent, red)
        \FTentry(ego_LEAD_AGENT,stationary in front of agent, red)
        \FTentry(ego_LEAD_AGENT,stop, red)
        \FTentry(ego_LEAD_AGENT,decelerate, red)
        \FTentry(ego_LEAD_AGENT,left turn, red)
        \FTentry(ego_LEAD_AGENT,reverse, red)
        \FTentry(ego_LEAD_AGENT,accelerate, red)
        \FTentry(ego_LEAD_AGENT,moving right of agent, red)
        \FTentry(ego_LEAD_AGENT,u turn, red)
        \FTentry(ego_LEAD_AGENT,follow agent, red)
        \FTentry(ego_LEAD_AGENT,moving left of agent, red)
        \FTentry(ego_LEAD_AGENT,lane change, red)
        \FTentry(ego_LEAD_AGENT,stationary left of agent, red)
        \FTentry(ego_LEAD_AGENT,right turn, red)
        \FTentry(ego_LEAD_AGENT,wait ped cross, red)
}
    \FTagentperson(group_agent-agent,ego_WAIT_PED_CROSS,wait ped cross){
        \FTentry(ego_WAIT_PED_CROSS,pass agent, red)
        \FTentry(ego_WAIT_PED_CROSS,overtake agent, red)
        \FTentry(ego_WAIT_PED_CROSS,overtake ego, red)
        \FTentry(ego_WAIT_PED_CROSS,follow agent, red)
        \FTentry(ego_WAIT_PED_CROSS,lead agent, red)
        \FTentry(ego_WAIT_PED_CROSS,follow ego, red)
        \FTentry(ego_WAIT_PED_CROSS,lead ego, red)
        \FTentry(ego_WAIT_PED_CROSS,decelerate, red)
        \FTentry(ego_WAIT_PED_CROSS,left turn, red)
        \FTentry(ego_WAIT_PED_CROSS,right turn, red)
        \FTentry(ego_WAIT_PED_CROSS,stop, red)
        \FTentry(ego_WAIT_PED_CROSS,reverse, red)
        \FTentry(ego_WAIT_PED_CROSS,u turn, red)
}
    \FTagentagent(group_agent-agent,ego_OVERTAKE_AGENT,overtake agent){
        \FTentry(ego_OVERTAKE_AGENT,pass agent, red)
        \FTentry(ego_OVERTAKE_AGENT,moving left of agent, red)
        \FTentry(ego_OVERTAKE_AGENT,moving right of agent, red)
        \FTentry(ego_OVERTAKE_AGENT,stationary left of agent, red)
        \FTentry(ego_OVERTAKE_AGENT,stationary right of agent, red)
        \FTentry(ego_OVERTAKE_AGENT,lead agent, red)
        \FTentry(ego_OVERTAKE_AGENT,follow agent, red)
        \FTentry(ego_OVERTAKE_AGENT,stationary in front of agent, red)
        \FTentry(ego_OVERTAKE_AGENT,stationary behind agent, red)
        \FTentry(ego_OVERTAKE_AGENT,wait ped cross, red)
        \FTentry(ego_OVERTAKE_AGENT,accelerate, red)
        \FTentry(ego_OVERTAKE_AGENT,decelerate, red)
        \FTentry(ego_OVERTAKE_AGENT,stop, red)
        \FTentry(ego_OVERTAKE_AGENT,lane change, red)
        \FTentry(ego_OVERTAKE_AGENT,u turn, red)
        \FTentry(ego_OVERTAKE_AGENT,reverse, red)
        \FTentry(ego_OVERTAKE_AGENT,left turn, red)
        \FTentry(ego_OVERTAKE_AGENT,right turn, red)
        \FTentry(ego_OVERTAKE_AGENT,walk alongside, red)
        \FTentry(ego_OVERTAKE_AGENT,walk opposite, red)
        \FTentry(ego_OVERTAKE_AGENT,run, red)
        \FTentry(ego_OVERTAKE_AGENT,stand, red)
        \FTentry(ego_OVERTAKE_AGENT,cross, red)
        \FTentry(ego_OVERTAKE_AGENT,jaywalk, red)
}
};
\end{tikzpicture}
    \end{subfigure}
    \caption{\textbf{Scenario catalog.} 
    Agent-Agent scenarios with assigned negative scenarios in red.}
    \label{fig:supp:scenatio_catalog:agent_agent}
\end{figure*}

\clearpage

\begin{table}[t]
    \centering
    \caption{\textbf{Ego-to-agent scenario definition.}
    Textual description of scenarios where the ego-vehicle interacts with other agents.}
    \label{tab:supp:scenatio_catalog:description:ego_agent}
    \vspace{.5em}
    \begin{tabularx}{\linewidth}{@{}>{\bfseries}r@{\hspace{1em}}X@{}}
        \toprule
        \textbf{lead agent} & Ego travels ahead of agent at a similar speed while maintaining a consistent distance.\\
        \textbf{follow agent} & Ego is driving behind agent at a similar speed while maintaining a consistent distance.\\
        \textbf{pass agent} & Ego in the adjacent lane overtakes the stopped agent.\\
        \textbf{overtake agent} & Ego in the adjacent lane and moves ahead of agent while both are in motion.\\
        \textbf{stationary left of agent} & Agent is fully stopped and remains stationary to the left of agent, which is also stationary, such as when waiting at a traffic light or in a parking lot.\\
        \textbf{stationary in front of agent} & Agent is fully stopped and remains stationary in front of agent, such as when waiting at a traffic light, in a parking lot, or any other situation requiring queuing.\\
        \textbf{wait ped cross} & Ego comes to a stop or remains stationary, yielding the right-of-way to agent who is crossing or preparing to cross the road, while maintaining awareness of the agent's movement and ensuring a safe distance until the crossing is complete.\\
        \bottomrule
    \end{tabularx}    
\end{table}

\begin{table}[t]
    \centering
    \caption{\textbf{Agent scenario definition.}
    Textual description of agent scenarios.}
    \label{tab:supp:scenatio_catalog:description:agent}
    \vspace{.5em}
    \begin{tabularx}{\linewidth}{@{}>{\bfseries}r@{\hspace{1em}}X@{}}
        \toprule
        \textbf{stand} & Agent (pedestrian) remains stationary in the traffic environment, either waiting at a crossing, observing surroundings, or pausing for other reasons.\\
        \textbf{walk} & Agent (pedestrian) moves at a steady, moderate pace, typically following designated paths or crosswalks.\\
        \textbf{jaywalk} & Agent (pedestrian) crosses the street outside of designated crossing areas or against traffic signals, often requiring heightened awareness of vehicle movements, quick decision-making to avoid conflicts, and potentially creating unpredictable interactions with other agents in the traffic environment.\\
        \textbf{run} & Agent (pedestrian) is running and moves rapidly.\\
        \textbf{cross} & Agent (pedestrian) moves from one side of the road to the other, at a designated crossing point or intersection.\\
        \midrule
        \textbf{accelerate} & Agent is increasing its speed, either gradually or abruptly, to adapt to traffic conditions, maintain flow, or comply with traffic rules and signals.\\
        \textbf{stop} & Agent is reducing its speed, either gradually or abruptly, in response to traffic conditions, obstacles, or to comply with traffic rules and comes to a complete stop.\\
        \textbf{reverse} & Agent is moving in reverse, either to park, navigate a tight space, or adjust its position.\\
        \textbf{left turn} & Agent is executing a left turn at an intersection or junction.\\
        \textbf{right turn} & Agent is executing a right turn at an intersection or junction.\\
        \textbf{u turn} & Agent is performing a 180-degree turn at an intersection or junction, reversing its direction of travel.\\
        \textbf{lane change} & Agent is transitioning from its current lane to an adjacent lane.\\
        \textbf{lead ego} & Agent travels ahead of ego at a similar speed while maintaining a consistent distance.\\
        \textbf{follow ego} & Agent is driving behind ego at a similar speed while maintaining a consistent distance.\\
        \textbf{overtake ego} & Agent in the adjacent lane moves ahead of ego while both are in motion.\\
        \textbf{stationary right of ego} & Agent is fully stopped and remains stationary to the right of ego, which is also stationary, such as when waiting at a traffic light or in a parking lot.\\
        \textbf{stationary behind ego} & Agent is fully stopped and remains stationary behind ego (which is also stopped), such as when waiting at a traffic light, in a parking lot, or in any other queuing scenario.\\
        \bottomrule
    \end{tabularx}
\end{table}

\begin{table}[t]
    \centering
    \caption{\textbf{Ego scenario definition.}
    Textual description of ego-vehicle scenarios.}
    \label{tab:supp:scenatio_catalog:description:ego}
    \vspace{.5em}
    \begin{tabularx}{\linewidth}{@{}>{\bfseries}r@{\hspace{1em}}X@{}}
        \toprule
        \textbf{accelerate} & Ego is increasing its speed, either gradually or abruptly, to adapt to traffic conditions, maintain flow, or comply with traffic rules and signals.\\
        \textbf{decelerate} & Ego is reducing its speed, either gradually or abruptly, in response to traffic conditions, obstacles, or to comply with traffic rules, without coming to a complete stop.\\
        \textbf{stop} & Ego is reducing its speed, either gradually or abruptly, in response to traffic conditions, obstacles, or to comply with traffic rules and comes to a complete stop.\\
        \textbf{left turn} & Ego is executing a left turn at an intersection or junction.\\
        \textbf{right turn} & Ego is executing a right turn at an intersection or junction.\\
        \textbf{lane change} & Ego is transitioning from its current lane to an adjacent lane.\\
        \bottomrule
    \end{tabularx}
\end{table}

\begin{table}[t]
    \centering
    \caption{\textbf{Agent-to-agent scenario definition.}
    Textual description of scenarios where agents interact with other agents.}
    \label{tab:supp:scenatio_catalog:description:agent_agent}
    \vspace{.5em}
    \begin{tabularx}{\linewidth}{@{}>{\bfseries}r@{\hspace{1em}}X@{}}
        \toprule
        \textbf{walk alongside} & Agent (pedestrian) and agent (pedestrian) walk side by side at a steady, moderate pace.\\
        \textbf{walk opposite} & Agent (pedestrian) and agent (pedestrian) walk toward each other at a moderate pace, cross paths, and proceed.\\
        \midrule
        \textbf{lead agent} & Agent travels ahead of agent at a similar speed while maintaining a consistent distance.\\
        \textbf{follow agent} & Agent is driving behind agent at a similar speed while maintaining a consistent distance.\\
        \textbf{pass agent} & Agent in the adjacent lane overtakes the stopped agent.\\
        \textbf{overtake agent} & Agent in the adjacent lane and moves ahead of agent while both are in motion.\\
        \textbf{moving left of agent} & Agent is traveling in parallel to the left of agent (e.g., in adjacent lanes or side by side), with one vehicle maintaining a leftward offset relative to the other. This could occur during lane-matched driving on a multi-lane road or synchronized movement from a traffic light.\\
        \textbf{moving right of agent} & Agent is traveling in parallel to the right of agent (e.g., in adjacent lanes or side by side), with one vehicle maintaining a rightward offset relative to the other. This could occur during lane-matched driving on a multi-lane road or synchronized movement from a traffic light.\\
        \textbf{stationary left of agent} & Agent is fully stopped and remains stationary to the left of agent, which is also stationary, such as when waiting at a traffic light or in a parking lot.\\
        \textbf{stationary right of agent} & Agent is fully stopped and remains stationary to the right of agent, which is also stationary, such as when waiting at a traffic light or in a parking lot.\\
        \textbf{stationary in front of agent} & Agent is fully stopped and remains stationary in front of agent, such as when waiting at a traffic light, in a parking lot, or any other situation requiring queuing.\\
        \textbf{stationary behind agent} & Agent is fully stopped and remains stationary behind agent, such as when waiting at a traffic light, in a parking lot, or any other situation requiring queuing.\\
        \textbf{wait ped cross} & Agent comes to a stop or remains stationary, yielding the right-of-way to a agent who is crossing or preparing to cross the road, while maintaining awareness of the agent's movement and ensuring a safe distance until the crossing is complete.\\
        \bottomrule
    \end{tabularx}
\end{table}

\clearpage

\subsection{Scenario verification}
\begin{figure}[t]
    \centering
    \includegraphics[width=0.5\textwidth]{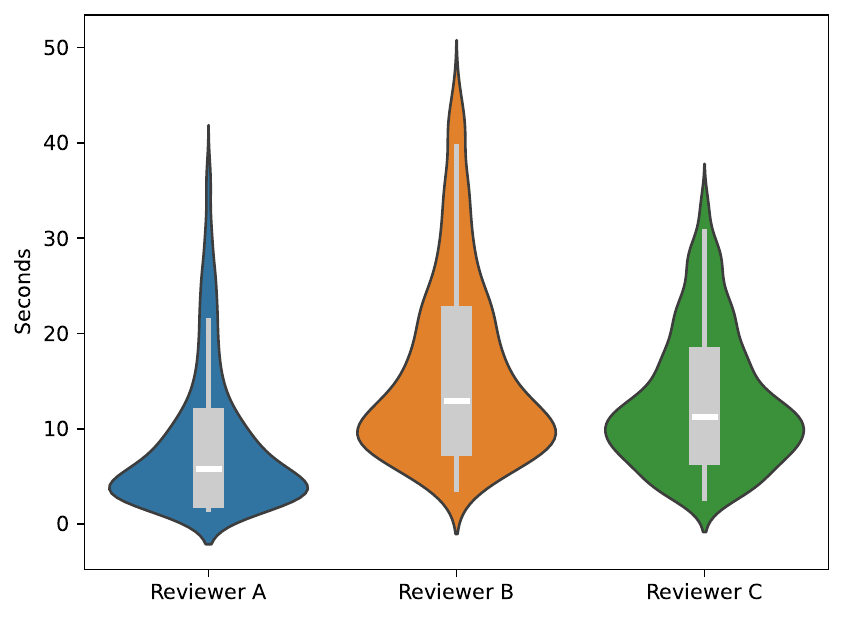}
    \caption{\textbf{Validation time.}
    Comparison of the validation duration for each reviewer.}
    \label{fig:sup:validation_time}
\end{figure}

\begin{figure}[t]
    \centering
    \begin{subfigure}{1\textwidth}
        \centering
        \includegraphics[height=0.565\linewidth, width=1\linewidth]{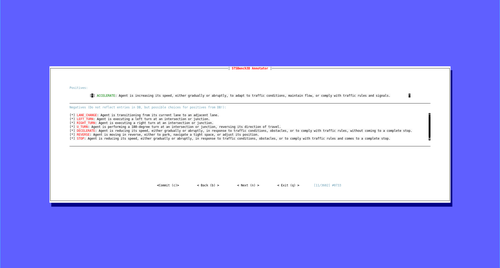}
        \caption{Verification GUI.}
        \label{fig:supp:verification:tool:gui}
    \end{subfigure}
    \vspace{1em}\\
    \begin{subfigure}{1\textwidth}
        \centering
        \includegraphics[height=0.565\linewidth, width=1\linewidth]{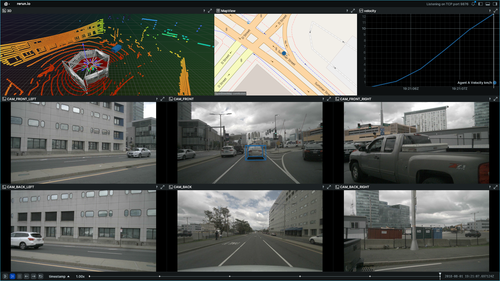}
        \caption{Verification rerun.}
        \label{fig:supp:verification:tool:rerun}
    \end{subfigure}
    \caption{
        Verification tool for fast and simple verification~(\subref{fig:supp:verification:tool:gui}) of mined traffic scenarios by inspecting~(\subref{fig:supp:verification:tool:rerun}) recordings from various available modalities.
    }
    \label{fig:supp:verification:tool}
\end{figure}

\paragraph{Verification tool.}
Fast and simple verification requires a clean and versatile visualization interface paired with a simple input mask. 
In Fig.~\ref{fig:supp:verification:tool}, we show the verification command line tool~(\subref{fig:supp:verification:tool:gui}) of \ourBenchmark, that can be used via mouse or keyboard only. The visualization interface~(\subref{fig:supp:verification:tool:rerun}) uses Rerun~\cite{rerunteam_2025_rerunsdk} and shows various available input modalities, such as LiDAR point clouds, agent trajectories on a map, velocities, and multi-view camera images for consecutive time frames. The presented validation tool enables fast inspection of mined scenarios in different modalities and effortless acceptance or rejection of samples.

\paragraph{Verification insights.} 
\label{sec:supp:benchmark_details:verification_details}

To demonstrate the efficiency of our verification, we provide a detailed time analysis in Fig.~\ref{fig:sup:validation_time}. We observe an average of 8.3 seconds, 16.0 seconds, and 12.9 seconds for reviewers A, B, and C, respectively. While some scenes require more attention, most of the scenarios are obvious and can therefore be confirmed quickly.

\begin{figure*}[h]
    \centering
    \begin{subfigure}[t]{0.5\textwidth}
        \centering
        \includegraphics[trim=57 0 57 0,clip,width=1.0\textwidth]{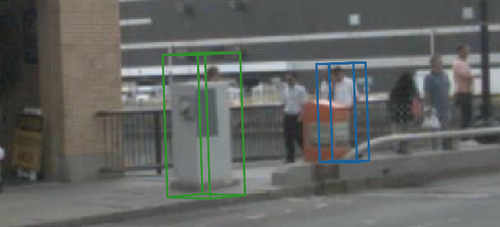}
        \caption{First frame (CAM\_FRONT\_LEFT).}
        \label{fig:supp:verification:walk_opposite_occ:1}
    \end{subfigure}%
    ~ 
    \begin{subfigure}[t]{0.5\textwidth}
        \centering
        \includegraphics[trim=57 0 57 0,clip,width=1.0\textwidth]{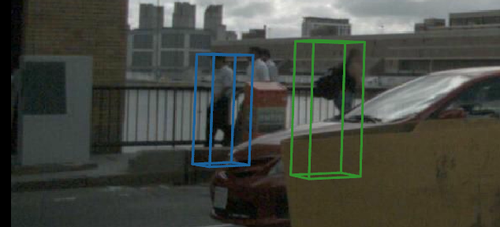}
        \caption{Fourth frame (CAM\_FRONT\_LEFT).}
        \label{fig:supp:verification:walk_opposite_occ:2}
    \end{subfigure}
    \caption{\textbf{Agent walk opposite.}
    Accepted scenario sample after review, since the occlusion level has been rated difficult but reasonable. The bounding box is not visible for the tested models.}
    \label{fig:supp:verification:walk_opposite_occ}
\end{figure*}

\begin{figure*}[th]
    \centering
    \begin{subfigure}[t]{0.5\textwidth}
        \centering
        \includegraphics[trim=57 0 57 0,clip,width=1.0\textwidth]{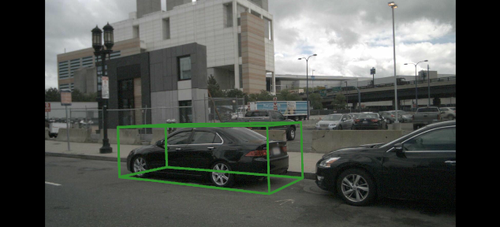}
        \caption{First frame (CAM\_FRONT\_RIGHT).}
        \label{fig:supp:verification:ego_pass_agent:1}
    \end{subfigure}%
    ~ 
    \begin{subfigure}[t]{0.5\textwidth}
        \centering
        \includegraphics[trim=57 0 57 0,clip,width=1.0\textwidth]{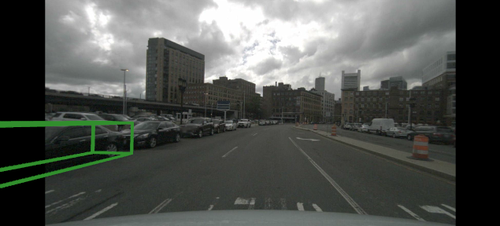}
        \caption{Last frame (CAM\_BACK).}
        \label{fig:supp:verification:ego_pass_agent:2}
    \end{subfigure}
    \caption{\textbf{Ego-vehicle passes agent.}
    The scenario sample was rejected after review since the ego-vehicle is not in the adjacent lane of the refereed agent. The bounding box is not visible for the tested models.}
    \label{fig:supp:verification:ego_pass_agent}
\end{figure*}

\begin{figure*}[th]
    \centering
    \begin{subfigure}[t]{0.5\textwidth}
        \centering
        \includegraphics[trim=57 0 57 0,clip,width=1.0\textwidth]{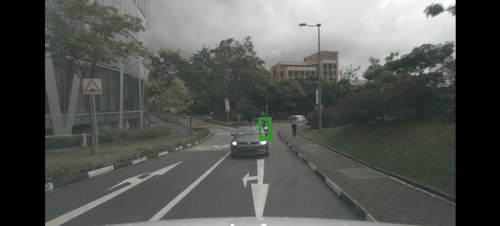}
        \caption{First frame (CAM\_BACK).}
        \label{fig:supp:verification:agent_follow_ego:1}
    \end{subfigure}%
    ~ 
    \begin{subfigure}[t]{0.5\textwidth}
        \centering
        \includegraphics[trim=57 0 57 0,clip,width=1.0\textwidth]{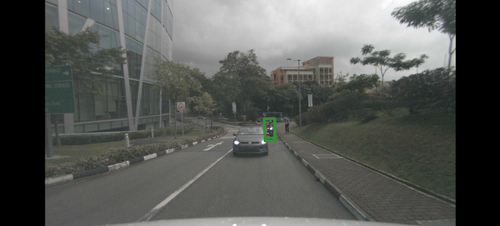}
        \caption{Last frame (CAM\_BACK).}
        \label{fig:supp:verification:agent_follow_ego:2}
    \end{subfigure}
    \caption{\textbf{Agent following ego-vehicle.}
    Rejected scenario sample after review since there is another agent between the ego-vehicle and the refereed agent. The bounding box is not visible for the tested models.}
    \label{fig:supp:verification:agent_follow_ego}
\end{figure*}

\begin{figure*}[!th]
    \centering
    \begin{subfigure}[t]{0.5\textwidth}
        \centering
        \includegraphics[trim=57 0 57 0,clip,width=1.0\textwidth]{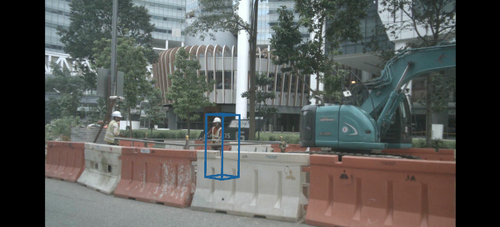}
        \caption{Construction worker in a safe space.}
        \label{fig:supp:verification:jaywalk_neg:1}
    \end{subfigure}%
    ~ 
    \begin{subfigure}[t]{0.5\textwidth}
        \centering
        \includegraphics[trim=57 0 57 0,clip,width=1.0\textwidth]{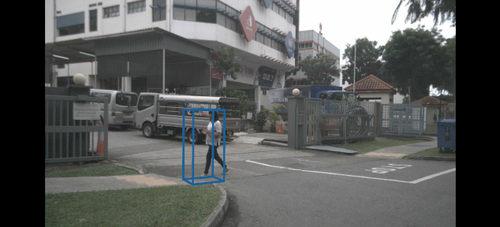}
        \caption{Pedestrian on sidewalk.}
        \label{fig:supp:verification:jaywalk_neg:2}
    \end{subfigure}
    \caption{\textbf{Agent jaywalking.}
    Rejected scenario sample after review since there is no jaywalking. The bounding box is not visible for the tested models.}
    \label{fig:supp:verification:jaywalk_neg}
\end{figure*}

In the following, we show typical verification mistakes and disagreements between reviewers leading to either accepted or rejected scenarios. These verification differences are solved by majority voting. 
Fig.~\ref{fig:supp:verification:walk_opposite_occ} shows a highly occluded person (blue), which has been considered too difficult by one of the three reviewers. 
However, although only a small part of the person is visible, this sample can serve as a difficult scenario for the benchmark.
The next two scenes in Fig.~\ref{fig:supp:verification:ego_pass_agent} and Fig.~\ref{fig:supp:verification:agent_follow_ego} illustrate scenarios that the reviewers have rejected because they do not follow the scenario definition exactly. In Fig.~\ref{fig:supp:verification:ego_pass_agent}, the ego-vehicle \emph{passing} the agent (green) is not in the adjacent line. Fig.~\ref{fig:supp:verification:agent_follow_ego} shows a similar issue, where the agent (green) \emph{following} the ego-vehicle has another agent in between, which is not considered following the ego-vehicle in our scenario definition. 
For \emph{jaywalking}, the definition states that a pedestrian crosses the street outside designated crossing areas. However, not all designated areas are properly annotated in map data, as shown in Fig.~\ref{fig:supp:verification:jaywalk_neg}. Construction workers operating in a safe construction space (Fig.~\ref{fig:supp:verification:jaywalk_neg:1}) or pedestrians crossing a driveway between two sidewalks (Fig.~\ref{fig:supp:verification:jaywalk_neg:2}) are not considered jaywalking.

\begin{figure*}[!t]
    \centering
    \begin{subfigure}[t]{0.5\textwidth}
        \centering
        \includegraphics[trim=57 0 57 0,clip,width=1.0\textwidth]{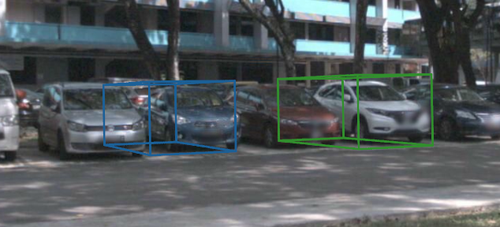}
        \caption{First frame (CAM\_FRONT\_LEFT).}
        \label{fig:supp:verification:agent_right_of_agent:1}
    \end{subfigure}%
    ~ 
    \begin{subfigure}[t]{0.5\textwidth}
        \centering
        \includegraphics[trim=148 0 149 0,clip,width=1.0\textwidth]{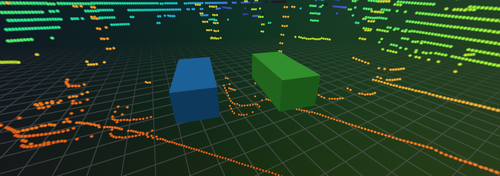}
        \caption{3D LiDAR scene.}
        \label{fig:supp:verification:agent_right_of_agent:2}
    \end{subfigure}
    \caption{\textbf{Agent right of agent.}
    The scenario sample was rejected after review since there is another agent between the two involved agents. The bounding box is not visible for the tested models.}
    \label{fig:supp:verification:agent_right_of_agent}
\end{figure*}

Disagreements can also occur due to inaccurate visualizations of projected 3D bounding boxes onto the camera image. In Fig.~\ref{fig:supp:verification:agent_right_of_agent}, we illustrate an example where bounding boxes indicate two neighboring vehicles (green, blue) in the camera image~(\subref{fig:supp:verification:agent_right_of_agent:1}). 
However, a closer look at the LiDAR scan~(\subref{fig:supp:verification:agent_right_of_agent:2}) reveals another agent between them. Consequently, the ambiguous verification of such samples could either stem from the inaccurate projection or a misunderstanding of the scenario description.

\begin{figure*}[!th]
    \centering
    \begin{subfigure}[t]{0.5\textwidth}
        \centering
        \includegraphics[trim=57 0 57 0,clip,width=1.0\textwidth]{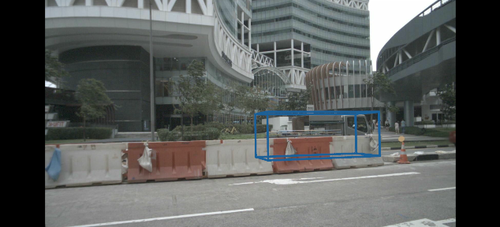}
        \caption{First frame (CAM\_BACK\_RIGHT).}
        \label{fig:supp:verification:lane_change:1}
    \end{subfigure}%
    ~ 
    \begin{subfigure}[t]{0.5\textwidth}
        \centering
        \includegraphics[trim=0 0 23 0,clip,width=1.0\textwidth]{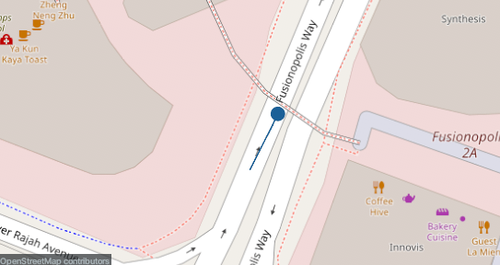}
        \caption{Trajectory on map.}
        \label{fig:supp:verification:lane_change:2}
    \end{subfigure}
    \caption{\textbf{Agent lane change.}
    The scenario sample was rejected after review since no clear lane change was visible. The bounding box is not visible for the tested models.}
    \label{fig:supp:verification:lane_change}
\end{figure*}

Finally, subjective perception can also lead to ambiguous verification results. Fig.~\ref{fig:supp:verification:lane_change} shows a lane change scenario hardly recognizable from the trajectory in the map view~(\subref{fig:supp:verification:lane_change:2}). However, the vehicle (blue) is heavily occluded, and lane markings are not visible in the camera image~(\subref{fig:supp:verification:lane_change:1}). Thus, such hardly recognizable scenarios have been rejected in the manual verification stage.

\clearpage
\subsection{Multiple-choice question generation}
\label{sec:supp:details_question_gen}
The evaluation on our benchmark requires the generation of input prompts that contain the task description and a question.
In addition to the preamble (setting the context) and scenario descriptions (for all answer choices), we ask the models one of the following questions, depending on the scenario category:
\begin{itemize}
    \item \textbf{Ego:} Which of the following options best describes ego driving maneuver?
    \item \textbf{Agent:} Which of the following options best describes the driving behavior of the <reference to agent>?
    \item \textbf{Ego-to-agent:} Which of the following options best describes the ego driving behavior with respect to the <reference to agent>?
    \item \textbf{Agent-to-agent:} Which of the following options best describes <reference to agent 1> maneuver with respect to the <reference to agent 2>?
\end{itemize}
Notice that this example is only valid for \drivemm and requires adaptation for other models in order to get better performance. For instance, the fine-tuning of \sennavlm always refers in a first-person manner to the ego-vehicle, which requires altering questions respectively.
Detailed prompting examples for all evaluated models are provided in Sec.\ref{sec:supp:prompt_examples}.

\section{Driving expert baselines}
\begingroup
    \setlength\tabcolsep{6.6pt}
    \begin{table}[t!]
        \caption{\textbf{End-to-end driving models.}
        Driving expert VLMs for end-to-end driving, \ie, open-loop planning (OLP), closed-loop driving (CLD), and control signal prediction (CSP). Additional proxy tasks are scene generation (SG), dense captioning (CAP), and counterfactual reasoning (CR). Rows marked in gray provide explicit reasoning, code, and pre-trained model weights. Code marked with "\textbf{\textasciitilde}" does not include evaluation scripts or configurations for \nus.}
        \label{tab:intro:model_comparison}
        \scriptsize
        \begin{center}
            \begin{tabular*}{\textwidth}{lccccccccc}
                \toprule
                Method                 & Evaluation Dataset & Environment & Task           & Explicit reasoning & Model Weights      & Code            \\
                \midrule
                \orion                 & \nus, \carlac      & Real, Sim   & CLD, OLP       & \cmark             & \cmark             & \textbf{\textasciitilde} \\
                \rowcolor{gray!20}
                \drivemm               & \nus               & Real        & VQA            & \cmark             & \cmark             & \cmark          \\
                \rdadriver             & \nus               & Real        & OLP            & \cmark             & \xmark             & \xmark          \\
                \emma                  & \wod, \nus         & Real        & OLP            & \cmark             & \xmark             & \xmark          \\
                \rowcolor{gray!20}
                \sennavlm              & \nus, \sennads     & Real        & CSP            & \cmark             & \cmark             & \cmark          \\
                \gpvl                  & \nus               & Real        & OLP            & \cmark             & \xmark             & \cmark          \\
                \vlp                   & \nus               & Real        & OLP            & \xmark             & \xmark             & \xmark          \\
                \lmdrive               & \carlac            & Sim         & CLD            & \xmark             & \cmark             & \cmark          \\
                \drivevlmdual          & \nus               & Real        & OLP            & \cmark             & \xmark             & \xmark          \\
                \insightdrive          & \nus               & Real        & OLP            & \cmark             & \xmark             & \xmark          \\
                \rowcolor{gray!20}
                \omnidrive             & \nus               & Real        & OLP, CR        & \cmark             & \cmark             & \cmark          \\
                \bevdriver             & \carlac            & Sim         & CLD            & \xmark             & \xmark             & \xmark          \\
                \scetwodrivex          & \nus               & Real        & OLP, CSP       & \cmark             & \xmark             & \xmark          \\
                \hermes                & \nus               & Real        & SG, OLP        & \cmark             & \xmark             & \xmark          \\
                \opendrivevla          & \nus               & Real        & OLP, CAP       & \cmark             & \xmark             & \xmark          \\
                \bottomrule        
            \end{tabular*}
        \end{center}
    \end{table}
\endgroup

In this section, we discuss the choice of driving expert models from recent publications listed in Table~\ref{tab:intro:model_comparison}. To be considered for evaluation, the model must have explicit reasoning capabilities in addition to its primary task. Furthermore, the code and model weights must be publicly available to guarantee fair comparison. Therefore, we opted for \drivemm, \sennavlm, and \omnidrive.

\paragraph{DriveMM.} 
Not particularly developed for the end-to-end driving task, \drivemm performs joint training on multiple driving datasets. The model architecture is inspired by LLaVa~\cite{liu_2023_llava} and comprises a SigLIP~\cite{zhai_2023_siglip} vision encoder and Llama-3.1~\cite{meta_2025_llama3} as LLM. In the first step, the model gets pre-trained on a multi-modal dataset that includes text-image pairs for images and videos. For the final fine-tuning, the model learns from single images, multi-view images, and videos, leveraging various driving datasets, \ie, \drivelm, \lingoqa, \nuinstruct, \omnidrive, \maplm, and \codalm. Hence, the model learns to process various input modalities from single images to multi-view videos.

\paragraph{OmniDrive.} 
This model is designed to provide driving decisions and waypoints for planning. It encodes multi-view images into BEV features leveraging the StreamPETR~\cite{wang_2023_streampetr} encoder and maps this representation into language space, incorporating a Q-Former~\cite{li_2023_blip2}. The training contains a series of objectives, including 3D grounding, open-loop planning, counterfactual reasoning, and scene understanding. Annotations for training on the \nusc dataset are automatically generated leveraging ground truth information and VLMs. 

\paragraph{Senna.}
This end-to-end driving model consists of two modules, Senna-VLM and SennaE2E. The idea is to predict high-level planning decisions from multi-view video inputs with Senna-VLM and encode them into meta-action features, enhancing the final planning trajectories of SennaE2E, which is a state-of-the-art end-to-end driving planner~\cite{jiang_2023_vad}. The images are encoded with CLIP~\cite{radford_2021_clip} and compressed in order to reduce the number of input tokens. The model is trained in multiple steps, starting with pre-training on multiple sources, including the instruction following data from LlaVa~\cite{liu_2023_llava}. Two fine-tuning steps follow on private data with automatically generated VLM annotations, leveraging ground truth information. First, the model learns to describe the scene, detect traffic signals, identify vulnerable road users, and so on. Second, the model learns to predict meta-actions such as accelerate, decelerate, left, right, \etc.

\section{Detailed results \& analysis}
\label{sec:supp:detailed_analysis}
We provide detailed evaluations for ego, agent, ego-to-agent, and agent-to-agent scenarios in Table~\ref{tab:detailed_res_ego},~\ref{tab:detailed_res_agent},~\ref{tab:detailed_res_ego_agent}, and~\ref{tab:detailed_res_agent_agent}, respectively. Therefore, we list the accuracy for each scenario and the weighted average for the corresponding category. 
In addition to evaluating the performance of the individual models, we also gain the following insights:

\paragraph{Ego scenarios.}
Large language models (LLMs) perform exceptionally well for ego scenarios, shown in Table~\ref{tab:detailed_res_ego}, as they have the ground truth trajectory of the ego-vehicle at their disposal. Especially \gptfo with advanced reasoning can successfully derive scenarios from given trajectories. We illustrate the model's reasoning after correctly identifying a \emph{left turn} in Fig.~\ref{fig:prompt:gptfo_ego_lt}.
An exception is the differentiation between the closely related scenarios \emph{decelerate} and \emph{stop}. Since the vehicle may not be fully stationary in the trajectory data, such as when moving at less than 1 km/h, the language model typically interprets the situation as deceleration. In contrast, powerful vision-language models (VLMs) often recognize it as a potential stop, based on contextual cues like a vehicle ahead, a red traffic light, or a stop sign. 
However, end-to-end driving models fail to perform well across different scenarios. Only \drivemm provides reasonable results for more obvious scenarios such as \emph{left turn} or \emph{right turn}.

\begin{table*}[t]
\centering
\caption{
    \textbf{Ego scenario evaluation.}
    Scenario-level performance comparison of LLMs, off-the-shelf VLMs, and driving expert VLMs for ego scenarios.
    Accuracy grouped by scenario categories: Left Turn (\textbf{LT}), Accelerate (\textbf{Acc}), Right Turn (\textbf{RT}), Lane Change (\textbf{LC}), Decelerate (\textbf{Dec}), Stop and Average (\textbf{Avg}).
    The best results are highlighted in bold.
}
\label{tab:detailed_res_ego}
\begin{tabular}{l*{6}{c}c}
\toprule
            & \textbf{LT}                & \textbf{Acc}       & \textbf{RT}        & \textbf{LC}        & \textbf{Dec}       & \textbf{Stop}      & \textbf{Avg}       \\
\midrule
\llama      & \phantom{0}\phantom{0}9.1 & 24.3              & \phantom{0}0.0    & 18.8              & 29.4              & 18.2              & 16.6              \\
\deepseek   & \phantom{0}27.3           & 86.5              & 10.0              & 12.5              & 41.2              & \phantom{0}9.1    & 31.1              \\
\gptfo      & \bestresult{100.0}        & \bestresult{91.9} & 50.0              & \bestresult{68.8} & 52.9              & 18.2              & \bestresult{63.6} \\
\midrule
\internvlob & \phantom{0}\phantom{0}0.0 & 29.7              & 20.0              & 37.5              & 23.5              & \phantom{0}9.1    & 20.0              \\
\quenvlsb   & \phantom{0}36.4           & \phantom{0}2.7    & \bestresult{70.0} & \phantom{0}6.3    & 52.9              & 45.5              & 35.6              \\
\internvlab & \phantom{0}18.2           & 21.6              & 50.0              & \phantom{0}6.3    & \bestresult{70.6} & \bestresult{63.6} & 38.4              \\
\midrule
\sennavlm   & \phantom{0}18.2           & 10.8              & \phantom{0}0.0    & \phantom{0}6.3    & 17.7              & \phantom{0}0.0    & \phantom{0}8.8    \\
\omnidrive  & \phantom{0}25.0           & 25.0              & 23.8              & 35.3              & 27.8     & 25.0              & 24.4              \\
\drivemm    & \phantom{0}63.6           & 51.4              & \bestresult{70.0} & 18.8              & 11.8              & 45.5              & 43.5              \\
\bottomrule
\end{tabular}
\end{table*}

\paragraph{Agent scenarios.}
Table~\ref{tab:detailed_res_agent} shows the detailed evaluation of various models on agent scenarios. These scenarios that involve a single agent are more difficult for LLMs. A simple reason is that some of these scenarios, \ie, \emph{jaywalking}, \emph{crossing}, require road markings to be detected correctly. Surprisingly, \gptfo performs above random guessing. However, this outcome is less indicative of true scenario understanding and more a reflection of the model’s strong general reasoning capabilities. We illustrate this in Fig.~\ref{fig:prompt:gptfo_jaywalking}, where \gptfo arrives at the correct answer through a process of elimination. The best overall performance for this category reaches \internvlab. The VLM has difficulties with challenging scenarios like \emph{jaywalking} and \emph{lane change}, requiring in-depth visual understanding, but also with scenarios that involve the ego vehicle, such as \emph{overtaking ego}, \emph{lead ego}, and \emph{follow ego}. The relatively bad performance of driving expert models highlights the need to enhance spatio-temporal understanding for these models. 

\begingroup
    \setlength\tabcolsep{1.5pt}
\begin{table*}[t]
\scriptsize
\centering
\caption{
    \textbf{Agent scenario evaluation.}
    Scenario-level performance comparison of LLMs, off-the-shelf VLMs, and driving expert VLMs for ego-to-agent scenarios.
    Accuracy grouped by scenario categories: Jaywalk (\textbf{JW}), Left Turn (\textbf{LT}), Walk (\textbf{W}), Stand (\textbf{S}), Cross (\textbf{C}), Accelerate (\textbf{Acc}), Right Turn (\textbf{RT}), Lane Change (\textbf{LC}), Overtake Ego (\textbf{OE}), Follow Ego (\textbf{FE}), Stop, Run, Lead Ego (\textbf{LE}), Reverse (\textbf{Rev}), U Turn (\textbf{UT}), Stationary Right Of Ego (\textbf{SRoE}), Stationary Behind Ego (\textbf{SBE}) and Average (\textbf{Avg}).
    The best results are highlighted in bold.
}
\label{tab:detailed_res_agent}

\begin{tabular*}{\textwidth}{lccccccccccccccccccc}
\toprule
            & JW                            & LT                            & W                             & S                             & C                             & Acc                           & RT                            & LC                            & OE                            & FE                            & Stop                          & Run                           & LE                            & Rev                 & UT              & SRoE                & SBE                 & Avg                           \\
\midrule
\llama      & \phantom{0}16.0              & \phantom{0}16.7              & \phantom{0}20.8              & \phantom{00}9.1               & \phantom{0}10.4              & \phantom{0}32.0              & \phantom{0}15.4              & \phantom{0}29.2              & \phantom{00}0.0               & \phantom{0}20.5              & \phantom{0}10.3              & \phantom{0}20.0              & \phantom{0}35.3              & \phantom{0}0.0     & \phantom{0}0.0 & \phantom{00}0.0     & \phantom{00}0.0     & \phantom{0}13.8              \\
\deepseek   & \phantom{0}28.0              & \phantom{0}45.8              & \phantom{0}33.3              & \phantom{0}13.6              & \phantom{0}41.7              & \phantom{0}20.0              & \phantom{0}30.8              & \phantom{0}12.5              & \bestresult{\phantom{0}83.3} & \phantom{0}89.7              & \phantom{0}20.7              & \phantom{00}0.0               & \bestresult{\phantom{0}94.1} & \phantom{0}0.0     & \phantom{0}0.0 & \bestresult{100.0} & \bestresult{100.0} & \phantom{0}42.0              \\
\gptfo      & \phantom{0}36.0              & \bestresult{\phantom{0}62.5} & \phantom{0}25.0              & \phantom{0}13.6              & \phantom{0}33.3              & \phantom{0}52.0              & \phantom{0}38.5              & \phantom{0}16.7              & \phantom{0}33.3              & \bestresult{\phantom{0}94.9} & \phantom{00}6.9              & \bestresult{\phantom{0}80.0} & \phantom{0}82.4              & \phantom{0}0.0     & \phantom{0}0.0 & \bestresult{100.0} & \bestresult{100.0} & \phantom{0}45.6              \\
\midrule
\internvlob & \phantom{0}16.0              & \phantom{0}29.2              & \phantom{0}35.4              & \phantom{00}0.0              & \phantom{0}70.8              & \phantom{0}22.0              & \phantom{0}26.9              & \phantom{0}16.7              & \phantom{0}66.7              & \phantom{0}15.4              & \phantom{0}17.2              & \phantom{0}20.0              & \phantom{0}11.7              & \phantom{00}0.0     & \phantom{0}0.0 & \phantom{00}0.0     & \phantom{00}0.0     & \phantom{0}20.5              \\
\quenvlsb   & \phantom{00}4.0              & \phantom{0}29.2              & \phantom{0}68.8              & \phantom{0}72.7              & \phantom{0}79.2              & \phantom{00}2.0              & \phantom{0}42.3              & \phantom{0}20.8              & \phantom{0}33.3              & \phantom{0}61.5              & \phantom{0}10.3              & \bestresult{\phantom{0}80.0} & \phantom{0}17.7              & \phantom{00}0.0     & \phantom{0}0.0 & \phantom{0}50.0    & \phantom{0}50.0    & \phantom{0}36.6              \\
\internvlab & \phantom{00}8.0              & \phantom{0}16.7              & \bestresult{\phantom{0}85.4} & \bestresult{\phantom{0}75.0} & \bestresult{\phantom{0}89.6} & \phantom{0}28.0              & \bestresult{\phantom{0}65.4} & \phantom{0}33.3              & \phantom{0}33.3              & \phantom{0}64.1              & \phantom{0}20.7              & \phantom{0}40.0              & \phantom{0}47.1              & \bestresult{100.0} & \phantom{0}0.0 & \bestresult{100.0} & \bestresult{100.0} & \bestresult{\phantom{0}53.3} \\
\midrule
\sennavlm   & \bestresult{\phantom{0}32.0}              & \phantom{00}0.0              & \phantom{0}58.3              & \phantom{0}31.8              & \phantom{0}35.4              & \phantom{0}30.0              & \phantom{0}30.8              & \phantom{0}12.5              & \phantom{0}33.3              & \phantom{0}61.5              & \phantom{00}6.9              & \phantom{0}20.0              & \phantom{0}41.2              & \phantom{00}0.0     & \phantom{0}0.0 & \phantom{0}50.0    & \phantom{00}0.0     & \phantom{0}26.1              \\
\omnidrive  & \phantom{0}26.3 & \phantom{0}44.8 & \phantom{0}32.4 & \phantom{0}14.5 & \phantom{0}50.6 & \phantom{0}28.2 & \phantom{0}45.9 & \phantom{00}3.2 & \phantom{0}66.7 & \phantom{0}20.6 & \phantom{0}27.3 & \phantom{0}28.6 & \phantom{0}15.2 & \phantom{00}0.0 & \phantom{0}0.0 & \phantom{00}0.0 & \phantom{00}0.0 & \phantom{0}23.8              \\
\drivemm    & \phantom{00}0.0              & \phantom{0}16.7              & \phantom{0}75.0              & \phantom{0}22.7              & \phantom{0}50.0              & \bestresult{\phantom{0}82.0} & \phantom{0}30.8              & \phantom{00}8.3              & \phantom{0}16.7              & \phantom{0}79.5              & \bestresult{\phantom{0}37.9} & \phantom{00}0.0              & \phantom{0}76.5              & \phantom{00}0.0     & \phantom{0}0.0 & \phantom{0}50.0    & \bestresult{100.0} & \phantom{0}38.0              \\
\bottomrule
\end{tabular*}
\end{table*}
\endgroup

\paragraph{Ego-agent scenarios.}
The detailed results for all ego-to-agent scenarios are listed in Table~\ref{tab:detailed_res_ego_agent}. We observe that LLMs perform very well at this task. However, we also notice a respectable performance of driving expert models. This is unsurprising since these scenarios are also part of existing benchmarks and datasets. The objective of driving experts mostly includes reasoning about agents in their close vicinity. The most challenging cases are \emph{pass agent}, and both scenarios of identifying \emph{stationary} objects in a certain position, \ie, \emph{left} and \emph{right} of the ego-vehicle. The performance on these relatively unknown but simple scenarios highlights a strong bias toward previously seen scenarios and underscores the limited zero-shot generalization capabilities of these models.

\begin{table*}[t]
\centering
\caption{
    \textbf{Ego-to-agent scenario evaluation.}
    Scenario-level performance comparison of LLMs, off-the-shelf VLMs, and driving expert VLMs for ego-to-agent scenarios.
    Accuracy grouped by scenario categories: Lead Agent (\textbf{LA}), Pass Agent (\textbf{PA}), Wait Pedestrian Cross (\textbf{WPC}), Follow Agent (\textbf{FA}), Overtake Agent (\textbf{OA}), Stationary Left Of Agent (\textbf{SLoA}), Stationary In Front Of Agent (\textbf{SiFoA}) and Average (\textbf{Avg}).
    The best results are highlighted in bold.
}
\label{tab:detailed_res_ego_agent}
\begin{tabular}{l*{7}{c}c}
\toprule
            & \textbf{LA}        & \textbf{PA}        & \textbf{WPC}        & \textbf{FA}        & \textbf{OA}                & \textbf{SLoA}              & \textbf{SiFoA}             & \textbf{Avg}       \\
\midrule
\llama      & 27.5              & 18.2              & \phantom{0}57.1    & 35.3              & \phantom{0}\phantom{0}0.0 & \phantom{0}\phantom{0}0.0 & \phantom{0}\phantom{0}0.0 & 19.7              \\
\deepseek   & 50.0              & \bestresult{72.7} & \phantom{0}57.1    & 70.6              & \bestresult{100.0}        & 50.0                      & \bestresult{100.0}        & 71.5             \\
\gptfo      & \bestresult{77.5} & 25.0              & \phantom{0}57.1    & 70.6              & \bestresult{100.0}        & \bestresult{100.0}        & \bestresult{100.0}        & \bestresult{75.8} \\
\midrule
\internvlob & 17.5              & 68.2              & \phantom{0}78.6    & 52.9              & \phantom{0}\phantom{0}0.0 & \phantom{0}\phantom{0}0.0 & \phantom{0}50.0           & 38.2              \\
\quenvlsb   & 10.0              & \phantom{0}0.0    & \bestresult{100.0} & 76.5              & \phantom{0}\phantom{0}0.0 & \phantom{0}\phantom{0}0.0 & \phantom{0}\phantom{0}0.0 & 26.6              \\
\internvlab & 32.5             & \phantom{0}9.1    & \bestresult{100.0} & \bestresult{94.1} & \phantom{0}\phantom{0}0.0 & \phantom{0}\phantom{0}0.0 & \phantom{0}50.0           & 40.8              \\
\midrule
\sennavlm   & 27.5              & 27.3              & \phantom{0}85.7    & 70.6              & \bestresult{100.0}        & \phantom{0}\phantom{0}0.0 & \phantom{0}\phantom{0}0.0 & 44.4              \\
\omnidrive  & 32.4 & 44.1 & \phantom{0}80.0 & 19.4 & \bestresult{100.0} & \phantom{00}0.0 & \phantom{0}25.0 & 43.0              \\
\drivemm    & \bestresult{77.5} & 29.6              & \phantom{0}92.9    & 88.2              & \phantom{0}\phantom{0}0.0 & \phantom{0}\phantom{0}0.0 & \phantom{0}50.0           & 48.3              \\
\bottomrule
\end{tabular}
\end{table*}

\paragraph{Agent-to-agent scenarios.}
The interactions between two agents not involving the ego-vehicle are the most difficult to reason about for the assessed models. In Table~\ref{tab:detailed_res_agent_agent}, \internvlab reaches the best average score across all scenarios. We can observe that \omnidrive is surprisingly good at scenarios like \emph{overtaking} and \emph{passing agent}. However, the model does not recognize pedestrian scenarios, \ie \emph{walk opposite} and \emph{walk alongside}, and is not able to identify positions of objects, \ie, \emph{left} and \emph{right} for both motion states \emph{stationary} and \emph{moving}. In contrast, \internvlab is exceptionally good at scenarios that can be identified by visual cues, but does not work well for scenarios spanned across different camera viewpoints, such as \emph{overtaking} or \emph{passing agent}. 

\begingroup
    \setlength\tabcolsep{3.5pt}
\begin{table*}[t]
\scriptsize
\centering
\caption{
    \textbf{Agent-to-agent scenario evaluation.}
    Scenario-level performance comparison of LLMs, off-the-shelf VLMs, and driving expert VLMs for Agent-to-agent scenarios.
    Accuracy grouped by scenario categories: Follow Agent (\textbf{FA}), Stationary Left Of Agent (\textbf{SLoA}), Stationary Right Of Agent (\textbf{SRoA}), Walk Alongside (\textbf{WA}), Stationary In Front Of Agent (\textbf{SiFoA}), Walk Opposite (\textbf{WO}), Lead Agent (\textbf{LA}), Wait Pedestrian Cross (\textbf{WPC}), Overtake Agent (\textbf{OA}), Stationary Behind Agent (\textbf{SBA}), Pass Agent (\textbf{PA}), Moving Left Of Agent (\textbf{MLoA}), Moving Right Of Agent (\textbf{MRoA}) and Average (\textbf{Avg}).
    The best results are highlighted in bold.
}
\label{tab:detailed_res_agent_agent}
\resizebox{\textwidth}{!}{
\begin{tabular*}{\textwidth}{l*{14}{c}c}
\toprule
            & \textbf{FA}                   & \textbf{SLoA}                 & \textbf{SRoA}                 & \textbf{WA}                   & \textbf{SiFoA}                & \textbf{WO}                   & \textbf{LA}                   & \textbf{WPC}        & \textbf{OA}                   & \textbf{SBA}                  & \textbf{PA}                   & \textbf{MLoA}       & \textbf{MRoA}       & \textbf{Avg}                  \\
\midrule
\llama      & \phantom{0}43.2              & \phantom{00}5.9              & \phantom{0}16.7              & \phantom{0}25.0              & \phantom{0}12.5              & \phantom{0}32.5              & \phantom{0}37.8              & \phantom{0}20.0    & \phantom{0}25.0              & \phantom{0}11.1              & \phantom{0}20.9              & \phantom{0}33.3    & \phantom{00}0.0    & \phantom{0}21.8              \\
\deepseek   & \phantom{0}52.3              & \phantom{00}5.9              & \phantom{00}8.3              & \phantom{0}65.0              & \phantom{00}6.3              & \bestresult{\phantom{0}95.0} & \phantom{0}26.7              & \phantom{0}96.0    & \phantom{0}25.0              & \phantom{00}5.6              & \phantom{0}55.8              & \phantom{0}33.3    & \bestresult{100.0} & \phantom{0}44.2              \\
\gptfo      & \bestresult{\phantom{0}65.9} & \phantom{00}5.9              & \phantom{00}5.6              & \bestresult{\phantom{0}75.0}              & \phantom{00}6.3              & \bestresult{\phantom{0}95.0} & \bestresult{\phantom{0}71.1} & \phantom{0}84.0    & \phantom{0}41.7              & \phantom{00}0.0              & \phantom{0}30.2              & \phantom{0}33.3    & \phantom{0}50.0    & \phantom{0}43.4              \\
\midrule
\internvlob & \phantom{00}6.8              & \phantom{00}2.9              & \phantom{00}2.8              & \phantom{0}17.5              & \bestresult{\phantom{0}25.0} & \phantom{0}27.5              & \phantom{0}24.4              & \phantom{0}40.0    & \bestresult{\phantom{0}50.0} & \phantom{0}16.7              & \bestresult{\phantom{0}58.1}              & \phantom{0}66.7    & \phantom{0}50.0    & \phantom{0}29.9              \\
\quenvlsb   & \phantom{0}63.6              & \phantom{0}23.5              & \phantom{0}36.1              & \phantom{0}55.0              & \bestresult{\phantom{0}25.0} & \phantom{0}45.0              & \phantom{0}20.0              & \phantom{0}84.0    & \phantom{0}25.0              & \phantom{0}38.9              & \phantom{0}20.9              & \phantom{00}0.0    & \phantom{0}50.0    & \phantom{0}37.5              \\
\internvlab & \phantom{0}43.2              & \bestresult{\phantom{0}61.8} & \bestresult{\phantom{0}63.9} & \phantom{0}72.5              & \phantom{00}0.0              & \phantom{0}72.5              & \phantom{0}26.7              & \bestresult{100.0} & \phantom{00}0.0              & \bestresult{\phantom{0}50.0} & \phantom{0}32.6              & \bestresult{100.0} & \phantom{0}50.0    & \bestresult{\phantom{0}51.8} \\
\midrule
\sennavlm   & \phantom{0}63.6              & \phantom{0}23.5              & \phantom{0}25.0              & \phantom{0}30.0              & \bestresult{\phantom{0}25.0} & \phantom{0}27.5              & \phantom{0}28.9              & \phantom{0}68.0    & \phantom{0}25.0              & \phantom{0}16.7              & \phantom{0}25.6              & \phantom{00}0.0    & \phantom{0}50.0    & \phantom{0}31.5              \\
\omnidrive  & \phantom{0}54.7 & \phantom{0}12.7 & \phantom{0}13.5 & \phantom{00}1.8 & \phantom{00}9.4 & \phantom{00}4.5 & \phantom{0}44.3 & \phantom{0}57.1 & \bestresult{\phantom{0}52.9} & \phantom{0}32.6 & \phantom{0}56.5 & \phantom{00}0.0 & \phantom{00}0.0 & \phantom{0}26.2              \\
\drivemm    & \phantom{0}34.1              & \phantom{0}35.3              & \phantom{0}55.6              & \phantom{0}32.5              & \phantom{00}6.3              & \phantom{0}17.5              & \phantom{0}31.1              & \phantom{0}24.0    & \phantom{00}8.3              & \phantom{0}27.8              & \phantom{0}11.6              & \phantom{0}33.3    & \phantom{0}50.0    & \phantom{0}28.3              \\
\bottomrule
\end{tabular*}
}
\end{table*}
\endgroup

\begin{figure}
    \tiny
    \centering
\begin{tikzpicture}[font=\ttfamily,
  mymatrix/.style={fill=PromptGray, matrix of nodes, nodes=typetag, row sep=1em, draw=none},
  mycontainer/.style={draw=gray, inner sep=1ex},
  typetag/.style={draw=gray, inner sep=1ex, anchor=west},
  title/.style={draw=none, color=PromptBorder, inner sep=0pt}
  ]
  \matrix[mymatrix] (mx1) {
    |[title]|System Prompt: \\
    \begin{minipage}{\textwidth}
You are a helpful traffic control expert specializing in the analysis and identification of temporal actions and maneuvers performed by various agents in diverse driving scenarios. 
    \end{minipage}\\
    |[title]|User Prompt: \\
    \begin{minipage}{\textwidth}
An agent refers to any participant in the traffic environment, including but not limited to cars, buses, construction vehicles, trucks, trailers, motorcycles, pedestrians, and bicycles. The traffic environment is captured using a sophisticated suite of sensors mounted on an ego vehicle, which is a standard passenger car.\\The sensor suite includes a Light Detection and Ranging (LiDAR) sensor labeled LIDAR\_TOP, mounted on the roof of the car. This LiDAR sensor provides high-precision 3D spatial data about the surrounding environment. LiDAR data is crucial for precise spatial positioning and size, which helps in differentiating vehicle types and detecting movement. For each agent, the LiDAR data includes the frame number, the center position of the agent (x, y, z) in meters relative to the LiDAR, and a heading angle in degrees relative to the LiDAR, and size dimensions such as width, length, and height in meters.\\In addition to the LiDAR, six cameras are strategically positioned around the vehicle to provide a 360-degree field of view. Camera data can aid in visual confirmation and potentially understanding intent. The first camera, CAM\_FRONT, is positioned directly in front of the LiDAR sensor and faces forward. To the right of CAM\_FRONT is CAM\_FRONT\_RIGHT, which is oriented at a 45-degree angle relative to the front-facing camera. On the right side of the car, CAM\_BACK\_RIGHT is positioned at a 135-degree angle relative to CAM\_FRONT. The rear-facing camera, CAM\_BACK, is oriented directly opposite to CAM\_FRONT. To the left of CAM\_FRONT is CAM\_FRONT\_LEFT, which is oriented at a -45-degree angle relative to the front-facing camera. Finally, CAM\_BACK\_LEFT is positioned on the left side of the car, oriented at a -135-degree angle relative to CAM\_FRONT. For each agent, the camera data includes the frame number, the center of the agent given with pixel location (x, y) in the image frame and in which camera is agent visible (e.g. CAM\_FRONT or CAM\_BACK).\\The system also integrates a GPS sensor, which provides the ego vehicle's precise global position. The GPS data includes the frame number, coordinates (x, y) in meters within a global coordinate system and the vehicle's orientation in radians relative to the global frame.\\Together, this comprehensive sensor suite enables detailed monitoring and analysis of the dynamic behaviors of all traffic agents. Your task is to leverage this data to identify and interpret the temporal actions and maneuvers of each agent within the traffic environment.\\The following are driving maneuvers and actions along with their respective descriptions: \\Agent 1 crossing street: Agent 1 (pedestrian) moves from one side of the road to the other, at a designated crossing point or intersection.\\Agent 1 jaywalking: Agent 1 (pedestrian) crosses the street outside of designated crossing areas or against traffic signals, often requiring heightened awareness of vehicle movements, quick decision-making to avoid conflicts, and potentially creating unpredictable interactions with other agents in the traffic environment.\\Agent 1 running: Agent 1 (pedestrian) is running and moves rapidly.\\Agent 1 decelerating: Agent 1 is reducing its speed, either gradually or abruptly, in response to traffic conditions, obstacles, or to comply with traffic rules, without coming to a complete stop.\\Agent 1 stopping: Agent 1 is reducing its speed, either gradually or abruptly, in response to traffic conditions, obstacles, or to comply with traffic rules and comes to a complete stop.\\Your input consists of sequential data, captured over 6 frames and 6 seconds with the described sensor suite.\\Ego:\\  Frame number: 0\\  x: 0.00\\  y: 0.00\\  rotation: 0.00\\  $\cdots$ \\  Frame number: 5\\  x: 7.95\\  y: 0.33\\  rotation: 0.03\\Agent 1:\\  Class: pedestrian\\  Frame number: 0\\  LiDAR x: 6.45\\  LiDAR y: -22.25\\  LiDAR rotation: 2.93\\  CAM x: 580.66\\  CAM y: 509.22\\  CAM: CAM\_BACK\\  $\cdots$ \\  Frame number: 5\\  LiDAR x: 4.63\\  LiDAR y: -29.92\\  LiDAR rotation: 2.90\\  CAM x: 696.33\\  CAM y: 513.36\\  CAM: CAM\_BACK\\Which of the following options best describes agent 1 driving maneuver?\\Options:\\A. Agent 1 crossing street\\B. Agent 1 jaywalking\\C. Agent 1 running\\D. Agent 1 decelerating\\E. Agent 1 stopping\\Please answer only with the letter of an option from the multiple choice list, e.g. A or B or C or D, and nothing else.
    \end{minipage} \\
    |[title]|Answer: \\
    \begin{minipage}{\textwidth}
    \gptfo: B \hfill Correct: B
    \end{minipage} \\
    |[title]|User Prompt: \\
    \begin{minipage}{\textwidth}
    Can you elaborate your decision? \hfill
    \end{minipage} \\
    |[title]|Answer: \\
    \begin{minipage}{\textwidth}
    \gptfo: Agent 1 is moving consistently and progressively away from the ego vehicle without any indication of stopping, running rapidly, or interacting with specific crossing points. The movement doesn't suggest crossing at a designated area, so the best interpretation given the information is that Agent 1 is likely crossing outside designated areas, which aligns with jaywalking behavior.
    \end{minipage} \\
  };
\end{tikzpicture}
\caption{\gptfo reasoning for a jaywalking scenario.}
\label{fig:prompt:gptfo_jaywalking}
\end{figure}

\begin{figure}
    \tiny
    \centering
\begin{tikzpicture}[font=\ttfamily,
  mymatrix/.style={fill=PromptGray, matrix of nodes, nodes=typetag, row sep=1em, draw=none},
  mycontainer/.style={draw=gray, inner sep=1ex},
  typetag/.style={draw=gray, inner sep=1ex, anchor=west},
  title/.style={draw=none, color=PromptBorder, inner sep=0pt}
  ]
  \matrix[mymatrix] (mx1) {
    |[title]|System Prompt: \\
    \begin{minipage}{\textwidth}
You are a helpful traffic control expert specializing in the analysis and identification of temporal actions and maneuvers performed by various agents in diverse driving scenarios. 
    \end{minipage}\\
    |[title]|User Prompt: \\
    \begin{minipage}{\textwidth}
An agent refers to any participant in the traffic environment, including but not limited to cars, buses, construction vehicles, trucks, trailers, motorcycles, pedestrians, and bicycles. The traffic environment is captured using a sophisticated suite of sensors mounted on an ego vehicle, which is a standard passenger car.\\The sensor suite includes a Light Detection and Ranging (LiDAR) sensor labeled LIDAR\_TOP, mounted on the roof of the car. This LiDAR sensor provides high-precision 3D spatial data about the surrounding environment. LiDAR data is crucial for precise spatial positioning and size, which helps in differentiating vehicle types and detecting movement. For each agent, the LiDAR data includes the frame number, the center position of the agent (x, y, z) in meters relative to the LiDAR, and a heading angle in degrees relative to the LiDAR, and size dimensions such as width, length, and height in meters.\\In addition to the LiDAR, six cameras are strategically positioned around the vehicle to provide a 360-degree field of view. Camera data can aid in visual confirmation and potentially understanding intent. The first camera, CAM\_FRONT, is positioned directly in front of the LiDAR sensor and faces forward. To the right of CAM\_FRONT is CAM\_FRONT\_RIGHT, which is oriented at a 45-degree angle relative to the front-facing camera. On the right side of the car, CAM\_BACK\_RIGHT is positioned at a 135-degree angle relative to CAM\_FRONT. The rear-facing camera, CAM\_BACK, is oriented directly opposite to CAM\_FRONT. To the left of CAM\_FRONT is CAM\_FRONT\_LEFT, which is oriented at a -45-degree angle relative to the front-facing camera. Finally, CAM\_BACK\_LEFT is positioned on the left side of the car, oriented at a -135-degree angle relative to CAM\_FRONT. For each agent, the camera data includes the frame number, the center of the agent given with pixel location (x, y) in the image frame and in which camera is agent visible (e.g. CAM\_FRONT or CAM\_BACK).\\The system also integrates a GPS sensor, which provides the ego vehicle's precise global position. The GPS data includes the frame number, coordinates (x, y) in meters within a global coordinate system and the vehicle's orientation in radians relative to the global frame.\\Together, this comprehensive sensor suite enables detailed monitoring and analysis of the dynamic behaviors of all traffic agents. Your task is to leverage this data to identify and interpret the temporal actions and maneuvers of each agent within the traffic environment.\\The following are driving maneuvers and actions along with their respective descriptions: \\Ego stopping: Ego is reducing its speed, either gradually or abruptly, in response to traffic conditions, obstacles, or to comply with traffic rules and comes to a complete stop.\\Ego turning left: Ego is executing a left turn at an intersection or junction.\\Ego performing u-turn: Ego is performing a 180-degree turn at an intersection or junction, reversing its direction of travel.\\Ego turning right: Ego is executing a right turn at an intersection or junction.\\Ego decelerating: Ego is reducing its speed, either gradually or abruptly, in response to traffic conditions, obstacles, or to comply with traffic rules, without coming to a complete stop.\\Your input consists of sequential data, captured over 6 frames and 6 seconds with the described sensor suite.\\Ego:\\  Frame number: 0\\  x: 0.00\\  y: 0.00\\  rotation: -0.00\\ \vdots\\  Frame number: 5\\  x: 7.75\\  y: 3.50\\  rotation: 0.82\\\\Which of the following options best describes ego driving maneuver?\\Options:\\A. Ego stopping\\B. Ego turning left\\C. Ego performing u-turn\\D. Ego turning right\\E. Ego decelerating\\Please answer only with the letter of an option from the multiple choice list, e.g. A or B or C or D, and nothing else.
    \end{minipage} \\
    |[title]|Answer: \\
    \begin{minipage}{\textwidth}
    \gptfo: B \hfill Correct: B
    \end{minipage} \\
    |[title]|User Prompt: \\
    \begin{minipage}{\textwidth}
    Can you elaborate your decision? \hfill
    \end{minipage} \\
    |[title]|Answer: \\
    \begin{minipage}{\textwidth}
    \gptfo: The ego vehicle's motion data shows increasing x and y coordinates along with a gradually increasing rotation angle from frames 0 to 5. The trajectory and rotation suggest the vehicle is moving in an arc to the left. This pattern aligns with the maneuver described as "Ego turning left," hence the choice B.
    \end{minipage} \\
  };
\end{tikzpicture}
\caption{\gptfo reasoning for an ego left turning scenario.}
\label{fig:prompt:gptfo_ego_lt}
\end{figure}

\begin{figure}[t!]
\centering
\begin{tikzpicture}
  \begin{axis}[
    ybar,
    symbolic x coords={A,B,C,D,E},
    xtick=data,
    width=\textwidth,
    bar width=3pt,
  ]
  \addplot [draw=C0, fill=C0!20] coordinates { (A,201) (B,182) (C,180) (D,201) (E,207) };
  \addlegendentry{Ground Truth}
  \addplot [draw=C1, fill=C1!20] coordinates { (A,410) (B,341) (C,167) (D,27) (E,26) };
  \addlegendentry{\llama}
  \addplot [draw=C2, fill=C2!20] coordinates { (A,148) (B,156) (C,160) (D,219) (E,166) };
  \addlegendentry{\deepseek}
  \addplot [draw=C3, fill=C3!20] coordinates { (A,244) (B,170) (C,203) (D,183) (E,171) };
  \addlegendentry{\gptfo}
  \addplot [draw=C4, fill=C4!20] coordinates { (A,195) (B,178) (C,188) (D,198) (E,212) };
  \addlegendentry{\internvlob}
  \addplot [draw=C5, fill=C5!20] coordinates { (A,185) (B,211) (C,206) (D,193) (E,176) };
  \addlegendentry{\quenvlsb}
  \addplot [draw=C6, fill=C6!20] coordinates { (A,195) (B,178) (C,188) (D,198) (E,212) };
  \addlegendentry{\internvlab}
  \addplot [draw=C7, fill=C7!20] coordinates { (A,484) (B,101) (C,35) (D,18) (E,9) };
  \addlegendentry{\sennavlm}
  \addplot [draw=C8, fill=C8!20] coordinates { (A,194) (B,197) (C,187) (D,197) (E,196) };
  \addlegendentry{\omnidrive}
  \addplot [draw=C9, fill=C9!20] coordinates { (A,188) (B,192) (C,198) (D,191) (E,202) };
  \addlegendentry{\drivemm}
  \end{axis}
\end{tikzpicture}
\caption{
    Distribution of individual letters that were the correct multiple-choice option, comparing ground truth data with the predictions of each evaluated method.
}
\label{fig:letter_distribution}
\end{figure}
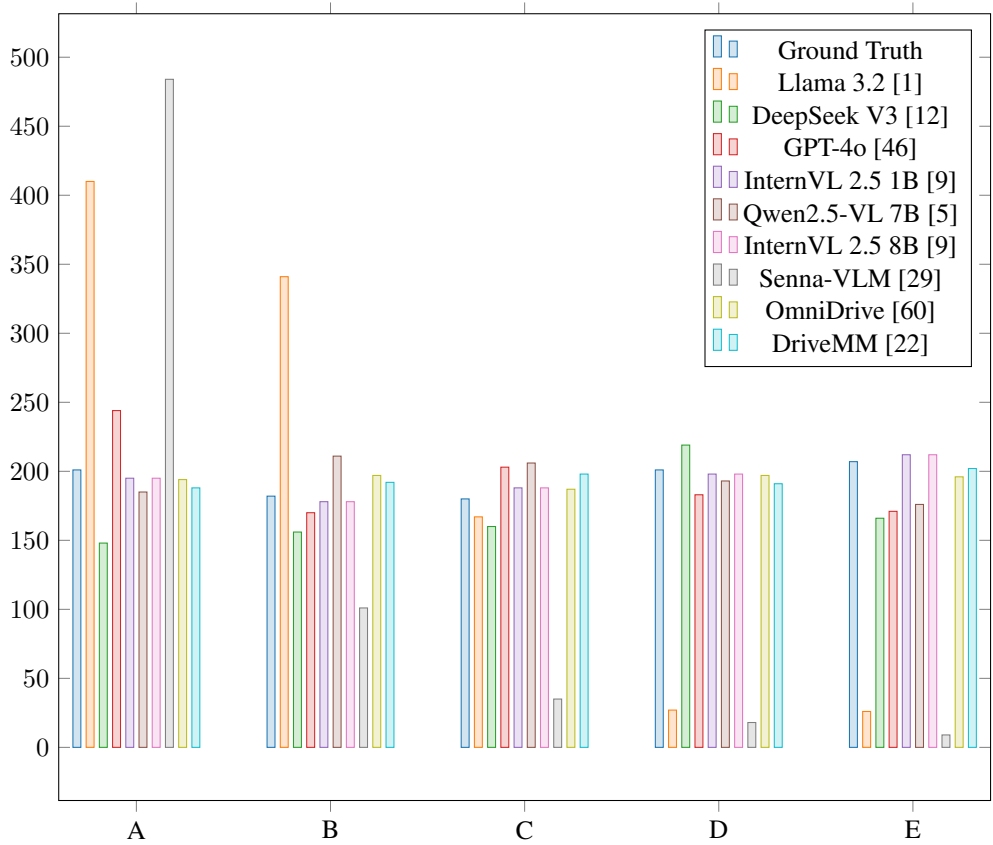

\begin{figure}[t]
    \centering
    \begin{subfigure}[b]{0.3\textwidth}
        \includegraphics[width=\textwidth]{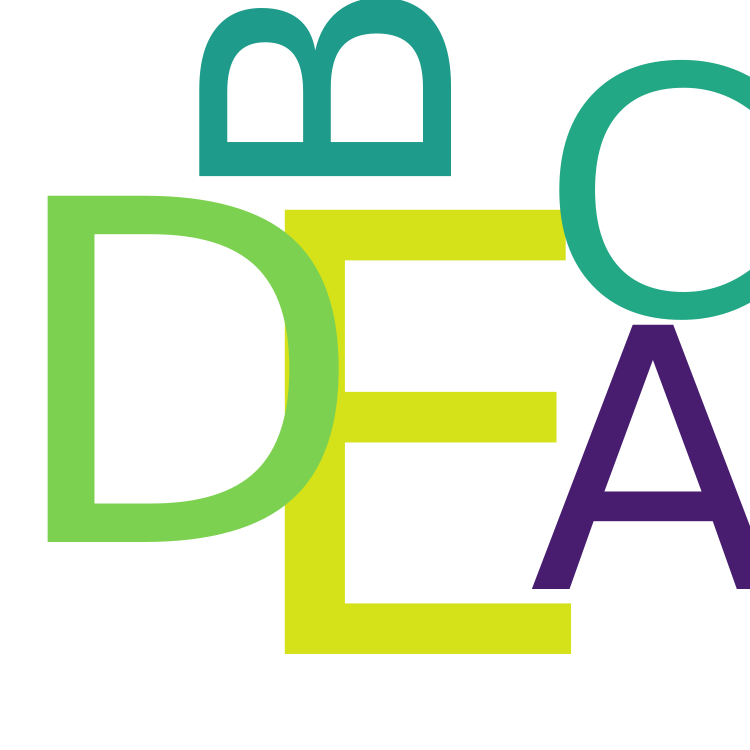}
        \caption{Ground truth}
        \label{fig:wordcloud:gt}
    \end{subfigure}%
    \hfill
    \begin{subfigure}[b]{0.3\textwidth}
        \includegraphics[width=\textwidth]{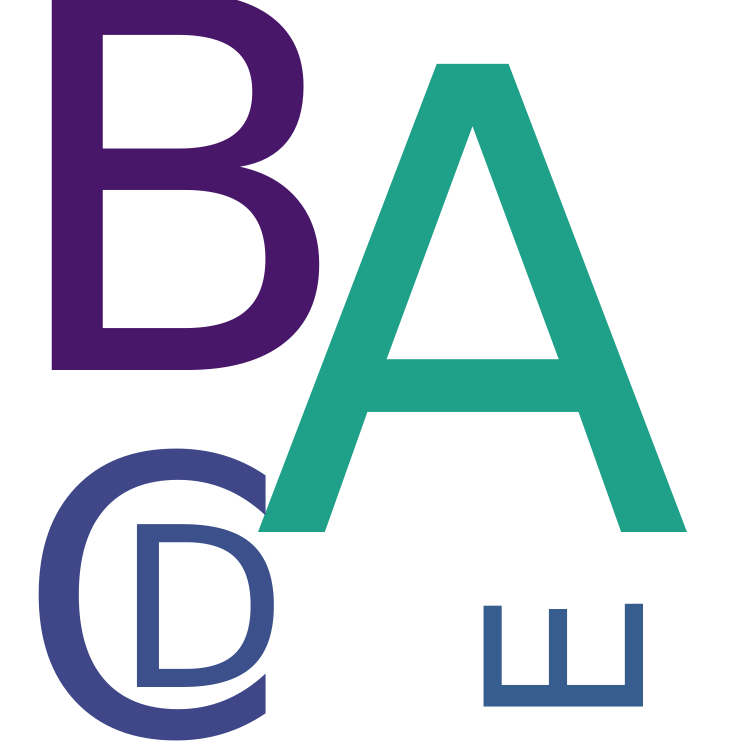}
    \caption{\llama}
        \label{fig:wordcloud:llama}
    \end{subfigure}%
    \hfill
    \begin{subfigure}[b]{0.3\textwidth}
        \includegraphics[width=\textwidth]{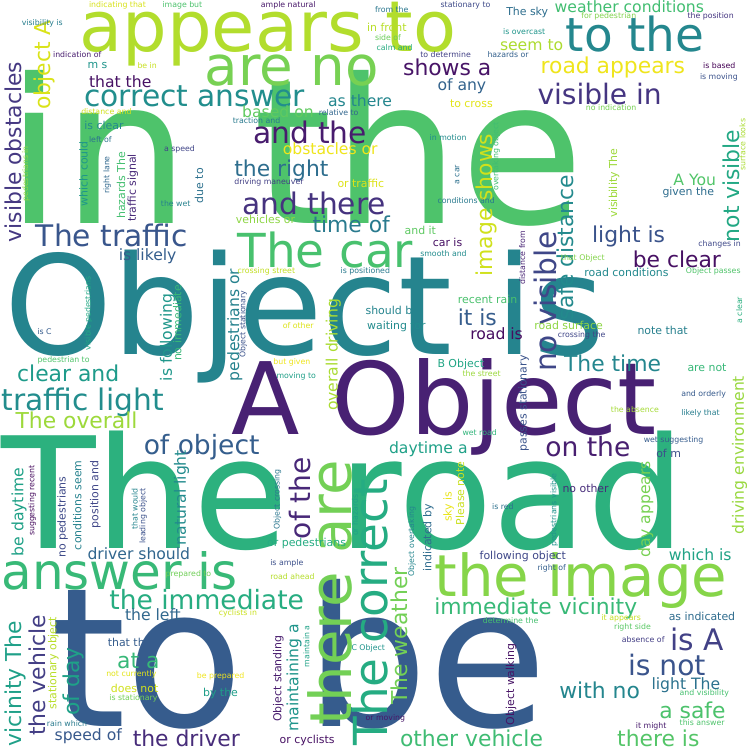}
    \caption{\sennavlm}
        \label{fig:wordcloud:senna}
    \end{subfigure}
    \caption{
        Distribution of individual letters that were the correct multiple-choice option \hyperref[fig:wordcloud:gt]{(a)}, and the outputs generated by the \llama \hyperref[fig:wordcloud:llama]{(b)} and \sennavlm \hyperref[fig:wordcloud:senna]{(c)} models.
    }
    \label{fig:wordcloud}
\end{figure}

\paragraph{Multiple-choice letter distribution.}
To mitigate potential biases arising from uneven ground truth letter distributions, our experimental design employs a uniform distribution across all multiple-choice options.
This critical aspect, visualized in Fig.~\ref{fig:letter_distribution}, addresses the known tendency of some LLMs to favor specific letter choices~\cite{zheng_2024_llmmcqbias}.
While methods such as \gptfo, \internvlab, and \drivemm appear to be free of this bias, our analysis reveals that \llama exhibits a preference for alphabetically earlier letters, a pattern similarly observed in the \sennavlm expert model.
Furthermore, \sennavlm demonstrates a tendency to disregard prompt instructions, such as the restriction to single-letter outputs, and occasionally includes irrelevant information (most probably from its training data), as illustrated in Fig.~\ref{fig:prompt:expert_senna_agent}.
The overall output vocabulary of Senna-VLM is depicted in the word cloud in Fig.~\ref{fig:wordcloud:senna}.

\clearpage
\section{Additional experiments}

\paragraph{Ablation of the query frame.}
The default model setting is to provide the initial frame of the sequence as the query frame.
We evaluate how the performance is affected by providing a later query frame as the initial input.
Specifically, we evaluate \omnidrive by querying with the first (default), middle, and last frames in each scenario.
Table~\ref{tab:omnidrive_additional_evals} shows that querying with the middle frame achieves the highest performance, presumably because it integrates more comprehensive temporal information than the alternatives.
Conversely, using the last frame as the query leads to a minor decrease in performance, which we hypothesize is due to limitations in OmniDrive's capacity for long-range temporal dependency modeling.

\paragraph{\omnidrive chain-of-though (CoT).}
OmniDrive leverages the chain-of-thought (CoT)~\cite{wei_2022_cot} capabilities of vision-language models (VLMs) to infer trajectories in a sequential, step-by-step manner, starting from scene descriptions, 3D grounding, and other relevant contextual information.
In our evaluation, we mirrored this approach by prompting the model at the conclusion of its internal CoT process.
This design choice aims to ensure the model encapsulates crucial spatio-temporal information necessary for answering our queries.
To ablate the impact of CoT, we conducted an evaluation of OmniDrive while disabling the CoT mechanism (Table~\ref{tab:omnidrive_additional_evals}).
As anticipated, the model's performance experienced a slight degradation without CoT, underscoring its positive contribution to overall performance.
 \begin{table*}
    \small
    \centering
    \caption{
        \omnidrive ablation by querying the model using $n^\text{th}$ frame and with or without the Chain of Thought (CoT).
    }
    \label{tab:omnidrive_additional_evals}
    \begin{tabular}{l c c c c c}
        \toprule
                              & \textbf{Ego} & \textbf{Ego-to-Agent} & \textbf{Agent} & \textbf{Agent-to-Agent} & \textbf{Average} \\
        \midrule
        $1^\text{st}$ w/ CoT            & 24.40 & 42.97 & 23.78 & 26.15 & 29.33            \\
        $3^\text{rd}$ w/ CoT  & 20.89        & \bestresult{46.92}                 & \bestresult{32.93}          & \bestresult{27.24}                   & \bestresult{32.00}            \\
        $6^\text{th}$ w/ CoT & 21.70        & 43.65                 & 23.60          & 26.72                   & 28.92            \\
        \midrule
        $1^\text{st}$ w/o CoT & \bestresult{29.91} & 35.87 & 21.76 & 25.39 & 28.23 \\
        \bottomrule
    \end{tabular}
\end{table*}

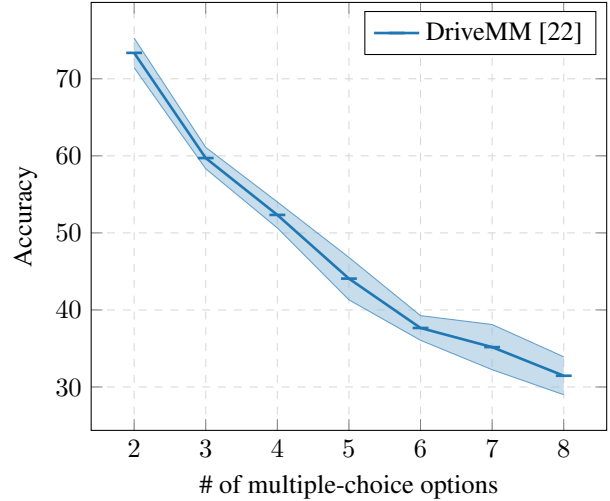
\begin{figure}
\centering
\begin{tikzpicture}
\begin{axis}[
    xlabel={\# of multiple-choice options},
    ylabel={Accuracy},
    grid=major,
    grid style={dashed, gray!30},
]
\addplot[mark=-,line width=1pt,mark size=3pt,C0] coordinates {
    (2,73.364)
    (3,59.712)
    (4,52.332)
    (5,44.066)
    (6,37.66)
    (7,35.176)
    (8,31.458)
};

\addplot[name path=top,color=C0!70] coordinates {
    (2,75.28249681)
    (3,61.10017146)
    (4,54.01996327)
    (5,46.81837715)
    (6,39.25851494)
    (7,38.10749791)
    (8,33.9174044)
};
\addplot[name path=bottom,color=C0!70] coordinates {
    (2,71.44550319)
    (3,58.32382854)
    (4,50.64403673)
    (5,41.31362285)
    (6,36.06148506)
    (7,32.24450209)
    (8,28.9985956)
};
\addplot[C0!50,fill opacity=0.5] fill between[of=top and bottom];

\addlegendentry{\drivemm}
\end{axis}
\end{tikzpicture}
\caption{
    Mean and variance of the DriveMMaccuracy across 5 independent evaluations, shown for increasing numbers of multiple-choice options (\eg, two: A-B; three: A-B-C).
    In each evaluation, the order of the multiple-choice options was randomly permuted.
}
\label{fig:number_mc_over_acc}
\end{figure}

\paragraph{Number of multiple-choice options.}
To allow for varying difficulty levels, our benchmark enables scaling the number of multiple-choice options.
Consistent with expectations, Fig.~\ref{fig:number_mc_over_acc} reveals a decrease in \drivemm model accuracy as the number of options grows.
This increase in options also leads to higher model variance, suggesting greater difficulty in selecting the correct answer.
We find that 5 multiple-choice options represents an optimal balance, preventing the benchmark from being trivial or excessively difficult.

\paragraph{Expert model LLMs with ground truth information.}
To further evaluate the expert models, we provided them with complete ground truth information in the prompt, mirroring the evaluation of LLM-only models.
As shown in Table~\ref{tab:expert_models_vlms}, this significantly degraded the performance of \sennavlm, suggesting that its training compromises the LLM's generalization capabilities, consistent with our findings in Fig.~\ref{fig:wordcloud:senna}.
The performance of \omnidrive remained consistent, indicating an inability to effectively utilize information outside its training data.
Notably, \drivemm's performance increased by nearly 9\%, surpassing \deepseek.
Given that DriveMM is trained on six diverse datasets, this highlights the importance of diverse training data in preserving the generalization abilities of LLMs.
 \begin{table*}
    \small
    \centering
    \caption{
        Expert model evaluation where complete ground truth information was included in the prompt, mirroring the evaluation of LLM-only models.
    }
    \label{tab:expert_models_vlms}
    \begin{tabular}{c l c c c c c}
        \toprule
        & \textbf{Ego} & \textbf{Ego-to-Agent} & \textbf{Agent} & \textbf{Agent-to-Agent} & \textbf{Average} \\
        \midrule
        \sennavlm & \phantom{0}3.46 & \phantom{0}0.00 & \phantom{0}4.35 & \phantom{0}4.81 & \phantom{0}3.16 \\
        \omnidrive & 27.02 & 34.61 & 22.75 & 29.27 & 28.41 \\
        \drivemm & \bestresult{53.11} & \bestresult{71.41} & \bestresult{39.54} & \bestresult{29.91} & \bestresult{48.49} \\
        \bottomrule
    \end{tabular}
\end{table*}

\section{Prompt examples}
\label{sec:supp:prompt_examples}
To demonstrate the required prompt adaptations, we include examples for all model types and scenario types (Ego, agent, ego-to-agent, and agent-to-agent) in the following.
In particular, Fig.~\ref{fig:prompt:llm_ego}--\ref{fig:prompt:llm_agent_agent} show the LLM prompts.
For the representative off-the-shelf VLMs, Fig.~\ref{fig:prompt:vlm_ego}--\ref{fig:prompt:vlm_agent_agent} provide prompt examples for InternVL 2.5~\cite{chen_2024_internvl}, while Fig.~\ref{fig:prompt:vlm_qwen_ego}--\ref{fig:prompt:vlm_qwen_agent_agent} provide prompt examples for Qwen2.5-VL 7B~\cite{bai_2023_qwenVL}.
For the driving expert models, we provide exemplary prompts for each scenario type for \sennavlm in Fig.~\ref{fig:prompt:expert_senna_ego}--\ref{fig:prompt:expert_senna_agent_agent}, OmniDrive~\cite{wang_2024_omnidrive} in Fig.~\ref{fig:prompt:expert_omnidrive_ego}--\ref{fig:prompt:expert_omnidrive_agent_agent}, and \drivemm in Fig.~\ref{fig:prompt:expert_drivemm_ego}--\ref{fig:prompt:expert_drivemm_agent_agent}.

\begin{figure}[t!]
    \tiny
    \centering
\begin{tikzpicture}[font=\ttfamily,
  mymatrix/.style={fill=PromptGray, matrix of nodes, nodes=typetag, row sep=1em, draw=none},
  mycontainer/.style={draw=gray, inner sep=1ex},
  typetag/.style={draw=gray, inner sep=1ex, anchor=west},
  title/.style={draw=none, color=PromptBorder, inner sep=0pt}
  ]
  \matrix[mymatrix] (mx1) {
    |[title]|System Prompt: \\
    \begin{minipage}{\textwidth}
You are a helpful traffic control expert specializing in the analysis and identification of temporal actions and maneuvers performed by various agents in diverse driving scenarios. 
    \end{minipage}\\
    |[title]|User Prompt: \\
    \begin{minipage}{\textwidth}
An agent refers to any participant in the traffic environment, including but not limited to cars, buses, construction vehicles, trucks, trailers, motorcycles, pedestrians, and bicycles. The traffic environment is captured using a sophisticated suite of sensors mounted on an ego vehicle, which is a standard passenger car.\\
The sensor suite includes a Light Detection and Ranging (LiDAR) sensor labeled LIDAR\_TOP, mounted on the roof of the car. This LiDAR sensor provides high-precision 3D spatial data about the surrounding environment. LiDAR data is crucial for precise spatial positioning and size, which helps in differentiating vehicle types and detecting movement. For each agent, the LiDAR data includes the frame number, the center position of the agent (x, y, z) in meters relative to the LiDAR, and a heading angle in degrees relative to the LiDAR, and size dimensions such as width, length, and height in meters.\\
In addition to the LiDAR, six cameras are strategically positioned around the vehicle to provide a 360-degree field of view. Camera data can aid in visual confirmation and potentially understanding intent. The first camera, CAM\_FRONT, is positioned directly in front of the LiDAR sensor and faces forward. To the right of CAM\_FRONT is CAM\_FRONT\_RIGHT, which is oriented at a 45-degree angle relative to the front-facing camera. On the right side of the car, CAM\_BACK\_RIGHT is positioned at a 135-degree angle relative to CAM\_FRONT. The rear-facing camera, CAM\_BACK, is oriented directly opposite to CAM\_FRONT. To the left of CAM\_FRONT is CAM\_FRONT\_LEFT, which is oriented at a -45-degree angle relative to the front-facing camera. Finally, CAM\_BACK\_LEFT is positioned on the left side of the car, oriented at a -135-degree angle relative to CAM\_FRONT. For each agent, the camera data includes the frame number, the center of the agent given with pixel location (x, y) in the image frame and in which camera is agent visible (e.g. CAM\_FRONT or CAM\_BACK).\\
The system also integrates a GPS sensor, which provides the ego vehicle's precise global position. The GPS data includes the frame number, coordinates (x, y) in meters within a global coordinate system and the vehicle's orientation in radians relative to the global frame.\\
Together, this comprehensive sensor suite enables detailed monitoring and analysis of the dynamic behaviors of all traffic agents. Your task is to leverage this data to identify and interpret the temporal actions and maneuvers of each agent within the traffic environment.\\
The following are driving maneuvers and actions along with their respective descriptions:\\
Ego stopping: Ego is reducing its speed, either gradually or abruptly, in response to traffic conditions, obstacles, or to comply with traffic rules and comes to a complete stop.\\
Ego turning left: Ego is executing a left turn at an intersection or junction.\\
Ego performing u-turn: Ego is performing a 180-degree turn at an intersection or junction, reversing its direction of travel.\\
Ego turning right: Ego is executing a right turn at an intersection or junction.\\
Ego decelerating: Ego is reducing its speed, either gradually or abruptly, in response to traffic conditions, obstacles, or to comply with traffic rules, without coming to a complete stop.
\\
Your input consists of sequential data, captured over 6 frames and 6 seconds with the described sensor suite.
\\
Ego:\\
Frame number: 0\\
x: 0.00\\
y: 0.00\\
rotation: 0.00\\
\vdots\\
Frame number: 5\\
x: 7.75\\
y: 3.50\\
rotation: 0.82\\
Which of the following options best describes ego driving maneuver?
Options:\\
A. Ego stopping\\
B. Ego turning left\\
C. Ego performing u-turn\\
D. Ego turning right\\
E. Ego decelerating\\
Please answer only with the letter of an option from the multiple choice list, e.g. A or B or C or D, and nothing else.
    \end{minipage} \\
    |[title]|Answer: \\
    \begin{minipage}{\textwidth}
    \gptfo: B \hfill \deepseek: D \hfill \llama: A \hfill Correct: B
    \end{minipage} \\
  };
\end{tikzpicture}
\caption{LLM Ego scenario prompt}
\label{fig:prompt:llm_ego}
\end{figure}
\begin{figure}[t!]
    \tiny
    \centering
\begin{tikzpicture}[font=\ttfamily,
  mymatrix/.style={fill=PromptGray, matrix of nodes, nodes=typetag, row sep=1em, draw=none},
  mycontainer/.style={draw=gray, inner sep=1ex},
  typetag/.style={draw=gray, inner sep=1ex, anchor=west},
  title/.style={draw=none, color=PromptBorder, inner sep=0pt}
  ]
  \matrix[mymatrix] (mx1) {
    |[title]|System Prompt: \\
    \begin{minipage}{\textwidth}
You are a helpful traffic control expert specializing in the analysis and identification of temporal actions and maneuvers performed by various agents in diverse driving scenarios. 
    \end{minipage}\\
    |[title]|User Prompt: \\
    \begin{minipage}{\textwidth}
You are a helpful traffic control expert specializing in the analysis and identification of temporal actions and maneuvers performed by various agents in diverse driving scenarios.|An agent refers to any participant in the traffic environment, including but not limited to cars, buses, construction vehicles, trucks, trailers, motorcycles, pedestrians, and bicycles. The traffic environment is captured using a sophisticated suite of sensors mounted on an ego vehicle, which is a standard passenger car.\\The sensor suite includes a Light Detection and Ranging (LiDAR) sensor labeled LIDAR\_TOP, mounted on the roof of the car. This LiDAR sensor provides high-precision 3D spatial data about the surrounding environment. LiDAR data is crucial for precise spatial positioning and size, which helps in differentiating vehicle types and detecting movement. For each agent, the LiDAR data includes the frame number, the center position of the agent (x, y, z) in meters relative to the LiDAR, and a heading angle in degrees relative to the LiDAR, and size dimensions such as width, length, and height in meters.\\In addition to the LiDAR, six cameras are strategically positioned around the vehicle to provide a 360-degree field of view. Camera data can aid in visual confirmation and potentially understanding intent. The first camera, CAM\_FRONT, is positioned directly in front of the LiDAR sensor and faces forward. To the right of CAM\_FRONT is CAM\_FRONT\_RIGHT, which is oriented at a 45-degree angle relative to the front-facing camera. On the right side of the car, CAM\_BACK\_RIGHT is positioned at a 135-degree angle relative to CAM\_FRONT. The rear-facing camera, CAM\_BACK, is oriented directly opposite to CAM\_FRONT. To the left of CAM\_FRONT is CAM\_FRONT\_LEFT, which is oriented at a -45-degree angle relative to the front-facing camera. Finally, CAM\_BACK\_LEFT is positioned on the left side of the car, oriented at a -135-degree angle relative to CAM\_FRONT. For each agent, the camera data includes the frame number, the center of the agent given with pixel location (x, y) in the image frame and in which camera is agent visible (e.g. CAM\_FRONT or CAM\_BACK).\\The system also integrates a GPS sensor, which provides the ego vehicle's precise global position. The GPS data includes the frame number, coordinates (x, y) in meters within a global coordinate system and the vehicle's orientation in radians relative to the global frame.\\Together, this comprehensive sensor suite enables detailed monitoring and analysis of the dynamic behaviors of all traffic agents. Your task is to leverage this data to identify and interpret the temporal actions and maneuvers of each agent within the traffic environment.\\The following are driving maneuvers and actions along with their respective descriptions: \\Agent 1 leading ego: Agent 1 travels ahead of ego at a similar speed while maintaining a consistent distance.\\Ego overtaking agent 2: Ego in the adjacent lane and moves ahead of agent 2 while both are in motion.\\Ego decelerating: Ego is reducing its speed, either gradually or abruptly, in response to traffic conditions, obstacles, or to comply with traffic rules, without coming to a complete stop.\\Ego waiting for pedestrian to cross: Ego comes to a stop or remains stationary, yielding the right-of-way to agent 2 who is crossing or preparing to cross the road, while maintaining awareness of the agent 2's movement and ensuring a safe distance until the crossing is complete.\\Ego passes stationary agent 2: Ego in the adjacent lane overtakes the stopped agent 2.\\Your input consists of sequential data, captured over 6 frames and 6 seconds with the described sensor suite.\\Ego:\\  Frame number: 0\\  x: 0.00\\  y: 0.00\\  rotation: 0.00\\ \vdots \\  Frame number: 5\\  x: 0.79\\  y: 0.03\\  rotation: 0.04\\Agent 2:\\  Frame number: 0\\  LiDAR x: -2.97\\  LiDAR y: 7.03\\  LiDAR rotation: -2.48\\  CAM x: 221.21\\  CAM y: 630.17\\  CAM: CAM\_FRONT\\ \vdots \\  Frame number: 5\\  LiDAR x: -5.41\\  LiDAR y: 4.08\\  LiDAR rotation: -2.30\\  CAM x: 828.79\\  CAM y: 578.67\\  CAM: CAM\_FRONT\_LEFT\\Which of the following options best describes ego driving behavior with respect to agent 2?\\Options:\\A. Agent 1 leading ego\\B. Ego overtaking agent 2\\C. Ego decelerating\\D. Ego waiting for pedestrian to cross\\E. Ego passes stationary agent 2\\Please answer only with the letter of an option from the multiple choice list, e.g. A or B or C or D, and nothing else.
    \end{minipage} \\
    |[title]|Answer: \\
    \begin{minipage}{\textwidth}
    \gptfo: C \hfill \deepseek: C \hfill \llama: B \hfill Correct: D
    \end{minipage} \\
  };
\end{tikzpicture}
\caption{LLM Ego-to-Agent scenario prompt}
\label{fig:prompt:llm_ego_agent}
\end{figure}
\begin{figure}[t!]
    \tiny
    \centering
\begin{tikzpicture}[font=\ttfamily,
  mymatrix/.style={fill=PromptGray, matrix of nodes, nodes=typetag, row sep=1em, draw=none},
  mycontainer/.style={draw=gray, inner sep=1ex},
  typetag/.style={draw=gray, inner sep=1ex, anchor=west},
  title/.style={draw=none, color=PromptBorder, inner sep=0pt}
  ]
  \matrix[mymatrix] (mx1) {
    |[title]|System Prompt: \\
    \begin{minipage}{\textwidth}
You are a helpful traffic control expert specializing in the analysis and identification of temporal actions and maneuvers performed by various agents in diverse driving scenarios. 
    \end{minipage}\\
    |[title]|User Prompt: \\
    \begin{minipage}{\textwidth}
An agent refers to any participant in the traffic environment, including but not limited to cars, buses, construction vehicles, trucks, trailers, motorcycles, pedestrians, and bicycles. The traffic environment is captured using a sophisticated suite of sensors mounted on an ego vehicle, which is a standard passenger car.\\The sensor suite includes a Light Detection and Ranging (LiDAR) sensor labeled LIDAR\_TOP, mounted on the roof of the car. This LiDAR sensor provides high-precision 3D spatial data about the surrounding environment. LiDAR data is crucial for precise spatial positioning and size, which helps in differentiating vehicle types and detecting movement. For each agent, the LiDAR data includes the frame number, the center position of the agent (x, y, z) in meters relative to the LiDAR, and a heading angle in degrees relative to the LiDAR, and size dimensions such as width, length, and height in meters.\\In addition to the LiDAR, six cameras are strategically positioned around the vehicle to provide a 360-degree field of view. Camera data can aid in visual confirmation and potentially understanding intent. The first camera, CAM\_FRONT, is positioned directly in front of the LiDAR sensor and faces forward. To the right of CAM\_FRONT is CAM\_FRONT\_RIGHT, which is oriented at a 45-degree angle relative to the front-facing camera. On the right side of the car, CAM\_BACK\_RIGHT is positioned at a 135-degree angle relative to CAM\_FRONT. The rear-facing camera, CAM\_BACK, is oriented directly opposite to CAM\_FRONT. To the left of CAM\_FRONT is CAM\_FRONT\_LEFT, which is oriented at a -45-degree angle relative to the front-facing camera. Finally, CAM\_BACK\_LEFT is positioned on the left side of the car, oriented at a -135-degree angle relative to CAM\_FRONT. For each agent, the camera data includes the frame number, the center of the agent given with pixel location (x, y) in the image frame and in which camera is agent visible (e.g. CAM\_FRONT or CAM\_BACK).\\The system also integrates a GPS sensor, which provides the ego vehicle's precise global position. The GPS data includes the frame number, coordinates (x, y) in meters within a global coordinate system and the vehicle's orientation in radians relative to the global frame.\\Together, this comprehensive sensor suite enables detailed monitoring and analysis of the dynamic behaviors of all traffic agents. Your task is to leverage this data to identify and interpret the temporal actions and maneuvers of each agent within the traffic environment.\\The following are driving maneuvers and actions along with their respective descriptions: \\Agent 1 turning left: Agent 1 is executing a left turn at an intersection or junction.\\Agent 1 changing lanes: Agent 1 is transitioning from its current lane to an adjacent lane.\\Agent 1 reversing: Agent 1 is moving in reverse, either to park, navigate a tight space, or adjust its position.\\Agent 1 decelerating: Agent 1 is reducing its speed, either gradually or abruptly, in response to traffic conditions, obstacles, or to comply with traffic rules, without coming to a complete stop.\\Agent 1 performing u-turn: Agent 1 is performing a 180-degree turn at an intersection or junction, reversing its direction of travel.\\Your input consists of sequential data, captured over 6 frames and 6 seconds with the described sensor suite.\\Ego:\\  Frame number: 0\\  x: 0.00\\  y: 0.00\\  rotation: 0.00\\ \vdots \\ Frame number: 5\\  x: 15.50\\  y: -0.17\\  rotation: -0.01\\Agent 1:\\  Class: car\\  Frame number: 0\\  LiDAR x: 5.17\\  LiDAR y: -5.98\\  LiDAR rotation: 1.47\\  CAM x: 1326.79\\  CAM y: 643.39\\  CAM: CAM\_BACK\_RIGHT\\ \vdots \\ Frame number: 5\\  LiDAR x: 5.91\\  LiDAR y: -11.37\\  LiDAR rotation: 1.52\\  CAM x: 1599.80\\  CAM y: 480.51\\  CAM: CAM\_BACK\_RIGHT\\Which of the following options best describes agent 1 driving maneuver?\\Options:\\A. Agent 1 turning left\\B. Agent 1 changing lanes\\C. Agent 1 reversing\\D. Agent 1 decelerating\\E. Agent 1 performing u-turn\\Please answer only with the letter of an option from the multiple choice list, e.g. A or B or C or D, and nothing else.
    \end{minipage} \\
    |[title]|Answer: \\
    \begin{minipage}{\textwidth}
    \gptfo: C. \hfill \deepseek: B \hfill \llama: B \hfill Correct: B
    \end{minipage} \\
  };
\end{tikzpicture}
\caption{LLM Agent scenario prompt}
\label{fig:prompt:llm_agent}
\end{figure}
\begin{figure}[t!]
    \tiny
    \centering
\begin{tikzpicture}[font=\ttfamily,
  mymatrix/.style={fill=PromptGray, matrix of nodes, nodes=typetag, row sep=1em, draw=none},
  mycontainer/.style={draw=gray, inner sep=1ex},
  typetag/.style={draw=gray, inner sep=1ex, anchor=west},
  title/.style={draw=none, color=PromptBorder, inner sep=0pt}
  ]
  \matrix[mymatrix] (mx1) {
    |[title]|System Prompt: \\
    \begin{minipage}{\textwidth}
You are a helpful traffic control expert specializing in the analysis and identification of temporal actions and maneuvers performed by various agents in diverse driving scenarios. 
    \end{minipage}\\
    |[title]|User Prompt: \\
    \begin{minipage}{\textwidth}
An agent refers to any participant in the traffic environment, including but not limited to cars, buses, construction vehicles, trucks, trailers, motorcycles, pedestrians, and bicycles. The traffic environment is captured using a sophisticated suite of sensors mounted on an ego vehicle, which is a standard passenger car.\\The sensor suite includes a Light Detection and Ranging (LiDAR) sensor labeled LIDAR\_TOP, mounted on the roof of the car. This LiDAR sensor provides high-precision 3D spatial data about the surrounding environment. LiDAR data is crucial for precise spatial positioning and size, which helps in differentiating vehicle types and detecting movement. For each agent, the LiDAR data includes the frame number, the center position of the agent (x, y, z) in meters relative to the LiDAR, and a heading angle in degrees relative to the LiDAR, and size dimensions such as width, length, and height in meters.\\In addition to the LiDAR, six cameras are strategically positioned around the vehicle to provide a 360-degree field of view. Camera data can aid in visual confirmation and potentially understanding intent. The first camera, CAM\_FRONT, is positioned directly in front of the LiDAR sensor and faces forward. To the right of CAM\_FRONT is CAM\_FRONT\_RIGHT, which is oriented at a 45-degree angle relative to the front-facing camera. On the right side of the car, CAM\_BACK\_RIGHT is positioned at a 135-degree angle relative to CAM\_FRONT. The rear-facing camera, CAM\_BACK, is oriented directly opposite to CAM\_FRONT. To the left of CAM\_FRONT is CAM\_FRONT\_LEFT, which is oriented at a -45-degree angle relative to the front-facing camera. Finally, CAM\_BACK\_LEFT is positioned on the left side of the car, oriented at a -135-degree angle relative to CAM\_FRONT. For each agent, the camera data includes the frame number, the center of the agent given with pixel location (x, y) in the image frame and in which camera is agent visible (e.g. CAM\_FRONT or CAM\_BACK).\\The system also integrates a GPS sensor, which provides the ego vehicle's precise global position. The GPS data includes the frame number, coordinates (x, y) in meters within a global coordinate system and the vehicle's orientation in radians relative to the global frame.\\Together, this comprehensive sensor suite enables detailed monitoring and analysis of the dynamic behaviors of all traffic agents. Your task is to leverage this data to identify and interpret the temporal actions and maneuvers of each agent within the traffic environment.\\The following are driving maneuvers and actions along with their respective descriptions: \\Agent 1 stationary to the right of agent 2: Agent 1 is fully stopped and remains stationary to the right of agent 2, which is also stationary, such as when waiting at a traffic light or in a parking lot.\\Agent 1 overtaking agent 2: Agent 1 in the adjacent lane and moves ahead of agent 2 while both are in motion.\\Agent 1 stationary in front of agent 2: Agent 1 is fully stopped and remains stationary in front of agent 2, such as when waiting at a traffic light, in a parking lot, or any other situation requiring queuing.\\Agent 1 stationary to the left of agent 2: Agent 1 is fully stopped and remains stationary to the left of agent 2, which is also stationary, such as when waiting at a traffic light or in a parking lot.\\Agent 1 moving to the right of agent 2: Agent 1 is traveling in parallel to the right of agent 2 (e.g., in adjacent lanes or side by side), with one vehicle maintaining a rightward offset relative to the other. This could occur during lane-matched driving on a multi-lane road or synchronized movement from a traffic light.\\Your input consists of sequential data, captured over 6 frames and 6 seconds with the described sensor suite.\\Ego:\\ \vdots \\Agent 1:\\  Class: car\\  Frame number: 0\\  LiDAR x: -11.89\\  LiDAR y: 28.12\\  LiDAR rotation: -1.58\\  CAM x: 291.60\\  CAM y: 546.14\\  CAM: CAM\_FRONT\\ \vdots \\ Frame number: 5\\  LiDAR x: -12.35\\  LiDAR y: -27.44\\  LiDAR rotation: -1.57\\  CAM x: 1223.86\\  CAM y: 520.56\\  CAM: CAM\_BACK\\Agent 2:\\  Class: car\\  Frame number: 0\\  LiDAR x: -15.65\\  LiDAR y: 26.08\\  LiDAR rotation: -1.58\\  CAM x: 90.13\\  CAM y: 564.20\\  CAM: CAM\_FRONT\\ \vdots \\ Frame number: 5\\  LiDAR x: -15.99\\  LiDAR y: -26.44\\  LiDAR rotation: -1.52\\  CAM x: 1351.97\\  CAM y: 527.46\\  CAM: CAM\_BACK\\Which of the following options best describes agent 1 driving behaviour with respect to agent 2?\\Options:\\A. Agent 1 stationary to the right of agent 2\\B. Agent 1 overtaking agent 2\\C. Agent 1 stationary in front of agent 2\\D. Agent 1 stationary to the left of agent 2\\E. Agent 1 moving to the right of agent 2\\Please answer only with the letter of an option from the multiple choice list, e.g. A or B or C or D, and nothing else.
    \end{minipage} \\
    |[title]|Answer: \\
    \begin{minipage}{\textwidth}
    \gptfo: B \hfill \deepseek: E \hfill \llama: E \hfill Correct: B
    \end{minipage} \\
  };
\end{tikzpicture}
\caption{LLM Agent-to-Agent scenario prompt}
\label{fig:prompt:llm_agent_agent}
\end{figure}

\input{prompts/vlm_ego}
\input{prompts/vlm_ego_agent}
\input{prompts/vlm_agent}
\input{prompts/vlm_agent_agent}

\input{prompts/vlm_qwen_ego}
\input{prompts/vlm_qwen_ego_agent}
\input{prompts/vlm_qwen_agent}
\input{prompts/vlm_qwen_agent_agent}

\input{prompts/expert_senna_ego}
\input{prompts/expert_senna_ego_agent}
\input{prompts/expert_senna_agent}
\input{prompts/expert_senna_agent_agent}

\input{prompts/expert_omnidrive_ego}
\input{prompts/expert_omnidrive_ego_agent}
\input{prompts/expert_omnidrive_agent}
\input{prompts/expert_omnidrive_agent_agent}

\input{prompts/expert_drivemm_ego}
\input{prompts/expert_drivemm_ego_agent}
\input{prompts/expert_drivemm_agent}
\input{prompts/expert_drivemm_agent_agent}

\end{document}